\documentclass{article} 
\usepackage{iclr2027_conference,times}

\usepackage[dvipsnames]{xcolor}
\definecolor{citecolor}{HTML}{1520a6}
\definecolor{linkcolor}{HTML}{900603}
\definecolor{bordo}{HTML}{ff0040}
\definecolor{comments}{HTML}{509187}
\definecolor{chestnut}{HTML}{CC6633}

\usepackage{hyperref}
\hypersetup{
  colorlinks=true,
  citecolor=citecolor,
  linkcolor=linkcolor,
}

\usepackage{url}
\usepackage{comment}
\usepackage{amsmath}
\usepackage{amsthm}
\usepackage{amssymb}
\usepackage{wrapfig}
\usepackage{colortbl}
\usepackage{xcolor}

\usepackage{algorithm}
\usepackage{algpseudocode}
\algrenewcommand\algorithmicindent{1.0em}
\algnewcommand{\LineComment}[1]{\State \textcolor{comments}{$\triangleright$ \textit{#1}}}

\usepackage{hyperref}
\hypersetup{
  colorlinks=true,
  citecolor=citecolor,
  linkcolor=linkcolor,
  urlcolor=black,
}
\usepackage[capitalize, noabbrev]{cleveref}
\usepackage{thmtools}

\definecolor{osisblue}{HTML}{0072B2}
\definecolor{osiscomparison}{HTML}{666666}
\definecolor{osisgreenbase}{HTML}{D2E6E3}
\colorlet{osisgreen}{osisgreenbase!50!white}

\theoremstyle{plain}
\declaretheorem[name=Theorem, numberwithin=section,
  refname={Theorem,Theorems}, Refname={Theorem,Theorems}]{theorem}

\declaretheorem[name=Definition, sibling=theorem,
  refname={Definition,Definitions}, Refname={Definition,Definitions}]{definition}

\usepackage[createShortEnv]{proof-at-the-end}

\theoremstyle{definition}

\makeatletter
\def\section{\@startsection {section}{1}{\z@}{-1.4ex plus
    -0.1ex minus -.1ex}{1.0ex plus 0.1ex
minus 0.1ex}{\large\sc\raggedright}}

\def\subsection{\@startsection{subsection}{2}{\z@}{-1.2ex plus
-0.1ex minus -.1ex}{0.8ex plus .1ex}{\normalsize\sc\raggedright}}
\def\subsubsection{\@startsection{subsubsection}{3}{\z@}{-1.1ex
plus      -0.2ex minus -.2ex}{0.2ex plus
.2ex}{\normalsize\sc\raggedright}}
\def\paragraph{\@startsection{paragraph}{4}{\z@}{0ex}{-1em}{\normalsize\bf}}
\def\subparagraph{\@startsection{subparagraph}{5}{\z@}{1.5ex plus
  0.5ex minus .2ex}{-1em}{\normalsize\sc}}

\crefname{section}{\S\@gobble}{\S\@gobble}
\crefname{subsection}{\S\@gobble}{\S\@gobble}

\newcommand{\zerodisplayskips}{%
  \setlength{\abovedisplayskip}{2mm}%
  \setlength{\belowdisplayskip}{2mm}%
  \setlength{\abovedisplayshortskip}{1mm}%
  \setlength{\belowdisplayshortskip}{1mm}}
\appto{\normalsize}{\zerodisplayskips}
\appto{\small}{\zerodisplayskips}
\appto{\footnotesize}{\zerodisplayskips}

\makeatother

\usepackage{mathtools}
\usepackage{booktabs}
\usepackage{tabularx}
\usepackage{array}
\newcolumntype{Y}{>{\raggedright\arraybackslash}X}

\usepackage{enumitem}
\usepackage{graphicx}
\usepackage{subcaption}
\usepackage{mathtools}
\usepackage{microtype}
\usepackage{float}
\usepackage{url}
\usepackage{booktabs}
\usepackage{enumitem}

\usepackage{tikz}
\usetikzlibrary{positioning, calc}
\usetikzlibrary{
    arrows.meta,shapes.geometric,decorations.markings,calc
}

\usepackage{newtxmath}
\usepackage{wrapfig}
\usepackage{booktabs}
\usepackage{fontawesome5}

\usepackage{multirow}
\usepackage{xspace}

\renewcommand{\thefootnote}{*}

\title{Amortised inference through\\one-step implicit sampling}

\iclrfinalcopy
\iclrpreprintcopy

\author{%
    Vincent Pauline\textsuperscript{*123}
    \quad
    Kirill Tamogashev\textsuperscript{*3}
    \quad
    Arran Carter\textsuperscript{3}
    \quad
    Sanghyeok Choi\textsuperscript{3}
    \\\bf
    Stefan Bauer\textsuperscript{124}
    \quad
    Esmeralda S. Whitammer\textsuperscript{34}
    \\
    \textsuperscript{1}Technical University of Munich
    \quad
    \textsuperscript{2}Munich Center for Machine Learning
    \\
    \textsuperscript{3}University of Edinburgh
    \quad
    \textsuperscript{4}CIFAR Fellow
    \quad
    \textsuperscript{*}Equal contribution\\[-1em]
}

\newcommand{\osis}{\textsc{OSIS}\xspace}

\newcommand{\Wp}[1]{\mathcal{W}_{#1}}

\newcommand{\sg}[1]{\text{sg}\left[#1\right]}
\newcommand{\W}[2]{\mathcal{W}_2(#1, #2)}
\newcommand{\sqW}[2]{\mathcal{W}^2_2(#1, #2)}
\newcommand{\KL}[2]{\operatorname{KL}\left(#1\!\parallel\!#2 \right)}
\newcommand{\FI}[2]{\operatorname{FI}\left(#1\!\parallel\!#2 \right)}

\newcommand{\pd}[2]{\frac{\partial}{\partial #1}#2}

\newcommand{\dv}{\nabla\cdot}
\newcommand{\dd}{\mathrm{d}}
\newcommand{\Id}{\operatorname{Id}}
\newcommand{\R}{\mathbb{R}}

\newcommand{\eg}{\emph{e.g.}}
\newcommand{\ie}{\emph{i.e.}}

\providecommand{\std}[1]{\ensuremath{{\scriptstyle\pm #1}}}

\begin{document}
\maketitle
{\renewcommand{\thefootnote}{*}\footnotetext{
\small
Correspondence to \href{vincent.paulinef@gmail.com}{\texttt{vincent.paulinef@gmail.com}} and \href{kirill.tamogashev@ed.ac.uk}{\texttt{kirill.tamogashev@ed.ac.uk}}.
}}

\begin{abstract}
    Efficient sampling from Bayesian posterior distributions in high dimensions remains a hard problem, despite advances in theory and algorithms. 
    Multiple steps are usually required to draw unbiased samples, for example, iterations of an MCMC algorithm or integration steps in a diffusion-based amortised sampler. 
    However, scalable one-step amortised inference remains a challenging open problem.
    In this work, we build upon recent progress in data-driven generative modelling to offer a possible solution for one-step amortised sampling. 
    Our \emph{one-step implicit sampler} (\osis) uses short-time MCMC dynamics to define a local target for the evolution of an implicitly modelled distribution.
    In the asymptotic limit, the modelled distribution is a fixed point of the MCMC kernel and thus samples from the target.
    We analyse the convergence properties of \osis and show connections with other generative modelling and sampling algorithms.
    We study various design choices and show that \osis performs well compared to multi-step amortised samplers 
    and non-amortised MCMC baselines while being far more efficient at sampling time.
\end{abstract}

\section{Introduction}\label{sec:intro}

\paragraph{The sampling problem.}\looseness=-1
Many problems in physics, biology, statistics, and machine learning \citep{welling2011bayesian, wirnsberger2020targeted,noe2019boltzmann,holdijk2023stochastic} require drawing samples from a distribution whose density is known up to a normalising constant. Formally, this means sampling from a distribution defined via an \emph{energy function} \mbox{$\mathcal{E}:\R^d\to\R$}, with density
\begin{equation}\label{eq:eq1}
    \pi(x) = {e^{-\mathcal{E}(x)}}/Z,\quad Z={\int_{\mathbb{R}^d} e^{-\mathcal{E}(x)}\,\dd x}.
\end{equation}
A typical instance of this problem is Bayesian inference: a posterior $\pi(x\mid y)\propto p(x)p(y\mid x)$ over a variable $x$ given an observation $y$ has the form in \eqref{eq:eq1} with $\mathcal{E}(x)=-\log p(x)-\log p(y\mid x)$.

\looseness=-1
The classic way to sample from a distribution as in \eqref{eq:eq1} is by using (training-free) Monte Carlo algorithms. Many such algorithms draw samples by simulating an ergodic chain using $\pi$-invariant kernels for a sufficiently long time. Popular Monte Carlo algorithms include Metropolis-Hastings \citep{metropolis1953equation, hastings1970monte}, unadjusted and Metropolis-adjusted Langevin \citep{rossky1978brownian, besag1994mala, roberts1996exponential}, Hamiltonian Monte Carlo \citep{duane1987hybrid,neal2011mcmc}, and sequential Monte Carlo \citep{del2006sequential}. The use of these methods comes at the cost of running chains for many steps, or with many simulated particles, to obtain high-quality samples from the target.

\looseness=-1
Generative modelling algorithms that learn to imitate unbiased samples from a distribution  -- \ie, approximate an empirical measure -- scale quite successfully, as evidenced by many large-scale applications of deep learning \citep{kong2020diffwave, hoogeboom2022equivariant, sahoo2024simple}. Meanwhile, generative modelling algorithms that rely only on a density function, without unbiased samples, continue to lag behind, and scaling them remains a challenge despite the deep significance of this problem.

\paragraph{Amortised sampling.}\looseness=-1
Deep learning has brought about new possibilities for introducing learnable components into sampling algorithms. Such components allow to spend more time and compute for training, yet gain speed when sampling and exploit generalisable structure in the target. Amortisation techniques include learning MCMC transition kernels \citep{song2017nice, samsonov2022local, wang2024reinforcement}, amortising MCMC dynamics \citep{li2017approximate}, and training density estimators to sample from unnormalised distributions \citep{rezende2015variational, midgley2022flow}.

\looseness=-1
The most successful family of amortised samplers adopts diffusion models \citep{sohl2015deep, ho2020denoising, song2020score} as approximators of the target distribution \citep[\eg,][]{zhang2021path, vargas2023denoising, richter2024improved, sendera2024improved, chen2025sequential, albergo2025nets, berner2026from,havens2025adjoint}. Diffusion samplers introduce a learnable stochastic process to model the transition dynamics from a known distribution to a target, usually by approximately reversing a fixed noising process. In contrast to diffusion models trained using a variational bound on the likelihood of empirical samples, diffusion samplers are trained using stochastic optimal control objectives \citep{zhang2021path, berner2024optimal} or path-space divergences \citep{richter2020vargrad, lahlou2023theory}, among other choices. Recent work has found and exploited connections between diffusion samplers and amortisation of SMC \citep{choi2025reinforced}.

\begin{figure}[t!]
    \centering
    \input{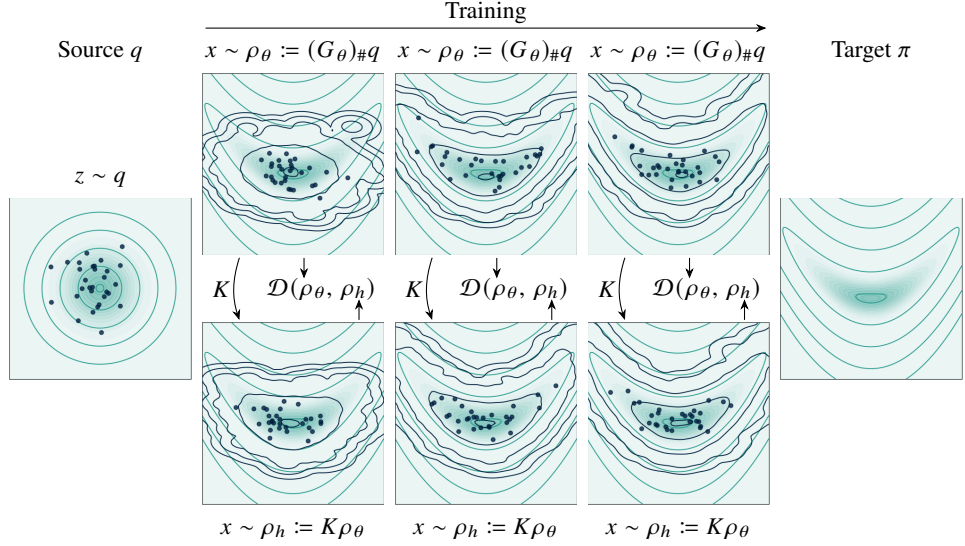}\\[-0.5em]
    \caption{
        \textbf{\osis evolves an implicit modelled distribution $\rho_\theta$ towards a target $\pi$ available through its unnormalised density (green contours).} $G_\theta$ maps samples $z$ the source distribution $q$ to samples $x=G_\theta(z)$ from the modelled distribution $\rho_\theta\coloneqq(G_\theta)_\#q$ (blue KDE contours). This pushforward distribution is evolved by a $\pi$-invariant MCMC kernel $K$, giving $\rho_h\coloneqq K\rho_{{\theta}}$. The parameters $\theta$ are updated to minimise an empirical divergence between $\rho_\theta$ and $\rho_h$. Upon convergence, $\rho_\theta\approx\pi$.
    }\vspace*{-1em}
\end{figure}

\paragraph{One-step amortised sampling.}
Although diffusion samplers allow for a considerable inference speed-up by avoiding long-chain MCMC, they require simulation of a neural stochastic dynamical system during both training and sampling.
Amortised sampling with a \emph{single} neural function evaluation, however, remains a challenge.
Although it is possible to use normalising flows for one-step sampling \citep{rezende2015variational}, they require the invertibility of a neural network and efficient computation of its Jacobian, both of which remain significant constraints. More recent work shows that such methods scale poorly and are prone to mode collapse \citep{he2025no}. These methods can further be placed within a Monte Carlo outer loop and augmented with MCMC sampling \citep{midgley2022flow}.

In an attempt to design an efficient one-step sampler, we take a different approach, drawing inspiration from 
the novel paradigm in generative modelling termed drifting models 
\citep{deng2026generative, gretton2026wasserstein}. Drifting models evolve a implicit distribution towards an empirical target by defining a local direction for its evolution via a \emph{drifting field} estimated from batches of target and generated samples. Such an approach, cannot be directly applied to sampling, as it requires access to samples from the target distribution to compute the drifting field. 

\paragraph{Our contributions.} We bypass this limitation by using an MCMC kernel invariant to the target distribution to implicitly define a drifting field. This leads us to the first main contribution:
\begin{enumerate}[nosep,label=(\arabic*), series=cont,left=0pt]
    \item We propose an algorithm for training a one-step implicit sampler (\osis) using an objective that enforces invariance of the sampler's output distribution to a target-invariant MCMC kernel (\Cref{subsec:osis}).
\end{enumerate}
There have been prior attempts to design a one-step amortised sampling algorithm, using kernel density approximations to circumvent data unavailability \citep{cao2026one} and using adversarial training \citep{li2017approximate}. 
We discuss the connections between \osis, drifting models, computational Wasserstein gradient flows, and this prior work in \Cref{sec:connections}.
We also study the theoretical properties of \osis, showing global convergence under assumptions similar to those in the MCMC literature:
\begin{enumerate}[cont]
    \item We prove convergence of \osis under asymptotic conditions and constraints on the target (\Cref{sec:theory}).
\end{enumerate}
To achieve scalability and efficiency in practice, we study the empirical performance of \osis under different design choices, such as choice of MCMC kernel, empirical divergence, and other techniques:
\begin{enumerate}[cont]
    \item We study design choices for \osis (\cref{sec:practical-considerations}) and their influence on performance (\cref{tab:kernel-comparison}).
\end{enumerate}
Finally, we benchmark our sampler on amortised inference problems ranging from synthetic targets to posterior sampling under large-scale image model priors, leading to our final contribution:
\begin{enumerate}[cont]
    \item We show that \osis performs on par with multi-step baselines, while being more efficient (\cref{sub:sampling_benchmark}) and remaining scalable (\cref{subsec:outsourced}).
\end{enumerate}

\section{
    Proposed method: 
    Training by distilling MCMC refinements
}\label{sec:algorithm}

\subsection{Preliminaries}\label{subsec:osis}

\begin{wrapfigure}[13]{r}{0.45\textwidth}
\vspace*{-2.5em}
\centering
\begin{minipage}[t]{\linewidth}
    \begin{algorithm}[H]
    \caption{OSIS using a divergence $\mathcal{D}$}\label{alg:osis-div}
    \begin{algorithmic}[1]
    \Require Prior distribution $q$,
        $\pi$-invariant kernel $K$,
        generator $G_\theta$,
        divergence $\mathcal{D}$.
    \While{not converged}
        \State $\rho_\theta \leftarrow G_{\theta\,\#}q$
        \LineComment{Apply the MCMC kernel}
        \State $\rho_h \leftarrow K\rho_{\sg{\theta}}$
        \LineComment{Compute loss using $\mathcal{D}$}
        \State $\mathcal{L}(\theta)\leftarrow
               \mathcal{D}\left(\rho_\theta , \rho_h\right)$

        \State Gradient update on $\theta$ with $\nabla_\theta\mathcal{L}$
    \EndWhile
    \State \Return $G_\theta$
    \end{algorithmic}\end{algorithm}
\end{minipage} 
\end{wrapfigure}

\paragraph{Problem setting: one-step amortised sampling.}

Suppose that $G:\R^{d_z} \to \R^d$ is a measurable function and $q$ is a simple distribution (\eg, $\mathcal{N}(0,I)$)
over $\R^{d_z}$, where $d_z$ is not necessarily equal to $d$. Define a random variable $z\sim q$ and 
$x=G(z)$. The distribution of $x$ is the pushforward of $q$ by $G$, denoted 
$\rho = G_{\#}q$. If $G$ is not (efficiently) invertible or differentiable, the density of $\rho$ is intractable to compute.
Given a target distribution $\pi$ over $\R^d$ defined by an unnormalised density as in \eqref{eq:eq1}, the goal of one-step amortised sampling is to find $G$ such that 
$\pi \approx G_{\#}q$. 
If the function $G$ depends on parameters $\theta$, we
denote it $G_\theta$ and abbreviate $\rho_\theta = (G_\theta)_\#q$. The goal is then to find $\theta$ such that $\rho_\theta\approx\pi$.

\paragraph{Stationary kernels.}

Suppose that $K$ is a stochastic kernel from $\R^d$ to $\R^d$ (\ie, a conditional distribution $K(y\mid x)$ satisfying measurability conditions). One can construct a Markov chain $(x_k)_{k=0}^\infty$ via $x_{k + 1}\sim K(x_{k + 1}\mid x_k)$.
Markov chain Monte Carlo algorithms select $K$ such that, regardless of the distribution of $x_0$, the chain converges to a unique stationary distribution $\pi$ with respect to which $K$ is invariant, \ie, $K\pi(\dd y)\coloneqq\int K(\dd y\mid x)\pi(x)\,\dd x=\pi(\dd y)$  \citep{douc2018markov}. Many choices for $K$ use a discrete-time approximation to Langevin dynamics, such as ULA,  MALA \citep{rossky1978brownian, besag1994mala, roberts1996exponential}, HMC \citep{duane1987hybrid,neal2011mcmc}, and improved Langevin integrators \citep{leimkuhler2015molecular}.

\subsection{Training objectives for one-step sampling}%
Training of a one-step sampler, by some algorithm, produces a sequence of parameter iterates $(\theta_k)_{k\geq0}$, and thus an evolution of the modelled distribution $\rho_{\theta_k} \coloneqq  (G_{\theta_k})_\# q$. The training algorithm should be designed to ensure that $\rho_{\theta_k}$ converges to $\pi$ as $k\to\infty$. One way to express such convergence is through invariance of the limiting distribution with respect to a $\pi$-invariant stochastic kernel $K$.

\paragraph{Divergence-based loss.}
Given a $\pi$-invariant kernel $K$, a one-step sampler $\rho_\theta=(G_{\theta})_\#q$ should satisfy $\rho_\theta\approx K\rho_\theta$ upon convergence. We propose to turn this into a training objective:
\begin{equation}\label{eq:div-objective}
    \mathcal{L}(\theta) = \mathcal{D}\left(\rho_\theta,\,K\rho_{\sg{\theta}}\right),
\end{equation}
where $\mathcal{D}$ is some measure of divergence between distributions, $K\rho(\dd y)\coloneqq \int K(\dd y\mid x)\rho(\dd x)$, and $\sg{\cdot}$ is the stop-gradient operator.  

The complete algorithm, given in \Cref{alg:osis-div}, leaves room for two design choices: the kernel $K$ and the divergence $\mathcal{D}$. Both should be efficient to 
compute, and the distance $\mathcal{D}$ should be differentiable with respect to (empirical approximations of) $\rho_\theta$. We discuss these choices in \cref{subsec:kernel-options} and \cref{subsec:div-options}, respectively. Additional aspects of training are discussed in \cref{sec:practical-considerations} and variants of \cref{alg:osis-div} in \cref{app:algorithm-additional}.

\section{Connections and related work}\label{sec:connections}

\paragraph{Connection to other implicit samplers.}

The connection between generative modelling and implicit models, those without a parametrisation of density, goes back to \citet{mohamed2016learning}. 
Subsequent works explored algorithms that use MCMC to train implicit samplers: \citet{feng2017learning} amortises Stein variational gradient descent \citep[SVGD;][]{liu2016stein}, while \citet{titsias2017learning} fits an
invertible map to sample from the distribution induced by a short MCMC chain.
Related approaches include \citet{hoffman2017learning},
\citet{ruiz2019contrastive}, \citet{gabrie2022adaptive}, and \citet{yi2022sliced}.

The closest precursor to \osis is \citet{li2017approximate}. That algorithm initialises an
MCMC chain from an implicit approximation network, applies a transition kernel, and updates the network with an adversarial loss matching the
initial samples to their few-step refinements. \osis differs from this line of work in three key ways. First, unlike \citet{li2017approximate}, \osis does not rely on an adversarial loss, but remains stable with a fixed choice of empirical divergence. Second, \osis is compatible with any MCMC kernel and any divergence between distributions, and we compare several choices of each (see
\Cref{tab:kernel-comparison} for kernels and \Cref{tab:main-results,tab:main-results} for divergences). Third, we link the convergence of these methods to the convergence of the underlying MCMC kernel and establish a connection to Wasserstein gradient flows that was not known before in the sampling context (see \Cref{sec:theory}).

\paragraph{Connection to drifting models.}
\looseness=-1
\osis can be linked to drifting models \citep{deng2026generative}, which define a local target for the evolution of a modelled distribution to guide it towards a target given empirically through data samples. Although \osis and drifting models solve different problems (approximating an empirical data distribution and an energy-based distribution, respectively), they share some similarities. In \citet{deng2026generative}, stationarity of an implicitly defined generative model is expressed through a drifting field, a vector field defining an evolution of particles under which the target distribution is stationary under asymptotic conditions. The field is defined similarly to the gradient of a kernel-based divergence and thus evolves model-generated samples towards the data distribution over the course of training. Unlike in diffusion and flow-based models, which approximate a time-dependent evolution, this field is a non-stationary training target, rather than part of the generation process.

More recent works propose other ways to define this drifting field, some of which can be identified with gradients of divergences between distributions, making the training process simulate a gradient flow \citep{he2026sinkhorn, gretton2026wasserstein, han2026one}. As we discuss below, \osis shares parts of this connection; however, drifting models and their successors \emph{approximate the vector field giving the gradient flow}, whereas \osis 
\emph{uses the vector field giving the gradient flow to define the local target for optimisation}.

\citet{cao2026one} propose several ways to adapt drifting models for sampling. Their method relies on either Monte Carlo simulation or second-order derivatives to approximate the gradient of the energy, both of which scale poorly with dimensionality. In contrast, \osis defines the drift implicitly by computing the divergence 
between two probability distributions.

\paragraph{Connection to computational Wasserstein gradient flows.}

A Wasserstein gradient flow (WGF) describes a continuous evolution $(\rho_t)_{t\geq0}$ on the space of absolutely continuous probability distributions with finite second moments $\mathcal{P}_2(\R^d)$. The flow moves a distribution $\rho$ along the gradients of a functional $\mathcal{J}:\mathcal{P}_2(\R^d)\to\R$. The flow admits two equivalent descriptions: as an evolution of particles $x_t \sim \rho_t$ \eqref{eq:particle-flow} and the corresponding continuity equation, the evolution of the density $\rho_t$ \eqref{eq:continuity-pde}:
\par\noindent
\begin{minipage}{0.48\textwidth}
\begin{equation}\label{eq:particle-flow}
    \frac{\dd}{\dd t}{x_t} = -\nabla \frac{\delta \mathcal{J}}{\delta \rho_t}(x_t),
\end{equation}
\end{minipage}%
\hfill
\begin{minipage}{0.48\textwidth}
\begin{equation}\label{eq:continuity-pde}
    \pd{t}{\rho_t} = \nabla \cdot \Big( \rho_t \, \nabla \frac{\delta \mathcal{J}}{\delta \rho_t} \Big),
\end{equation}
\end{minipage}
\par\noindent
where initial conditions are given by $x_0 \sim \rho_0$ and 
$\frac{\delta \mathcal{J}}{\delta \rho}:\R^d \to \R$ denotes the first variation of $\cal J$. 
The flow given in \eqref{eq:continuity-pde} admits a discrete-time approximation scheme by proximal steps in the Wasserstein geometry, known as the Jordan-Kinderlehrer-Otto (JKO) scheme \citep{jordan1998variational}:
\begin{equation}\label{eq:jko-scheme}
\rho_{k+1} = \underset{\rho}{\operatorname{argmin}} \Big\{
    2\eta\mathcal{J}(\rho) +\sqW{\rho}{\rho_k}
\Big\}
\end{equation}
that converges to the continuous-time flow as $\eta\to0$. 
Refer to  \citet{ambrosio2005gradient} for a rigorous treatment or to \citet{santambrogio2016euclidea} for a succinct introduction.

The scheme in \eqref{eq:jko-scheme} offers a tempting way to solve optimisation problems on the space of probability distributions, and recent years have seen multiple attempts to convert it into a scalable machine learning algorithm. \citet{mokrov2021large} proposes to learn WGF by parametrising  each step of the JKO scheme \eqref{eq:jko-scheme} via an input-convex neural network. An alternative approach was introduced in \citet{galashov2025deep}. Some more recent works \citep{chen2025regularized, tian2026sobolev} offer ways to compute WGF using different variations of MMD as the functional $\mathcal{J}$. Finally, \citet{gretton2026wasserstein} provides an interpretation 
of drifting (\Cref{sec:connections}) as WGF. \citet{he2026sinkhorn, han2026one} show that Sinkhorn divergence to the target distribution can be used as the functional and can achieve strong results on image generation. Notably, \emph{all methods that attempt to neurally approximate WGF to this day fundamentally rely on the availability of unbiased samples}.

\section{Theoretical analysis of \osis}\label{sec:theory}

In this section we analyse the convergence of \osis under idealised assumptions:
\begin{enumerate}[label=(\arabic*),left=0pt,nosep]
\item We take $K$ to be continuous Langevin diffusion and $\mathcal{D}$ to be $\frac12\Wp{2}^2$. In practice, this assumption is satisfied in the infinite-particle limit of empirical estimation of the divergence and with perfect numerical integration in the Monte Carlo updates.
\item \looseness=-1 We assume that optimisation of the modelled distribution is performed in Wasserstein space, rather than the parameter space ($\theta$). This can be achieved in practice by replacing the gradient step on $\Wp{2}^2$ with respect to $\theta$ by a proximal gradient step in the $\Wp{2}^2$ geometry and assuming the parametric family and optimisation algorithm are sufficiently expressive to perform this step exactly.
\end{enumerate}
Our analysis uses that the evolution of density given by Langevin dynamics is the WGF of the KL divergence functional; we show that \osis inherits the convergence properties of Langevin dynamics \citep[\eg,][]{durmus2017nonasymptotic,chewi2025analysis}.

\subsection{Setting}\label{subsec:theory-prelim}

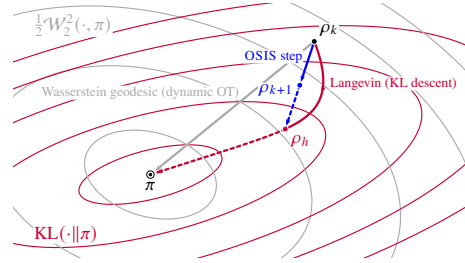
\begin{wrapfigure}[13]{r}{0.43\textwidth}
    \vspace*{-6em}
    \centering
    \resizebox{\linewidth}{!}{%
    % Figure 1 v4. All geometry and labels are editable native TikZ.
% Requires amsmath,tikz and arrows.meta,decorations.markings,calc.
\begingroup
\begin{tikzpicture}[x=1mm,y=1mm,line cap=round,line join=round,
 every node/.style={inner sep=0pt,outer sep=0pt,font=\fontsize{12}{12}\selectfont},
 mathlabel/.style={font=\fontsize{16}{16}\selectfont,fill=white,inner sep=.6pt,rounded corners=.3pt},
 operation/.style={fill=white,inner sep=.7pt,rounded corners=.3pt},
 onearrow/.style args={#1}{postaction={decorate},decoration={markings,mark=at position #1 with {\arrow{Stealth[length=6pt,width=4pt]}}}},
 state/.style={circle,inner sep=0pt,minimum size=1.8mm,draw=white,line width=.6pt}]
\definecolor{kernel}{HTML}{cc0033}
\definecolor{transport}{HTML}{aaaaaa}
\definecolor{osis}{HTML}{0000ff}

\definecolor{ink}{HTML}{000000}
\path[use as bounding box] (0,0) rectangle (139.7,78);
% Exact fixed-covariance Gaussian mean family, one common affine display.
\def\la{5}\def\lb{.7}\def\ma{1}\def\mb{2.8}
\def\kernelTime{.7}\def\stepsize{.5}\def\cx{42}\def\cy{25.5}
\def\saa{-18.181876736}\def\sab{24.493527408}
\def\sba{19.011800672}\def\sbb{7.838642624}
\pgfmathsetmacro{\fk}{.5*(\la*\ma^2+\lb*\mb^2)}
\pgfmathsetmacro{\gk}{.5*(\ma^2+\mb^2)}
\pgfmathdeclarefunction{px}{2}{\pgfmathparse{\cx+\saa*#1+\sab*#2}}
\pgfmathdeclarefunction{py}{2}{\pgfmathparse{\cy+\sba*#1+\sbb*#2}}
\pgfmathdeclarefunction{flowx}{1}{\pgfmathparse{px(\ma*exp(-\la*#1),\mb*exp(-\lb*#1))}}
\pgfmathdeclarefunction{flowy}{1}{\pgfmathparse{py(\ma*exp(-\la*#1),\mb*exp(-\lb*#1))}}
\coordinate (pi) at (\cx,\cy);
\coordinate (k) at ({flowx(0)},{flowy(0)});
\coordinate (h) at ({flowx(\kernelTime)},{flowy(\kernelTime)});
\coordinate (n) at ($(k)!\stepsize!(h)$);
% Selected level sets; the drawing window is not a functional domain.
% All three update states lie inside the two outer levels.
\begin{scope}
 \clip (0,0) rectangle (139.7,78);
 \foreach \ell in {.05,.30,.72,1.25,1.80}{
  \pgfmathsetmacro{\fa}{sqrt(2*\fk*\ell/\la)}
  \pgfmathsetmacro{\fb}{sqrt(2*\fk*\ell/\lb)}
  \pgfmathsetmacro{\gr}{sqrt(2*\gk*\ell)}
  \draw[kernel,line width=1pt]
   plot[domain=0:360,samples=241] ({px(\fa*cos(\x),\fb*sin(\x))},{py(\fa*cos(\x),\fb*sin(\x))}) -- cycle;
  \draw[transport,line width=1pt]
   plot[domain=0:360,samples=241] ({px(\gr*cos(\x),\gr*sin(\x))},{py(\gr*cos(\x),\gr*sin(\x))}) -- cycle;
 }
\end{scope}

% Direct optimal transport to the target, shown for comparison only.
\draw[transport,line width=2pt,shorten >=1.7mm,
 -{Stealth[length=6pt,width=4pt]},preaction={draw=white,line width=1.75pt}]
 (k)--(pi);
% Continue the SAME exact Langevin evolution towards its limiting law pi.
% Stop the visible terminal arrow just outside the target marker.
% The exact analytical continuation has limit pi as time tends to infinity.
\pgfmathsetmacro{\tailEnd}{ln(\mb*sqrt(\sab^2+\sbb^2)/1.9)/\lb}
\draw[kernel,line width=2pt,dash pattern=on 3pt off 3pt,
 -{Stealth[length=6pt,width=4pt]},
 preaction={draw=white,line width=1.9pt,solid}]
 plot[domain=\kernelTime:\tailEnd,samples=220] ({flowx(\x)},{flowy(\x)});
 
% Highlight the short Langevin curve and the current Euler step.
\draw[kernel,line width=2pt,preaction={draw=white,line width=2pt},onearrow=.50]
 plot[domain=0:\kernelTime,samples=200] ({flowx(\x)},{flowy(\x)});
 
\draw[osis,line width=2pt,shorten >=1mm,dash pattern=on 3pt off 3pt,
 -{Stealth[length=6pt,width=4pt]},preaction={draw=white,line width=2pt,solid}]
 (n)--(h);
\draw[osis,line width=2pt,shorten >=1mm,-{Stealth[length=6pt,width=4pt]}]
 (k)--(n);
 
% State labels: small opaque patches, kept local to the actual points.
\node[state,fill=ink] at (k) {};
\node[state,fill=osis] at (n) {};
\node[state,fill=kernel] at (h) {};
\fill[white] (pi) circle[radius=1.6];
\draw[ink,line width=.55pt] (pi) circle[radius=1.15];
\fill[ink] (pi) circle[radius=.55];

\node[mathlabel,anchor=north,text=ink] at ($(pi)+(0,-2.1)$) {$\pi$};
\node[mathlabel,anchor=south west,text=ink] at ($(k)+(1.5,1.6)$) {$\rho_k$};
\node[mathlabel,anchor=east,text=osis] at ($(n)+(-2.0,-1.3)$) {$\rho_{k+1}$};
\node[mathlabel,anchor=north west,text=kernel] at ($(h)+(1.5,-1.7)$) {$\rho_h$};

\node[operation,anchor=east,text=osis,font=\fontsize{12}{12}\selectfont]
 at (89,61.8) {\osis step};
 
\node[operation,anchor=east,text=transport,font=\fontsize{12}{12}\selectfont]
 at (69.5,50.7) {Wasserstein geodesic (dynamic OT)};
 
\node[operation,anchor=west,text=kernel] at (97,53.5) {Langevin (KL descent)};

\node[mathlabel,anchor=west,text=transport,font=\fontsize{16}{16}\selectfont]
 at (6,71.5) {$\frac12 \mathcal{W}_2^2(\cdot,\pi)$};
 
\node[mathlabel,anchor=west,text=kernel,font=\fontsize{16}{16}\selectfont]
 at (6,7.0) {$\mathrm{KL}(\cdot\Vert\pi)$};
\end{tikzpicture}%
\endgroup}\\[-0.75em]
    \caption{
        One idealised OSIS iteration. KL and Wasserstein distance share the minimiser $\pi$. However,
        KL decreases along the Langevin path (red), whereas Wasserstein decreases along the geodesic (grey).
        OSIS takes a  step along the Wasserstein gradient of $\frac12 \mathcal{W}_2^2(\cdot,\rho_h)$ 
        towards $\rho_{k+1}$ (blue).
    }
    \label{fig:osis-landscapes}
\end{wrapfigure}

We first formalise the algorithm described in \Cref{subsec:osis}. 

\paragraph{Notation and assumptions.} We assume that all distributions are elements of $\mathcal P_2(\R^d)$, the space of absolutely continuous distributions on $\R^d$ with finite second moment, and we identify them with their densities.
We assume that the target $\pi \propto e^{- \cal E }$ is defined by a twice continuously differentiable energy $\mathcal{E}:\R^d \to \R$.  Let $\{X_t\}_{t \geq 0}$ be a Langevin process, defined by an Itô SDE:
\begin{equation}\label{eq:main-exact-langevin}
  \dd X_t = - \nabla\mathcal{E}(X_t)\,\dd t + \sqrt{2}\,\dd B_t,
\end{equation}
where $\{B_t\}_{t\geq 0}$ is $d$-dimensional Brownian motion. For $h\geq0$, we define $K^h$ to be the stochastic integration kernel of the process \eqref{eq:main-exact-langevin}, \ie, $K^h(A\mid x) \coloneqq  \mathbb{P}(X_{t+h} \in A \mid X_t = x)$, which is $\pi$-invariant. 

For a distribution $\rho$ and an absolutely continuous kernel $K$ we denote by $K\rho$ the pushforward distribution, which has density $K\rho(y) \coloneqq  \int K (y\mid x)\rho(x)\dd x$. As before, we denote by $T_\#\rho$ the pushforward of a distribution $\rho$ by a measurable function $T$. Integration is assumed to be taken over the entire $\R^d$ unless stated otherwise.  We use $|\cdot|$ to denote Euclidean distance and define 2-Wasserstein distance and KL divergence between distributions $\rho$ and $\mu$ as follows:
\begin{equation*}
    \sqW{\rho}{\mu} = \inf_{T_\#\rho=\mu}\left\{
        \int |T(x) - x|^2\rho(x)\dd{x}
    \right\}, 
    \quad
    \KL{\rho}{\mu} = \int \log\frac{\rho(x)}{\mu(x)}\rho(x)\dd{x}.
\end{equation*}
We use the Monge formulation of 2-Wasserstein distance, since for absolutely continuous distributions, the Monge optimal transport maps exist and are unique. Other formulations are discussed in \Cref{app:additional-notation}.

\paragraph{One step of \osis in Wasserstein space.}
We now formalise one step of \osis, visualised in \Cref{fig:osis-landscapes}. At each training iteration $k \geq 0$, \osis has the initial distribution $\rho_k \coloneqq \rho_{\theta_k} = (G_{\theta_k})_\#q$.

\textbf{First}, we use $\rho_{h_k} \coloneqq  K^{h_k}\rho_k$ to denote the distribution after a continuous evolution of $\rho_k$ by a  stochastic kernel $K^{h_k}$ for the time $h_k > 0$. \textbf{Second}, we compute the 2-Wasserstein OT map $T^{h_k}$ between $\rho_k$  and $\rho_{h_k}$. With $\rho_{h_k}$ fixed, the functional
$\frac12\Wp{2}^2(\cdot,\rho_{h_k})$ has Wasserstein
gradient $\Id-T^{h_k}$ at $\rho_k$
\citep[Theorem~2.2]{alfonsi2020squared}.  \textbf{Third},  we set $T^{h_k}_{\eta_k} \coloneqq \eta_k T^{h_k} + (1-\eta_k)\Id$ with 
$\eta_k \in (0, 1]$, which is the result of taking a gradient step on $\frac12\Wp{2}^2(\cdot,\rho_{h_k})$ with step size $\eta_k$. Finally, one idealised iteration of \osis is expressed as
\begin{equation}\label{eq:one-iterate-osis}
    \rho_{k+1} \coloneqq (T^{h_k}_{\eta_k})_\#\rho_k.
\end{equation}

\subsection{Main results}

We prove convergence of the procedure \eqref{eq:one-iterate-osis} under various assumptions on $\pi$. 
Proofs appear in \Cref{subsec:proofs}.

\paragraph{Convergence for strongly log-concave target.}

\looseness=-1
We consider the case where $\pi$ is $\lambda$-strongly log-concave ($\lambda$-SLC), that is, $\nabla^2\mathcal{E}\succeq \lambda I$ for some $\lambda > 0$. 
\begin{lemmaE}[KL convexity along a Wasserstein geodesic][all end]
\label{lem:geodesic-kl-bound}
Let $\rho_0$, $\rho_1\in\mathcal{P}_2(\R^d)$, and let $\rho_t$ be the geodesic between $\rho_0$ and $\rho_1$ in $\mathcal{P}_2(\R^d)$. With $\pi(x)\propto e^{-\mathcal{E}(x)}$, $\KL{\rho_t}{\pi}$ is
\begin{itemize}[left=0pt]
    \item \textbf{geodesically convex} if $\pi$ is log-concave;
    \item \textbf{$\lambda$-strongly geodesically convex} if $\pi$ is $\lambda$-strongly log-concave.
\end{itemize}
Moreover, if  $\nabla^2\mathcal{E}\succeq -\kappa I$ for some $\kappa > 0$, that is, $\cal E$ is hypoconvex, then 
\begin{equation}
    \forall t\in[0, 1]
    \quad\mathcal{J}(\rho_t) \leq t\mathcal{J}(\rho_1) + (1 - t)\mathcal{J}(\rho_0) 
    + \frac{\kappa}{2} t(1-t)\sqW{\rho_0}{\rho_1}.
\end{equation}
\end{lemmaE}
\begin{proofE}
We prove the result for $\lambda$-strong convexity; convexity can be recovered by setting $\lambda=0$. Recall that $\rho_t=(T_t)_\#\rho_0$,  where $T_t(x) = tT + (1-t)\Id$ and $T$ is the Monge map from $\rho_0$ to $\rho_1$.  First, we rewrite $\KL{\rho_t}{\pi}$ as follows:
\begin{align}\label{eq:kl-representation}
    \KL{\rho_t}{\pi} 
    & = \int \rho_t(x)\log\frac{\rho_t(x)}{\pi(x)}\dd x
      = \int \rho_0(x)\log\frac{\rho_0(x)}{\pi(T_t(x)) \det(\nabla T_t) }\dd x  \\
    & = \int \mathcal{E}(T_t(x))\rho_0(x)\,\dd x
     - \int \log\det(\nabla T_t) \rho_0(x) \,\dd x  + \text{const}.
\end{align}
Next, we show the first term is $\lambda$-strongly convex and the second term is convex. Since $\pi$ is $\lambda$-strongly log-concave, for any $x$ we have
\begin{align}
    \mathcal{E}(T_t(x)) = \mathcal{E}(tT(x) + (1-t)x) 
    t \leq \mathcal{E}(T(x)) + (1 - t)  \mathcal{E}(x) - \frac{\lambda}{2}t(1-t)|T(x) - x|^2.
\end{align}
Integrating with $\rho_0(x)$, we obtain
\begin{equation}\label{eq:energy-int}
    \int \mathcal{E}(T_t(x)) \rho_0(x) \dd x \leq  \int t \mathcal{E}(T(x)) \rho_0(x) \dd x 
    + (1 - t) \int \mathcal{E}(x) \rho_0(x) \dd x - \frac{\lambda}{2}t(1-t)\sqW{\rho_0}{\rho_1},
\end{equation}
showing $\lambda$-strong convexity of the first term. For the second term, we have
\begin{equation}
    -\log\det(\nabla T_t) = -\log\det(t \nabla T(x) + (1 - t)\Id) = - \sum_i\log((1 - t) + t \sigma_i) \leq -t\log\det(\nabla T),
\end{equation}
where $\sigma_i$ are the eigenvalues of $\nabla T$. Because $T$ is the gradient of a convex function \citep{santambrogio2015optimal}
$t\log\det(\nabla T)>0$, integrating over $\rho_0$, we get 
\begin{equation}
    -\int\log\det(\nabla T_t)\rho_0(x)\,\dd x\leq -t\int\log\det(\nabla T)\rho_0(x)\,\dd x,
\end{equation}
which is exactly convexity of the second summand, noting that $T_0=\Id$ (so $\log\det(\nabla T_0)=0$) and $T_1=T$.

This proves $\lambda$-strong convexity of KL along $\rho_t$. The case of geodesic convexity can be recovered by setting $\lambda=0$.

Finally, we prove the inequality for hypoconvex $\mathcal{E}$, using the KL representation from 
\eqref{eq:kl-representation}. It suffices to consider the only the first term. Observe that since
$\mathcal{E} \succeq - \kappa I$, then for any $x$ the function
$\mathcal{E}\left(y\right) + \frac{\kappa}{2}|y - x|^2 $ is convex in $y$. Using $y = tT(x) + (1- t)x$ and setting
\begin{equation*}
    \phi(t) \coloneqq  \mathcal{E}\left( tT(x) + (1-t)x \right) + \frac{\kappa}{2}t^2|T(x) - x|^2,
\end{equation*} 
we obtain $\phi(t) \leq t \phi(1) + (1-t)\phi(0) \,\forall t \in [0, 1]$. Rearranging,
\begin{equation*}
    \mathcal{E}\left( tT(x) + (1-t)x \right) \leq
    t \mathcal{E}\left( T(x) \right)
    (1 - t) \mathcal{E}\left( x \right)
    + \frac{\kappa}{2}t(1-t)|T(x) - x|^2,
\end{equation*} 
By integrating with $\rho_0(x)$, we obtain an equivalent of \eqref{eq:energy-int}, which concludes the proof. 
\end{proofE}
\begin{lemmaE}[][all end]
\label{appxlemma:sqw-kl-bound}
Let $\rho_0\in\mathcal P_2(\R^d)$ satisfy $\KL{\rho_0}{\pi}<\infty$, and define $\rho_h=K^h\rho_0$,
Then, for any $h>0$,
\begin{equation}\label{eq:app-entropy-action-bound}
    \Wp{2}^2(\rho_0,\rho_h) \leq h\KL{\rho_0}{\pi}.
\end{equation}
\end{lemmaE}
\begin{proofE}
    We begin the proof by recalling the Benamou-Brenier formula for $\W{\rho_0}{\rho_h}$ 
    (see \eqref{eq:benamou-brenier}). For the velocity $v_t = \nabla\log\frac{\pi}{\rho_t}$, which is associated with the Langevin diffusion, we have
    \begin{align}
        \sqW{\rho_0}{\rho_h} 
        & \leq h \int_0^h \int |v_t|^2\rho_t(x)\dd x\dd t 
          = h^2 \int_0^1 \int |v_{hs}|^2\rho_{hs}(x)\dd x\dd s \\
          = h^2 \int_0^1 &\FI{\rho_{hs}}{\pi}ds 
          = h(\KL{\rho_0}{\pi} -  \KL{\rho_h}{\pi}) \leq h\KL{\rho_0}{\pi},
    \end{align}
    where the first inequality follows from the Langevin diffusion path having a greater dynamic transport cost than the geodesic, the second equality is due to rescaling $s = t / h$, and the last equality is due to the fact that $\frac{\dd}{\dd t} \KL{\rho_t}{\pi} = - \FI{\rho_t}{\pi}$, 
    \citep[][\S2.2]{gretton2026wasserstein}, 
    \citep[][Lemma 2]{vempala2019rapid}.
\end{proofE}
\begin{theoremE}[Contraction of one step of \osis under $\lambda$-SLC $\pi$][end, restate, text link=]
\label{thm:main-slc-contraction}
Let $\pi$ be $\lambda$-SLC. Let $\rho_k,\rho_{h_k},T^{h_k},T_{\eta_k}^{h_k}$, $\rho_{k+1}$ be as in \Cref{subsec:theory-prelim}. For any $h_k>0$, $\eta_k\in(0,1]$, the following bounds hold:
\begin{align}
    \W{\rho_{k + 1}}{\pi} & \leq \left((1 - \eta_k)(1 + e^{-\lambda h_k}) + e^{-\lambda h_k}\right) \W{\rho_k}{\pi}; 
    \label{eq:one-step-w2-bound} \\
    \KL{\rho_{k + 1}}{\pi}  & \leq (1 - \eta_k(1 - e^{-2 \lambda h_k})) \KL{\rho_k}{\pi}.
    \label{eq:one-step-kl-bound} 
\end{align}
\end{theoremE}
\begin{proofE}
\label{proofs:lambda-slc-step}
We start by recalling the Langevin convergence bound for $\lambda$-SLC distributions 
\citep[][Theorem 1.1(iii)]{vonrenesse2005transport}:
\begin{equation}
    \W{\rho_{h_k}}{\pi} \leq e^{-\lambda h_k}\W{\rho_k}{\pi},
\end{equation}
which, together with the triangular inequality, provides the following bound for $\W{\rho_{k+1}}{\pi}$:
\begin{equation*}
    \W{\rho_{k +1}}{\pi} \leq  \W{\rho_{k+1}}{\rho_{h_k}} + e^{-\lambda h_k}\W{\rho_k}{\pi}.
\end{equation*}
Next, we introduce the bound for $W_2(\rho_{k+1},\rho_{h_k})$
\begin{align*}
    \W{\rho_{k+1}}{\rho_{h_k}} 
        & = (1 - \eta_k) \W{\rho_k}{\rho_{h_k}} 
        \leq (1 - \eta_k) \Big(\W{\rho_{k}}{\pi} +  \W{\rho_{h_k}}{\pi}\Big) \\
        &\leq (1 - \eta_k) \Big(\W{\rho_{k}}{\pi} +  e^{-\lambda h_k} \W{\rho_{k}}{\pi}\Big)
        = (1 - \eta_k) (1 + e^{-\lambda h_k})  \W{\rho_{k}}{\pi}.
\end{align*}
This gives us the combined bound: 
\begin{equation*}
    \W{\rho_{k + 1}}{\pi} \leq 
    \underbrace{\Big(
        \underbrace{(1 - \eta_k) (1 + e^{-\lambda h_k})}_{\substack{
            \text{penalty for the
            } \\ \text{gradient descent}
        }} 
        + 
        \underbrace{e^{-\lambda h_k}}_{\substack{
            \text{convergence}\\\text{of exact Langevin}
        }}\Big)}_{
           = C(h_k, \eta_k)
        } \W{\rho_{k}}{\pi}.
\end{equation*}
Now we establish conditions for $C(h_k, \eta_k)<1$:
\begin{align*}
    & C(h_k, \eta_k) = 1 + 2e^{-\lambda h_k} - \eta_k(1 + e^{-\lambda h_k}) < 1, \\
    & 2e^{-\lambda h_k} < \eta_k(1 + e^{-\lambda h_k}), \quad 2/ (1 + e^{\lambda h_k}) < \eta_k.
\end{align*}
 So we get $\eta_k > 2  \sigma(-\lambda h_k)$, where $\sigma(s)$ is the sigmoid function.

We now prove the bound on KL divergence under the same assumptions.
We use the strong convexity of KL along geodesics (\Cref{lem:geodesic-kl-bound}) to obtain the following inequality:
\begin{equation}
  \KL{\rho_{k+1}}{\pi}
  \leq (1 - \eta_k)\KL{\rho_k}{\pi} + \eta_k\KL{\rho_{h_k}}{\pi}
  - \frac{\lambda}{2}\eta_k(1 - \eta_k)\Wp{2}^2(\rho_k,\rho_{h_k}).
  \label{appxeq:slc-displacement-convexity}
\end{equation}

We recall that $\rho_{h_k} = K^{h_k}\rho_k$. 
If $\rho_t$ is a continuous curve in $\mathcal{P}_2(\R^d)$ induced by Langevin diffusion,
using the Bakry-\'{E}mery criterion \citep{bakry1985diffusions}, together with the fact that 
$\frac{\dd}{\dd t} \KL{\rho_t}{\pi} = - \FI{\rho_t}{\pi}$, \citep[][\S2.2]{gretton2026wasserstein}, 
\citet[][Lemma 2]{vempala2019rapid}
gives
\begin{equation}
  \frac{\dd}{\dd t}\KL{\rho_t}{\pi}
  = -\FI{\rho_t}{\pi}  \leq -2\lambda\KL{\rho_t}{\pi},
\end{equation}
which consequently implies geometric convergence for KL:
\begin{align}
    \KL{\rho_{h_k}}{\pi} \leq
    e^{-2\lambda h_k}\KL{\rho_k}{\pi}.
\label{appxeq:slc-kl-endpoint}
\end{align}
Finally, combining \Cref{appxeq:slc-kl-endpoint} with \eqref{appxeq:slc-displacement-convexity} gives:
\begin{align}
    \KL{\rho_{k+1}}{\pi}
    & \leq (1 - \eta_k)\KL{\rho_k}{\pi} + \eta_k\KL{\rho_{h_k}}{\pi}
    - \frac{\lambda}{2}\eta_k(1 - \eta_k)\Wp{2}^2(\rho_k,\rho_{h_k}) \nonumber \\ 
    & \leq (1 - \eta_k(1 - e^{-2\lambda h_k}))\KL{\rho_k}{\pi}
    - \frac{\lambda}{2}\eta_k(1 - \eta_k)\Wp{2}^2(\rho_k,\rho_{h_k}) \nonumber  \\
    & \leq \underbrace{\left(1 - \eta_k(1 - e^{-2 \lambda h_k})\right)}_{C(\eta_k,h_k)}\KL{\rho_k}{\pi},
\end{align}
which proves the final bound for KL with 
$C(\eta_k, h_k) = 1 - \eta_k(1 - e^{-2\lambda h_k})$.
\end{proofE}
\begin{theoremE}[Convergence of \osis under $\lambda$-SLC $\pi$][end, restate, text link=]
\label{thm:main-slc-convergence}
Under the assumptions of \Cref{thm:main-slc-contraction}, let $(\rho_k)_{k\geq 0}$ be the sequence of probability distributions produced by the \osis iterations. 
Assuming $h_k>0$ and $\eta_k\in(0, 1]$ for all $k \geq 0$, the following bounds hold for 
any $N \geq 1$:
\begin{equation}\label{appxeq:main-slc-iterated-kl}
    \KL{\rho_N}{\pi} \leq e^{-\sum_{k=0}^{N-1}\eta_k(1-e^{-2 \lambda h_k})} \KL{\rho_0}{\pi}.
\end{equation}
Moreover, if $\bar{h}=\inf_{k\geq0} h_k > 0$, $\bar{\eta} = \inf_{k\geq0} \eta_k> 0$, and $\bar{\eta} > 2/(1+\exp(\lambda\bar{h}))$, then:
\begin{equation}\label{appxeq:main-slc-iterated-w2}
    \Wp{2}(\rho_{N},\pi) \leq  \big( (1 - \bar{\eta})(1 + e^{-\lambda \bar{h}}) + e^{-\lambda \bar{h}} \big)^N\Wp{2}(\rho_0,\pi);
\end{equation}
\end{theoremE}

\begin{proofE}
By iteratively applying the bound from \Cref{thm:main-slc-contraction} $N$ times and using $0<1 - x < e^{-x}$ for $x<1$, we get:
\begin{equation}
    \KL{\rho_N}{\pi}
    \leq 
    \prod_{k=0}^{N-1} \left(1 - \eta_k(1 - e^{-2\lambda h_k}) \right) \KL{\rho_0}{\pi}
    \leq 
    \exp\Bigl(-\sum_{k=0}^{N-1}\eta_k(1-e^{-2 \lambda h_k})\Bigr)\KL(\rho_0,\pi) \nonumber. \\
\end{equation}
Hence $\KL{\rho_k}{\pi}\to0$ whenever 
$\underset{N\to\infty}{\lim}\sum^{N - 1}_{k = 0}\eta_k(1-e^{-2\lambda h_k})\to\infty$. 

To prove the bound for $\Wp{2}$, we observe that $\forall k \geq0 \quad C(\bar{\eta}, \bar{h}) < C(\eta_k, h_k)$. By recursively applying the bound from \eqref{eq:one-step-w2-bound} we get:
\begin{equation}
    \Wp{2}(\rho_{N},\pi) \leq  \left( (1 - \bar{\eta})(1 + e^{-\lambda \bar{h}}) + e^{-\lambda \bar{h}} \right)^N\Wp{2}(\rho_0,\pi).
\end{equation}
\end{proofE}
As a result, we have shown a geometric convergence rate for \osis under a $\lambda$-SLC assumption on $\pi$.

\paragraph{Beyond strong log-concavity.}
The same update also converges for non-log-concave
$\pi$ satisfying an $\alpha$-log-Sobolev inequality ($\alpha$-LSI)
and a lower curvature bound
$\nabla^2\mathcal E\succeq-\kappa I$.
Here $\alpha>0$ and $\alpha$-LSI means
$2\alpha\KL{\rho}{\pi}
\le\int|\nabla\log(\rho/\pi)|^2\rho\,\dd x$ for all $\rho$.
In the appendix (\cref{thm:app-lsi-convergence}), we prove convergence rates, which in particular imply:

\begin{theoremE}[Contraction of one step of \osis under $\pi$ 
    satisfying logarithmic Sobolev inequality][all end]
\label{prop:app-semiconvex-entropy-descent}
Suppose that  $\mathcal E \in C^2(\R^d)$ and $\nabla^2(\mathcal E) \succeq - \kappa I$ for $\kappa \geq 0$ and that $\pi\propto e^{-\mathcal{E}}$ satisfies the logarithmic Sobolev inequality with parameter $\alpha > 0$ ($\alpha$-LSI), that is:
\begin{equation}
\forall \rho \in \mathcal P(\R^d)  \qquad 2\alpha\KL{\rho}{\pi}\le\FI{\rho}{\pi}.
\label{eq:app-lsi-convention}
\end{equation}
With the same notations as in \cref{thm:main-slc-contraction}, the following bound holds:
\begin{equation}
\KL{\rho_{k+1}}{\pi}
\le
\KL{\rho_k}{\pi} 
-
 \eta_k
\left[
1-\frac{ \kappa h_k}{2}(1- \eta_k)
\right]
\left[
\KL{\rho_k}{\pi}
-\KL{\rho_{h_k}}{\pi}
\right].
\label{eq:app-semiconvex-entropy-descent}
\end{equation}
Further, for $h>0$, $\eta \in (0,1]$, define
\begin{align}
     C_{\alpha, \kappa}( \eta,h)\coloneqq 1-\eta \left[1-\frac{\kappa h}{2}(1-\eta)\right](1-e^{-2\alpha h}).
     \label{eq:C-alpha-LSI}
\end{align}
If $\kappa h_k(1- \eta_k)\leq 2$, then
\begin{equation}
\KL{\rho_{k+1}}{\pi}
\le C_{\alpha, \kappa}( \eta_k,h_k)\KL{\rho_k}{\pi}.
\end{equation}
\end{theoremE}
\begin{proofE}
Apply \cref{lem:geodesic-kl-bound} to the
Wasserstein geodesic from $\rho_k$ to
$\rho_{h_k}=K^{h_k}\rho_k$, with curvature parameter
$- \kappa$ and interpolation parameter $ \eta_k$.
This gives
\begin{align}
\KL{\rho_{k+1}}{\pi}
&\le
(1- \eta_k)\KL{\rho_k}{\pi}
+ \eta_k\KL{\rho_{h_k}}{\pi} \nonumber\\
&\quad+
\frac{ \kappa}{2} \eta_k(1- \eta_k)
\Wp{2}^2(\rho_k,\rho_{h_k}).
\label{eq:app-semiconvex-geodesic-step}
\end{align}
The last term is nonnegative because $\kappa \geq 0$ and $\eta_k \in (0,1]$.

Rearranging the first two entropy terms gives
\begin{align}
\KL{\rho_{k+1}}{\pi}
&\le
\KL{\rho_k}{\pi}
- \eta_k
\left[
\KL{\rho_k}{\pi}
-\KL{\rho_{h_k}}{\pi}
\right] \nonumber\\
&\quad+
\frac{ \kappa}{2} \eta_k(1- \eta_k)
\Wp{2}^2(\rho_k,\rho_{h_k}).
\label{eq:app-semiconvex-before-action}
\end{align}

Using the proof of \cref{appxlemma:sqw-kl-bound},
\begin{align*}
\Wp{2}^2(\rho_k,\rho_{h_k})
\le
h_k
\left[
\KL{\rho_k}{\pi}
-\KL{\rho_{h_k}}{\pi}
\right].
\end{align*}
The multiplier
$\frac{ \kappa}{2} \eta_k(1- \eta_k)$ is
nonnegative, so substituting this upper bound into
\eqref{eq:app-semiconvex-before-action} we obtain
\begin{align}
\KL{\rho_{k+1}}{\pi}
&\le
\KL{\rho_k}{\pi} \nonumber\\
&\quad-
 \eta_k
\left[
1-\frac{ \kappa h_k}{2}(1- \eta_k)
\right]
\left[
\KL{\rho_k}{\pi}
-\KL{\rho_{h_k}}{\pi}
\right].
\label{eq:relative-entropy-decrease-LSI}
\end{align}
Exact Langevin performs descent on relative entropy, so
$
\KL{\rho_k}{\pi}
-\KL{\rho_{h_k}}{\pi}\ge0.
$
We conclude that, if $\kappa h_k(1- \eta_k)\le2$ then
\begin{align*}
1-\frac{ \kappa h_k}{2}(1- \eta_k)\ge0,
\end{align*}
and so the subtracted term is nonnegative, proving entropy
monotonicity.

For almost every $s>0$, the entropy dissipation identity and $\alpha$-LSI bound give
\begin{equation}
\frac{\dd}{\dd s}\KL{K^s \rho_k}{\pi}
=-\FI{K^s \rho_k}{\pi}
\le-2\alpha\KL{K^s \rho_k}{\pi}.
\end{equation}

Integrating as in the proof of \cref{appxlemma:sqw-kl-bound},

\begin{equation}
\KL{\rho_{h_k}}{\pi}\le e^{-2\alpha h_k}\KL{\rho_k}{\pi}.
\label{eq:app-lsi-endpoint}
\end{equation}

We substitute this upper-bound in \eqref{eq:relative-entropy-decrease-LSI} noting that the step condition makes $ \eta_k[1- \kappa h_k(1- \eta_k)/2]$ nonnegative and obtain

\begin{equation}
\KL{\rho_{k+1}}{\pi}
\le C_{\alpha, \kappa}( \eta_k,h_k)\KL{\rho_k}{\pi}.
\end{equation}
\end{proofE}
\begin{theoremE}[Convergence under LSI and bounded negative curvature][all end]
\label{thm:app-lsi-convergence}
Suppose that  $\mathcal E \in C^2(\R^d)$ and $\nabla^2(\mathcal E) \succeq - \kappa I$ for $\kappa \geq 0$ and that $\pi\propto e^{-\mathcal{E}}$ satisfies $\alpha$-LSI.
With the same notations as in \cref{thm:main-slc-contraction}, let $(\rho_k)_{k\geq0}$ be the sequence of probability distributions produced by \osis interations, define
\begin{equation}
C_{\alpha, \kappa}(\eta,h)
=1-\eta\left[1-\frac{ \kappa h}{2}(1-\eta)\right](1-e^{-2\alpha h}).
\label{eq:app-lsi-factor}
\end{equation}
Assuming $h_k>0$, $\eta_k\in(0, 1]$ for all $k\geq0$,
 the following bounds hold for 
any $N \geq 1$:
\begin{align}
\KL{\rho_N}{\pi}
&\le\left(\prod_{k=0}^{N-1}C_{\alpha, \kappa}( \eta_k,h_k)\right)\KL{\rho_0}{\pi}\\
&\le\exp\!\left(-\sum_{k=0}^{N-1}[1-C_{\alpha, \kappa}( \eta_k,h_k)]\right)\KL{\rho_0}{\pi},\label{eq:app-lsi-exponential-bound}\\
\Wp{2}^2( \rho_N,\pi)&\le\frac2\alpha\KL{\rho_N}{\pi}.
\label{eq:app-lsi-iteration}
\end{align}
\end{theoremE}
\begin{proofE}

The proof uses exact-Langevin KL decay under LSI \citep[Theorem~4]{vempala2019rapid} and Talagrand inequality \citep[Theorem~1]{otto2000generalization}, in addition to \cref{prop:app-semiconvex-entropy-descent}.

By \cref{prop:app-semiconvex-entropy-descent}, the one-step bound is
\begin{equation}
\KL{\rho_{k+1}}{\pi} \leq C_{\alpha, \kappa}(\eta_k,h_k)\KL{\rho_k}{\pi}.
\end{equation}

We note that $\kappa\ge0$ and $\kappa h_k(1-\eta_k)\le2$ imply $\eta_k[1- \kappa h_k(1-\eta_k)/2]\in [0,1]$ and $0<e^{-2\alpha h_k}\le C_{\alpha, \kappa}(\eta_k,h_k)\le1$. The upper inequality is strict if $\eta_k>0$ and $\kappa h_k(1-\eta_k)<2$.

Iterating the one-step descent over $k$ gives the product bound. Applying the inequality $C_{\alpha,\kappa}\le e^{-(1-C_{\alpha,\kappa})}$ to each factor gives the exponential bound in \eqref{eq:app-lsi-exponential-bound}.

We now apply Talagrand's inequality
\begin{equation}
\Wp{2}^2( \rho_k,\pi)
\le 2\KL{\rho_k}{\pi}/\alpha.
\end{equation}
Thus the previous bound provides convergence of both KL and $\Wp{2}$.
\end{proofE}

\begin{corollaryE}[Geometric convergence under LSI and bounded negative curvature][end,restate,text link=]
\label{cor:app-lsi-convergence}
Under the notations of \cref{subsec:theory-prelim}, and the $\alpha$-LSI and curvature assumptions above, with fixed $h_k=h>0$ and $\eta_k=\eta\in(0,1]$ with $ \kappa h(1-\eta)<2$, $\sqW{\rho_k}{\pi}$ and $\KL{\rho_k}{\pi}$ converge geometrically to 0.
\end{corollaryE}
\begin{proofE}
    Immediate from \Cref{thm:app-lsi-convergence}, as both discrepancies tend to zero if $\sum_{k\ge0}[1-C_{\alpha, \kappa}( \eta_k,h_k)]=\infty$.
\end{proofE}

\section{Practical considerations}\label{sec:practical-considerations}

The idealised setting, which is detailed in \Cref{alg:osis-div}  and theoretically analysed in \Cref{sec:theory} cannot be implemented in practice. Instead, we approximate the distributions and the divergence between them 8by their empirical counterparts and replace the continuous-time Langevin by a discrete-time MCMC simulation. The next paragraphs detail various design choices for \osis relevant in practice. An extended introduction to the practical considerations of the algorithm is provided in \cref{apx:ext-practical-considerations}.

\paragraph{Refinement kernel.} \osis can be instantiated with different MCMC kernels. The theory concerns exact Langevin, but we use the MALA kernel in practice and consider other choices in \cref{tab:kernel-comparison}.

\paragraph{Empirical divergence.} It is not possible to use infinite populations as considered in the theory, instead \osis computes finite-batch empirical divergences between generated samples and their kernel-refined versions. Our presented theory considers a squared 2-Wasserstein loss, but \osis can be extended to other discrepancies between distributions. For the experiments on the synthetic targets (\Cref{sub:sampling_benchmark}) we use sliced Wasserstein distances \citep[SWD;][]{bonneel2015sliced}, and for the experiments on Bayesian posteriors (\Cref{subsec:outsourced}) we use minibatch OT with squared-Euclidean cost, which converges to exact 2-Wasserstein in the limit \citep{fatras2021minibatch}.

\paragraph{Annealing.} Kernel refinements can initially target the tempered distribution $\pi_{\beta}\propto\exp(-\beta\mathcal E)$ with $\beta_t$ and then anneal $\beta$ to 1 over the course of training. The starting temperature and annealing schedule balance early exploration with learning the final target.

\paragraph{Complete algorithm.} To facilitate training stability, we make several gradient steps after each MCMC refinement. The complete training procedure detailing the use of empirical divergence, finite-step MCMC kernel, annealing, and multiple gradient steps per iteration is presented in \Cref{alg:osis-div-ful}.

\section{Experiments}\label{sec:experiments}
\subsection{Synthetic benchmarks}
\label{sub:sampling_benchmark}

\paragraph{Target distributions.} We evaluate \osis on standard sampling benchmarks: Neal's funnel ($d=10$) \citep{neal2000slice}, ManyWell ($d=32, 64$) \citep{noe2019boltzmann}, and 40GMM ($d=2, 10$) \citep{blessing2024beyond}. We also consider benchmarks in higher dimension: NICE-MNIST ($d=196$) and NICE-FMNIST ($d=728$) \citep{blessing2024beyond}.

\begin{table*}[t]
\centering
\caption{Sampler comparison on GMM, Funnel, and ManyWell targets. Entries are mean $\pm$ standard deviation over five seeds. Second column shows number of NN and energy evaluations per sample}\vspace*{-1em}
\label{tab:main-results}
\resizebox{\linewidth}{!}{%
\begin{tabular}{@{}lr*{10}{r}}
\toprule
Target density $\rightarrow$ &
 & \multicolumn{2}{c}{40GMM ($d=2$)}
 & \multicolumn{2}{c}{40GMM ($d=10$)}
 & \multicolumn{2}{c}{Funnel ($d=10$)}
 & \multicolumn{2}{c}{ManyWell ($d=32$)}
 & \multicolumn{2}{c}{ManyWell ($d=64$)} \\
\cmidrule(lr){3-4} \cmidrule(lr){5-6} \cmidrule(lr){7-8} \cmidrule(lr){9-10} \cmidrule(lr){11-12}
Method $\downarrow$ Metric $\rightarrow$\hspace*{-1em}
 & NN / $\mathcal E$
 & \multicolumn{1}{c}{$\mathcal{W}_2^2$} & \multicolumn{1}{c}{MMD ($\times10^{-3}$)}
 & \multicolumn{1}{c}{$\mathcal{W}_2^2$} & \multicolumn{1}{c}{MMD ($\times10^{-3}$)}
 & \multicolumn{1}{c}{$\mathcal{W}_2^2$} & \multicolumn{1}{c}{MMD ($\times10^{-3}$)}
 & \multicolumn{1}{c}{$\mathcal{W}_2^2$} & \multicolumn{1}{c}{MMD ($\times10^{-3}$)}
 & \multicolumn{1}{c}{$\mathcal{W}_2^2$} & \multicolumn{1}{c}{MMD ($\times10^{-3}$)} \\
\midrule
Reference
 & 
 & $7.68\std{3.22}$ & $7.17\std{1.51}$
 & $202.81\std{16.77}$ & $7.98\std{0.78}$
 & $87.60\std{1.20}$ & $9.22\std{1.54}$
 & $23.61\std{0.07}$ & $8.75\std{0.72}$
 & $78.36\std{0.11}$ & $9.58\std{0.65}$ \\
\midrule
\rowcolor{osisgreen} \osis
 & $1$ / $0$
 & $103.26\std{41.41}$ & $64.37\std{13.38}$
 & $3086.96\std{439.59}$ & $139.26\std{36.09}$
 & $93.99\std{3.33}$ & $28.06\std{3.06}$
 & $26.79\std{0.41}$ & $55.92\std{6.00}$
 & $75.88\std{0.34}$ & $91.05\std{3.98}$ \\
\midrule
FAB (flow only)
 & $1$ / $0$
 & $296.01\std{333.93}$ & $81.05\std{38.73}$
 & $6217.94\std{472.86}$ & $382.64\std{44.65}$
 & $583.48\std{233.71}$ & $12.94\std{2.34}$
 & $27.10\std{0.46}$ & $47.95\std{4.56}$
 & $125.58\std{0.04}$ & $401.28\std{0.64}$ \\
FAB
 & $1$ / $120$
 & $72.03\std{19.29}$ & $46.42\std{6.32}$
 & $6257.04\std{455.10}$ & $380.98\std{45.19}$
 & $582.97\std{233.17}$ & $11.87\std{1.74}$
 & $25.52\std{0.30}$ & $36.63\std{5.56}$
 & $132.55\std{0.48}$ & $299.11\std{1.51}$ \\
SMC
 & $0$ / $3825$
 & $14.67\std{1.79}$ & $16.85\std{2.11}$
 & $4362.81\std{139.06}$ & $208.62\std{11.73}$
 & $149.56\std{1.37}$ & $125.72\std{8.99}$
 & $26.04\std{0.16}$ & $22.48\std{1.10}$
 & $85.76\std{0.29}$ & $42.63\std{1.36}$ \\
DIS
 & $128$ / $128$
 & $29.26\std{7.00}$ & $34.71\std{10.99}$
 & $1650.04\std{60.09}$ & $85.37\std{5.26}$
 & $122.58\std{21.01}$ & $214.06\std{103.49}$
 & $36.44\std{2.44}$ & $162.78\std{20.94}$
 & $100.34\std{3.52}$ & $162.04\std{18.66}$ \\
PIS
 & $128$ / $128$
 & $40.35\std{25.29}$ & $34.40\std{12.85}$
 & $2702.24\std{66.69}$ & $142.54\std{5.24}$
 & $165.77\std{1.72}$ & $194.21\std{3.72}$
 & $35.47\std{0.10}$ & $227.64\std{1.74}$
 & $83.39\std{0.13}$ & $230.37\std{1.05}$ \\
TB$+$MALA
 & $128$ / $128$
 & $10.20\std{4.11}$ & $9.50\std{2.66}$
 & $408.80\std{70.18}$ & $17.38\std{2.13}$
 & $508.96\std{66.54}$ & $106.16\std{3.89}$
 & $23.71\std{0.18}$ & $15.31\std{2.42}$
 & $78.48\std{0.37}$ & $21.59\std{1.97}$ \\
SubTB$+$SMC
 & $128$ / $128$
 & $12.67\std{3.94}$ & $10.66\std{2.84}$
 & $314.98\std{23.25}$ & $14.87\std{3.26}$
 & $498.94\std{61.84}$ & $14.03\std{5.65}$
 & $23.75\std{0.12}$ & $13.86\std{1.33}$
 & $76.54\std{1.23}$ & $31.55\std{9.21}$ \\
\bottomrule
\end{tabular}%
}
\vspace*{-1.9em}
\end{table*}

\begin{wraptable}[12]{r}{0.5\textwidth}
\vspace*{-2.5em}
\centering
\caption{Sampler comparison on NICE targets. Conventions follow \Cref{tab:main-results}.}\vspace*{-1em}
\label{tab:main-results-nice}
\begingroup
\resizebox{\linewidth}{!}{%
\begin{tabular}{@{}lrrrrr}
\toprule
Target density $\rightarrow$ &
 & \multicolumn{2}{c}{NICE-MNIST ($d=196$)}
 & \multicolumn{2}{c}{NICE-FMNIST ($d=784$)} \\
\cmidrule(lr){3-4} \cmidrule(lr){5-6}
Method $\downarrow$ Metric $\rightarrow$\hspace*{-1em}
 & NN / $\mathcal E$
 & \multicolumn{1}{c}{$\mathcal{W}_2^2$} & \multicolumn{1}{c}{MMD ($\times10^{-3}$)}
 & \multicolumn{1}{c}{$\mathcal{W}_2^2$} & \multicolumn{1}{c}{MMD ($\times10^{-3}$)} \\
\midrule
Reference
 & 
 & $129.56\std{0.22}$ & $8.75\std{0.08}$
 & $549.42\std{1.35}$ & $9.24\std{1.09}$ \\
\midrule
\rowcolor{osisgreen} OSIS
 & $1$ / $0$
 & $189.59\std{20.34}$ & $184.55\std{68.71}$
 & $1743.96\std{180.58}$ & $382.13\std{59.92}$ \\
\midrule
FAB (flow only)
 & $1$ / $0$
 & $162.20\std{1.91}$ & $57.94\std{11.09}$
 & $1768.75\std{335.78}$ & $430.25\std{72.05}$ \\
FAB
 & $120$ / $120$
 & $143.83\std{3.10}$ & $80.83\std{14.39}$
 & $1746.47\std{335.35}$ & $438.79\std{71.94}$ \\
SMC
 & $0$ / $3825$
 & $268.32\std{4.16}$ & $246.17\std{13.54}$
 & $2856.36\std{15.26}$ & $629.18\std{3.24}$ \\
DIS
 & $128$ / $128$
 & $307.73\std{55.89}$ & $208.21\std{22.97}$
 & $5321.68\std{2656.55}$ & $421.53\std{124.32}$ \\
PIS
 & $128$ / $128$
 & $237.63\std{3.08}$ & $225.89\std{16.69}$
 & $2793.73\std{19.11}$ & $290.08\std{4.09}$ \\
TB$+$MALA
 & $128$ / $128$
 & $266.44\std{7.03}$ & $329.57\std{14.95}$
 & $3074.87\std{23.88}$ & $430.23\std{4.81}$ \\
SubTB$+$SMC
 & $128$ / $128$
 & $285.59\std{80.69}$ & $296.43\std{144.70}$
 & $4035.49\std{82.91}$ & $531.04\std{15.03}$ \\
\bottomrule
\end{tabular}
}
\endgroup
\end{wraptable}

\paragraph{Baselines.} We evaluate \osis against several  amortised samplers: the normalising-flow-based FAB \citep{midgley2022flow} and diffusion samplers trained with different algorithms -- PIS \citep{zhang2021path}, DIS
\citep{berner2024optimal}, and off-policy methods TB$+$MALA \citep{sendera2024improved} and SubTB$+$SMC \citep{choi2025reinforced}. Since FAB can either be used as a one-step sampler or placed in an annealed importance sampling loop, we compare \osis to the one-step version of FAB as well. Additionally, we compare \osis to SMC \citep{del2006sequential}, a training-free sampling algorithm.

\begin{wrapfigure}[14]{r}{0.45\textwidth}\vspace{-7.5em}
    \centering
    \includegraphics[width=1.0\linewidth, trim=8 6 0 6, clip]{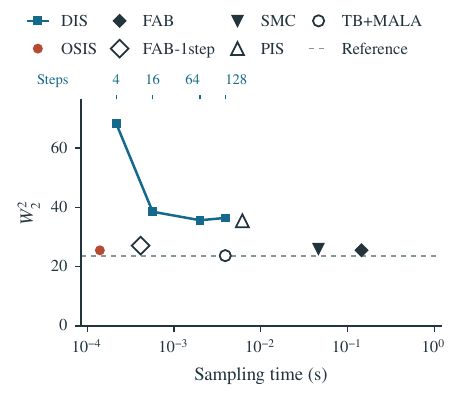}
    \vspace*{-1.4em}
    \caption{Sampling time comparison on ManyWell (median of $250$ runs). DIS is trained for different numbers of denoising steps. \osis achieves $\mathcal{W}_2^2$ comparable to the baselines while sampling up to $1000\times$ faster.}
    \label{fig:efficiency}
\end{wrapfigure}

\begin{figure}
    \centering
    \resizebox{\linewidth}{!}{
    \begin{tabular}{@{}c@{} *{7}{c @{\hspace{1pt}}} c@{}}
         &  & \multicolumn{2}{c}{OSIS}  & \multicolumn{2}{c}{FAB }& \multicolumn{2}{c}{Diff.\ samplers}     &  \\\cmidrule(lr){3-4}\cmidrule(lr){5-6}\cmidrule(lr){7-8}
         & Target & Base & + MALA & AIS & flow only & DIS  & PIS & SMC
         \\

        \raisebox{2mm}{\scriptsize \raisebox{0.5cm}{\rotatebox[origin=c]{90}{40GMM}}} &
        \includegraphics[width=0.125\linewidth]{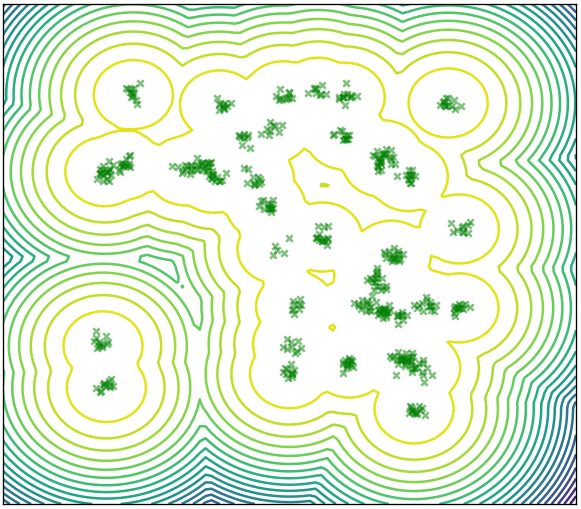}&
        \includegraphics[width=0.125\linewidth]{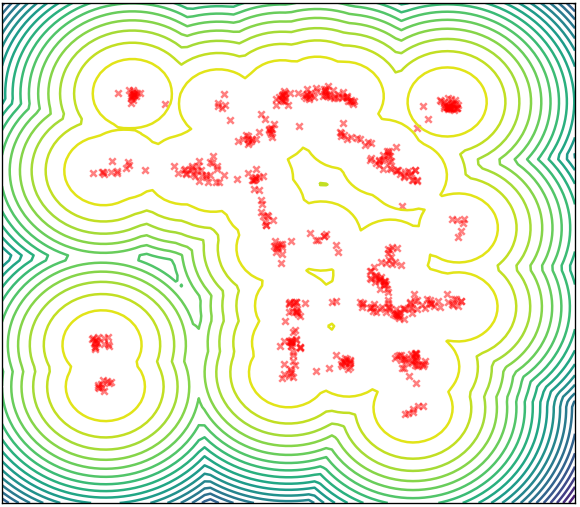} &
        \includegraphics[width=0.125\linewidth]{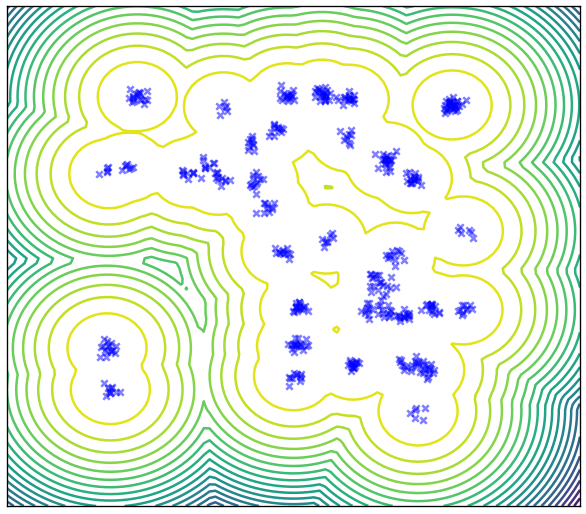} &
        \includegraphics[width=0.125\linewidth]{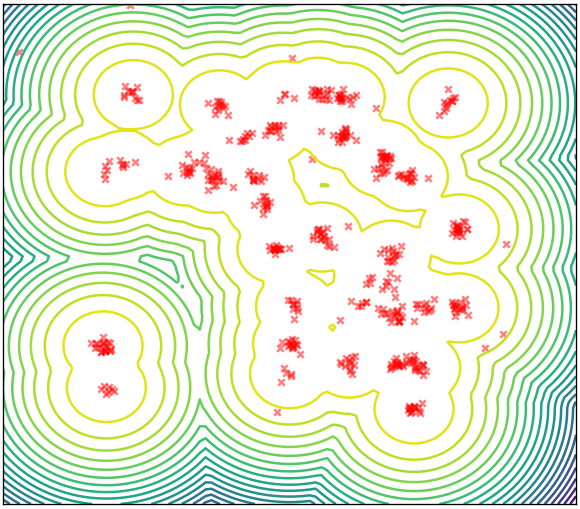} &
        \includegraphics[width=0.125\linewidth]{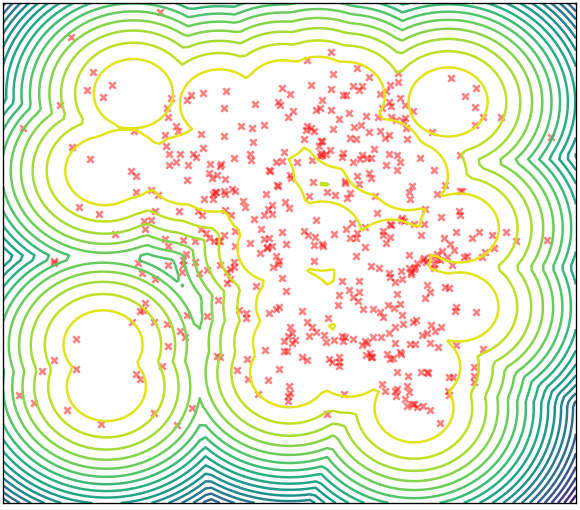} &
        \includegraphics[width=0.125\linewidth]{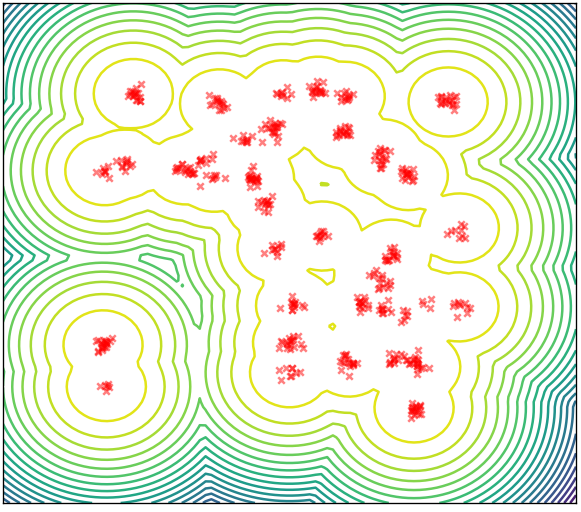} &
        \includegraphics[width=0.125\linewidth]{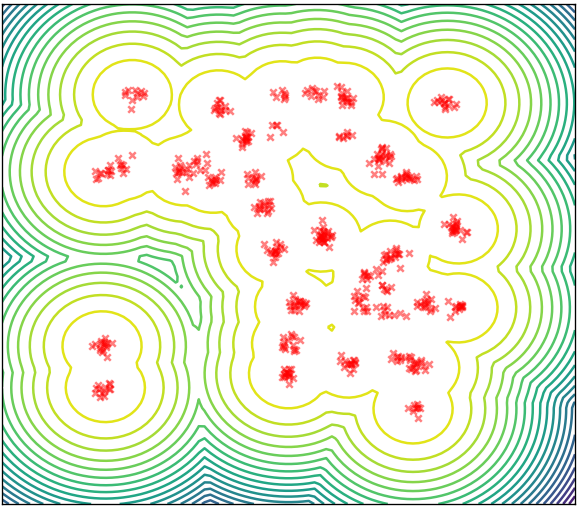} &
        \includegraphics[width=0.125\linewidth]{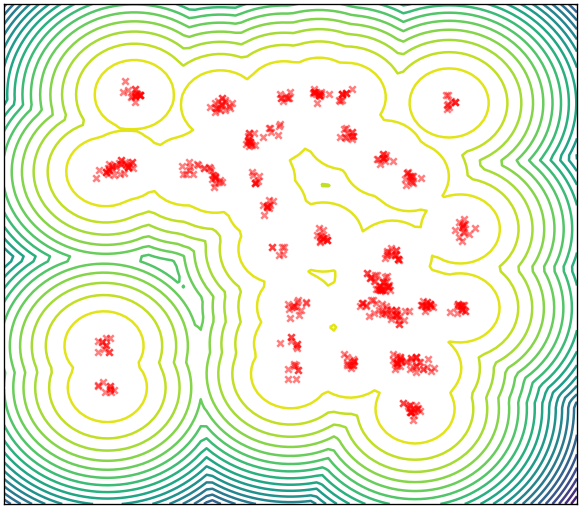} \\[-2pt]
        
        \raisebox{2mm}{\scriptsize \raisebox{0.5cm}{\rotatebox[origin=c]{90}{Funnel}}} &
        \includegraphics[width=0.125\linewidth]{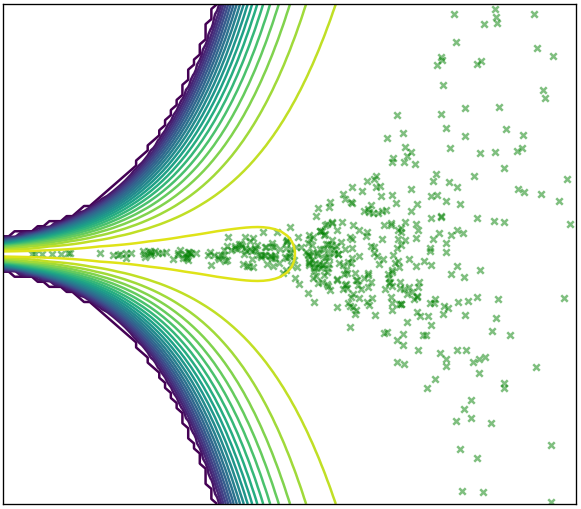}&
        \includegraphics[width=0.125\linewidth]{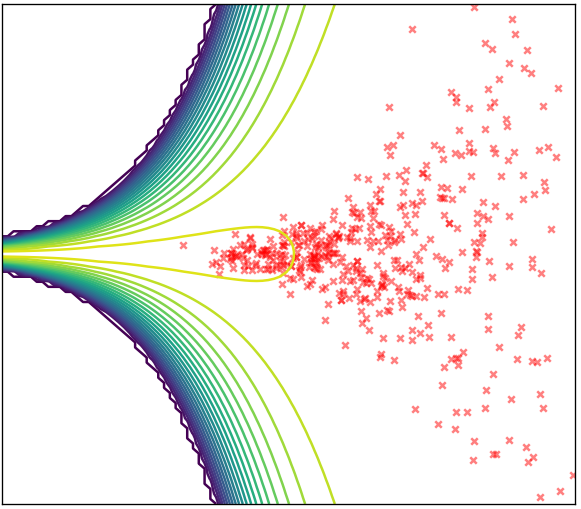} &
        \includegraphics[width=0.125\linewidth]{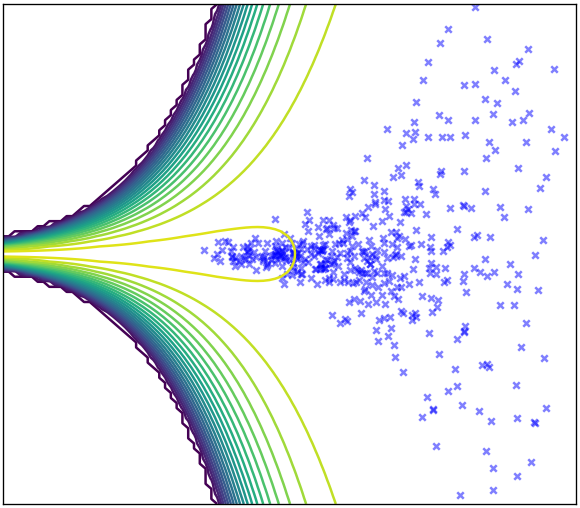} &
        \includegraphics[width=0.125\linewidth]{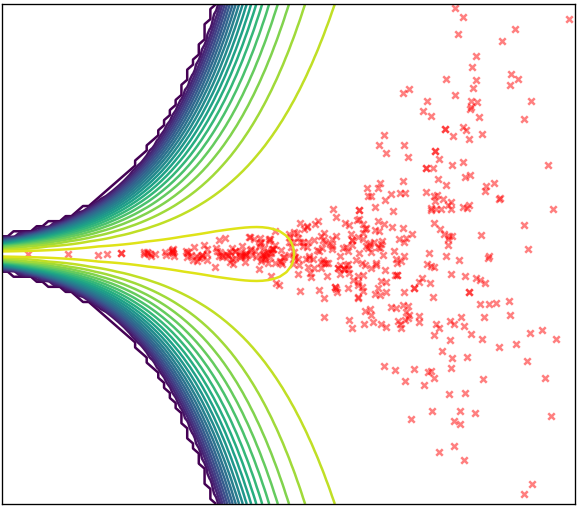} &
        \includegraphics[width=0.125\linewidth]{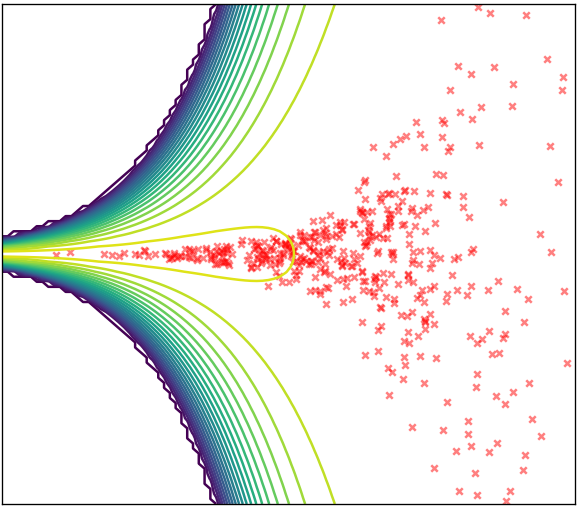} &
        \includegraphics[width=0.125\linewidth]{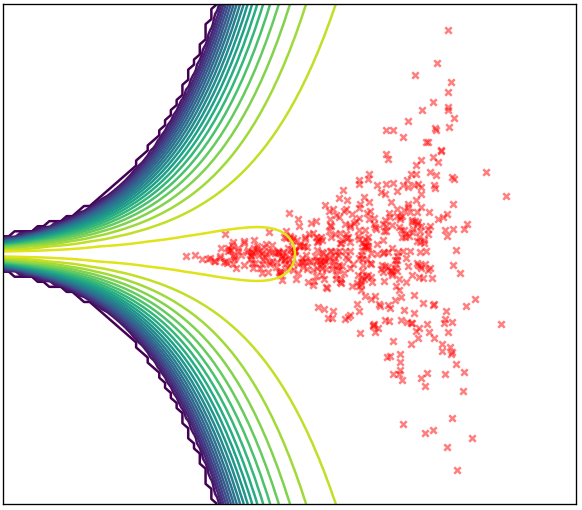} &
        \includegraphics[width=0.125\linewidth]{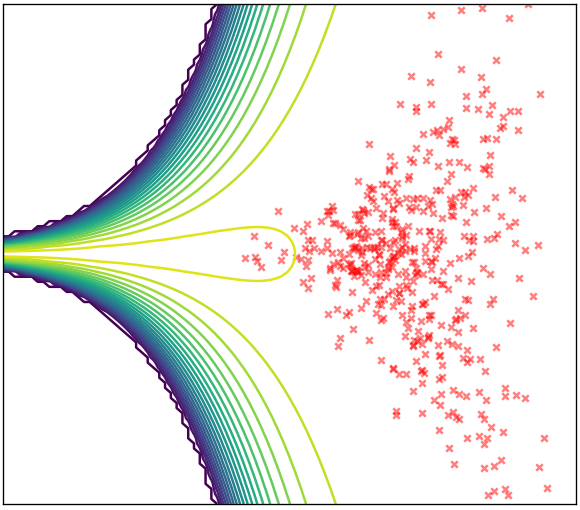} &
        \includegraphics[width=0.125\linewidth]{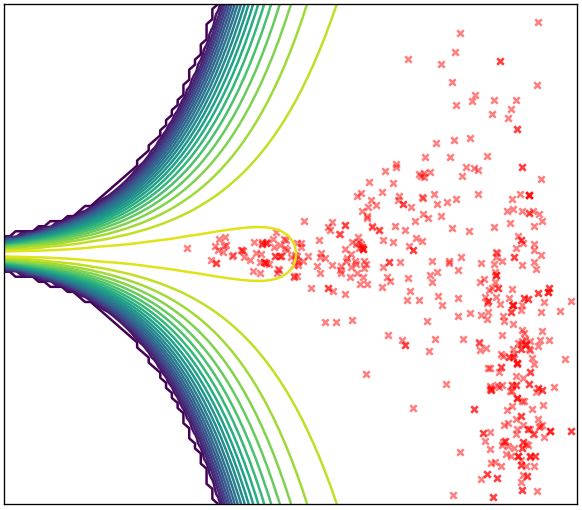} \\[-2pt]

        \raisebox{2mm}{\scriptsize \raisebox{0.5cm}{\rotatebox[origin=c]{90}{ManyWell}}} &
        \includegraphics[width=0.125\linewidth]{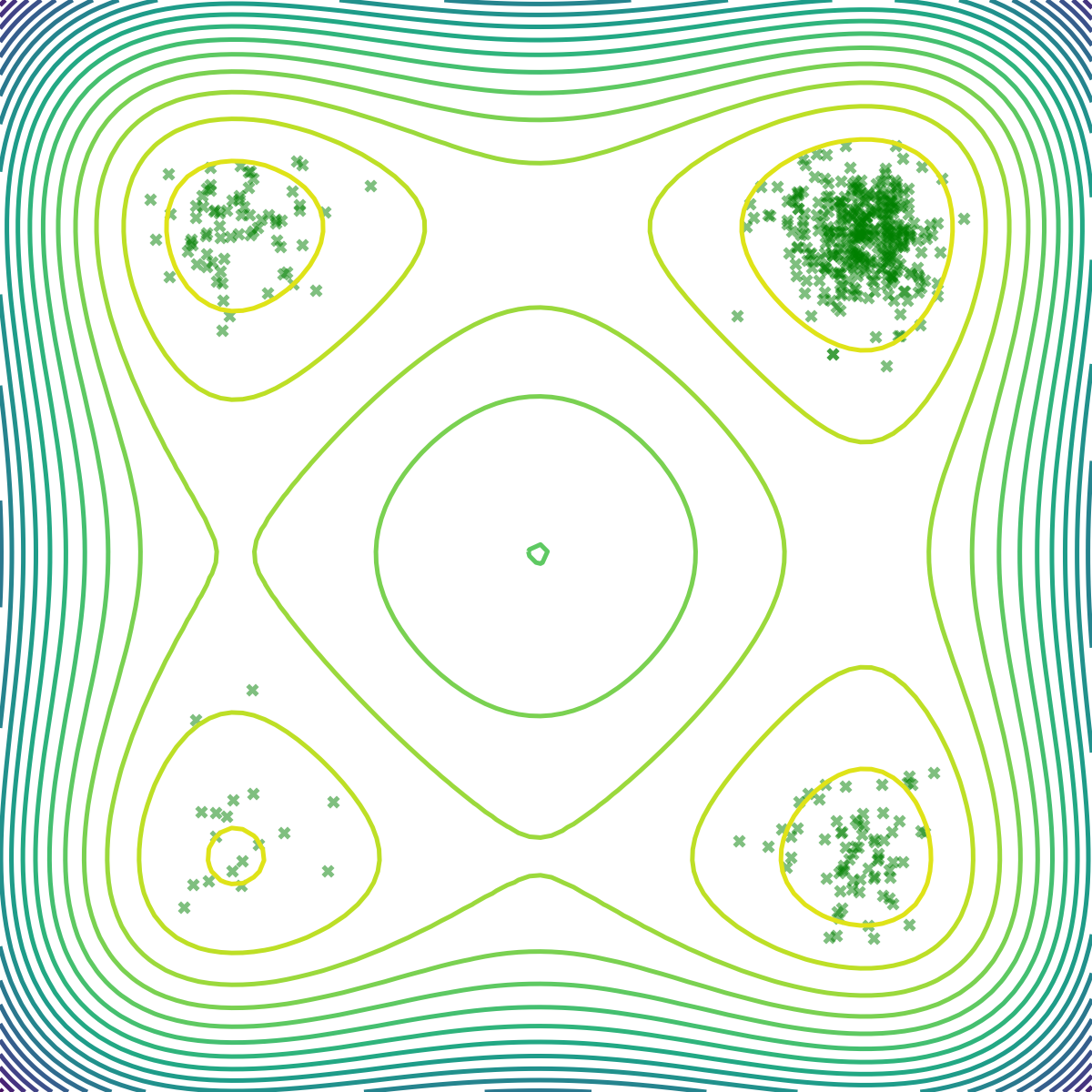}&
        \includegraphics[width=0.125\linewidth]{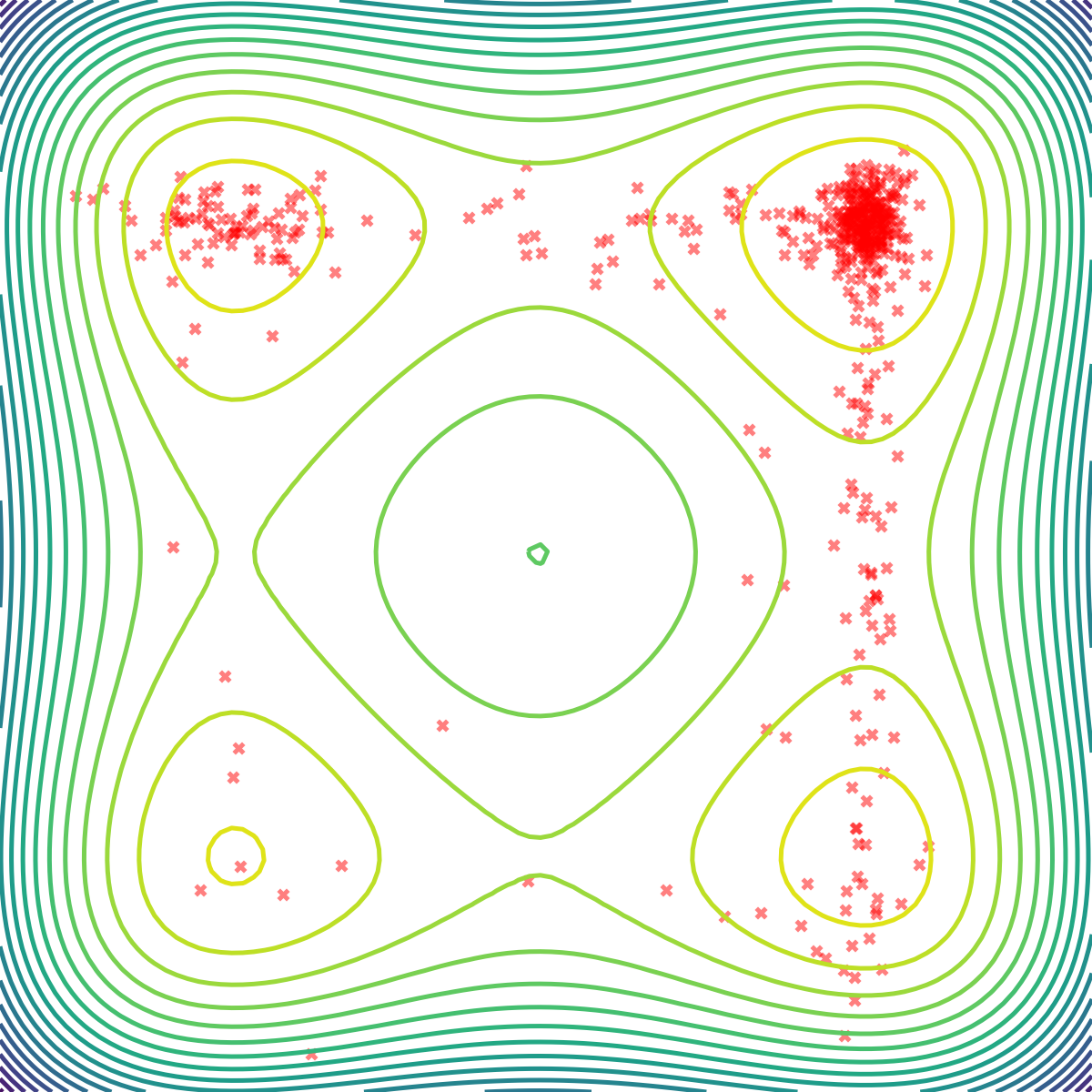} &
        \includegraphics[width=0.125\linewidth]{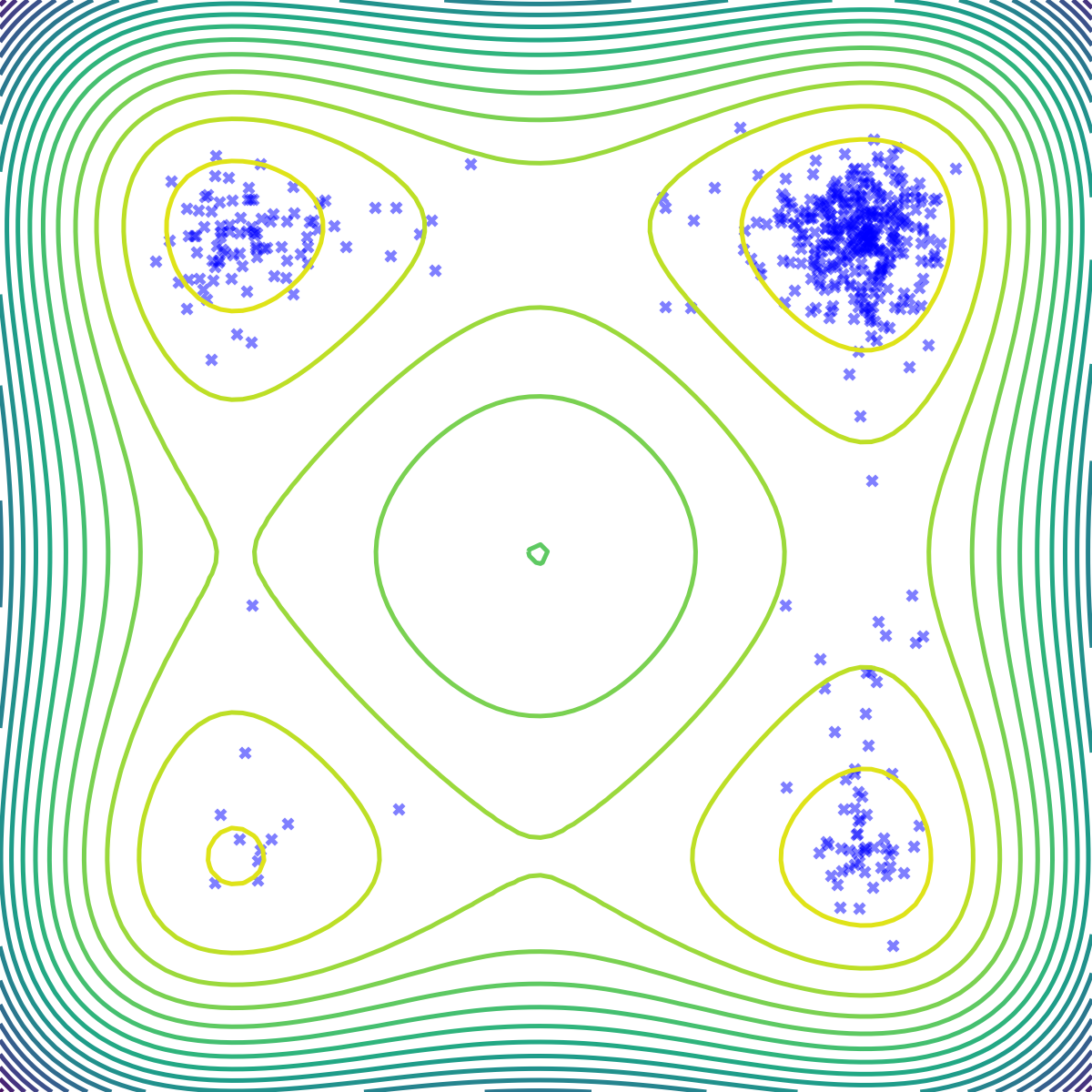} &
        \includegraphics[width=0.125\linewidth]{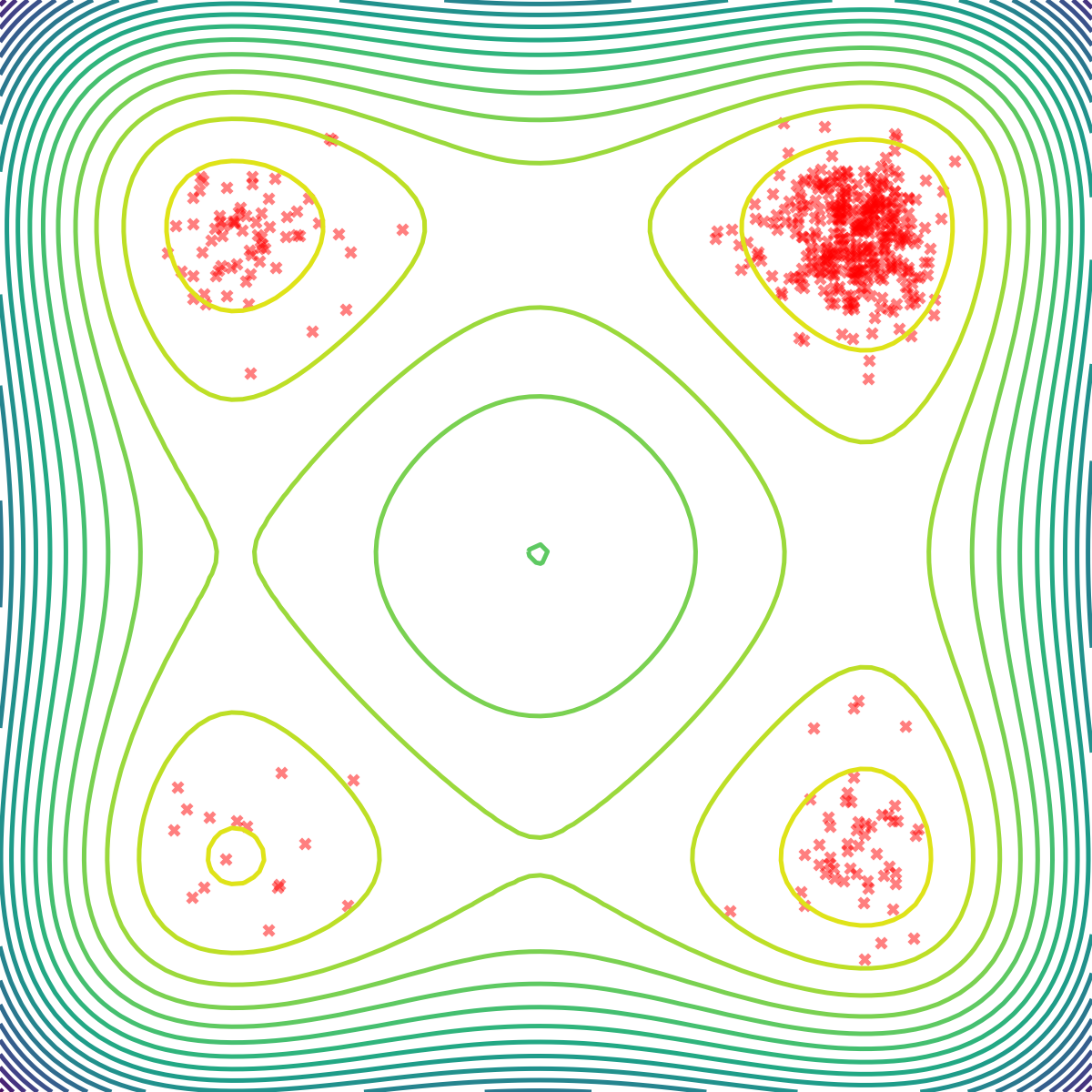} &
        \includegraphics[width=0.125\linewidth]{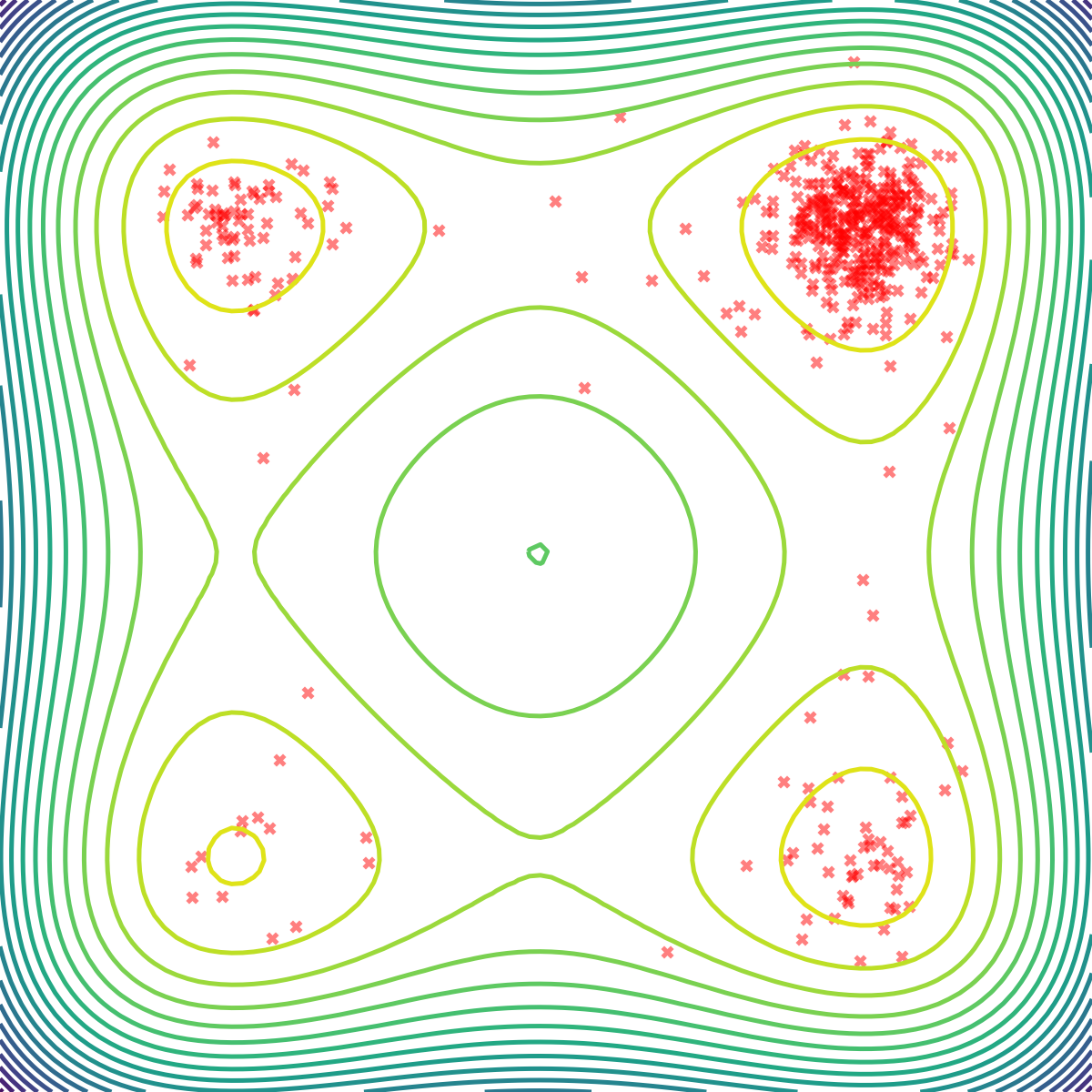} &
        \includegraphics[width=0.125\linewidth]{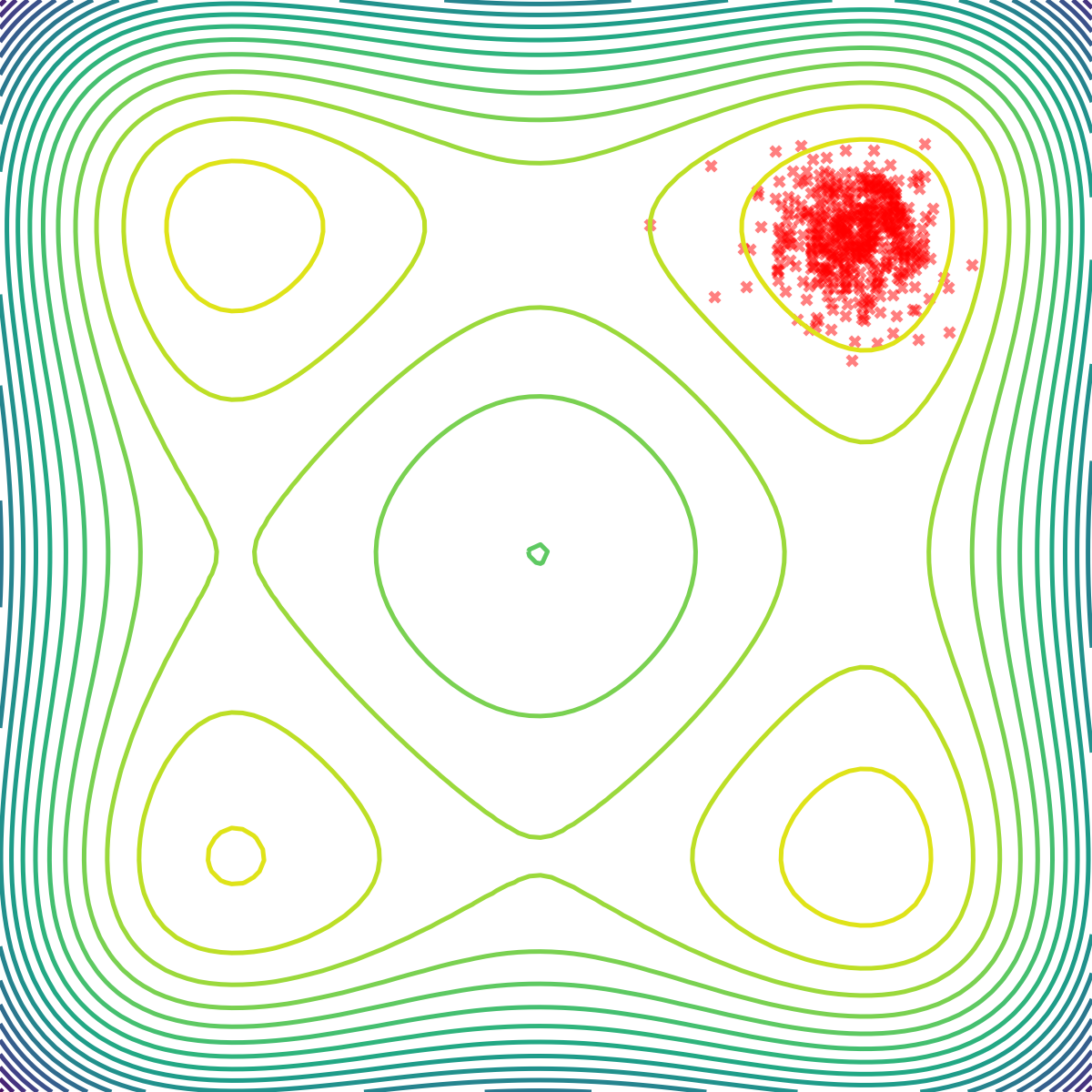} &
        \includegraphics[width=0.125\linewidth]{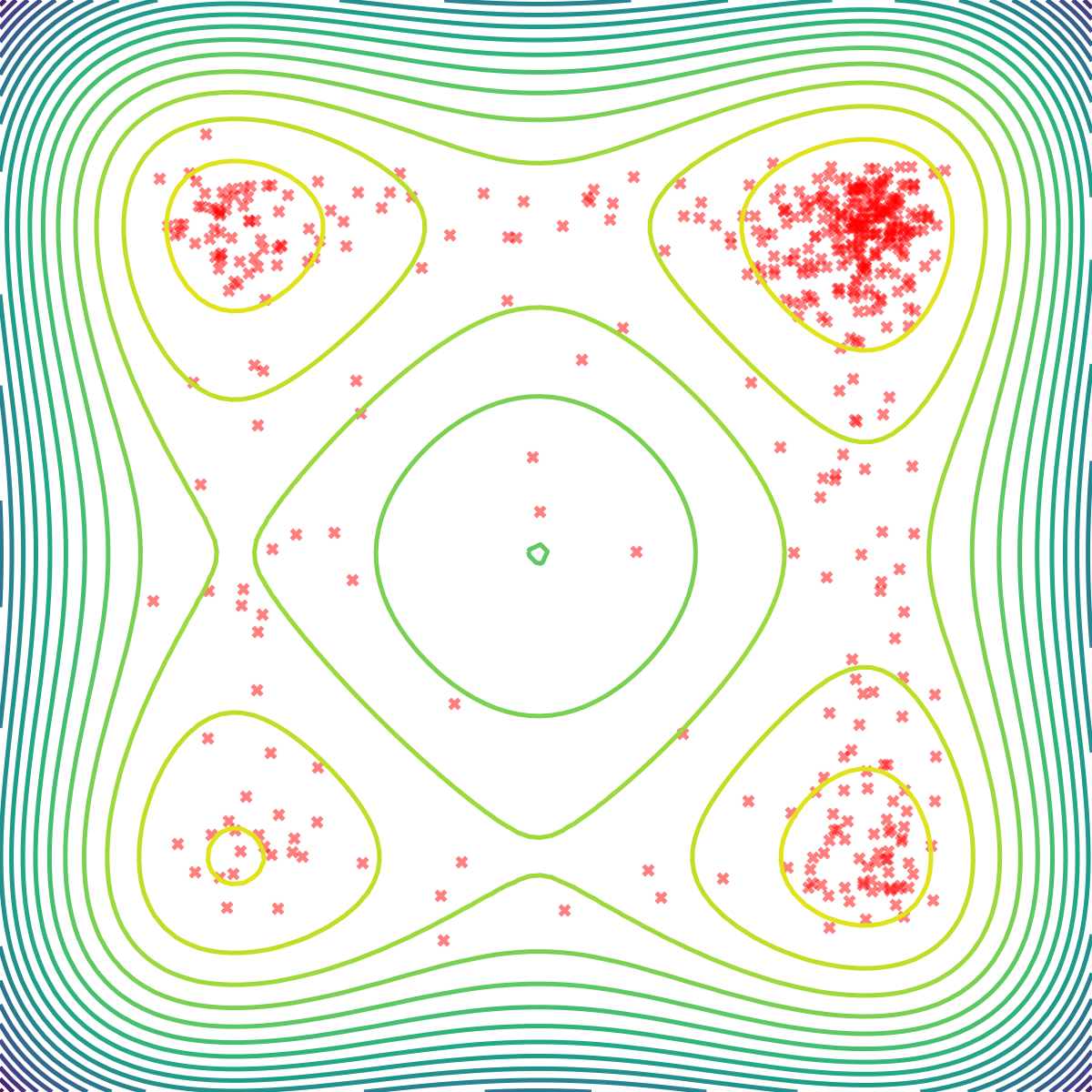} &
        \includegraphics[width=0.125\linewidth]{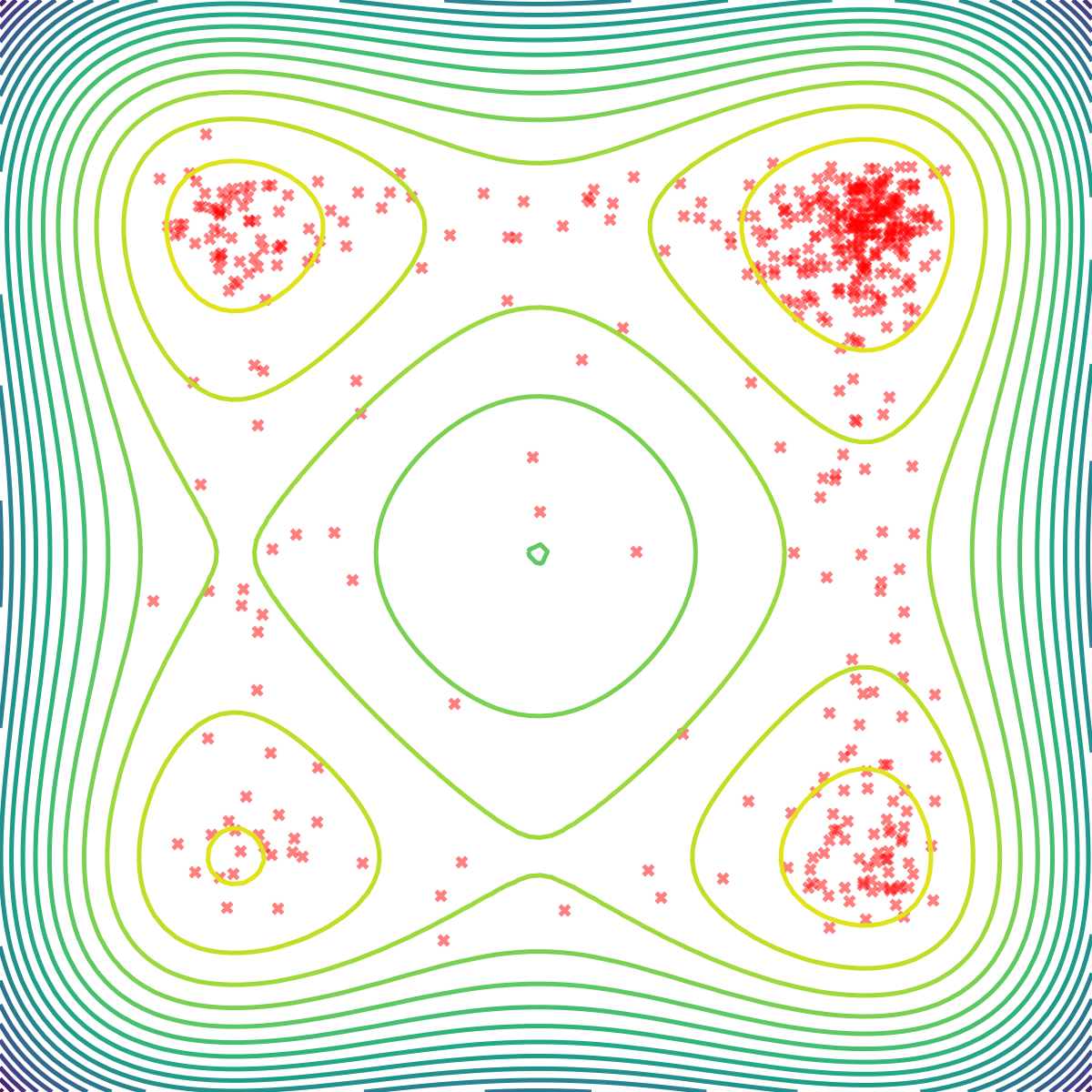} \\[-2pt]

        \raisebox{2mm}{\scriptsize \raisebox{0.5cm}{\rotatebox[origin=c]{90}{MNSIT}}} &
        \includegraphics[width=0.125\linewidth, trim=0 112 280 0, clip]{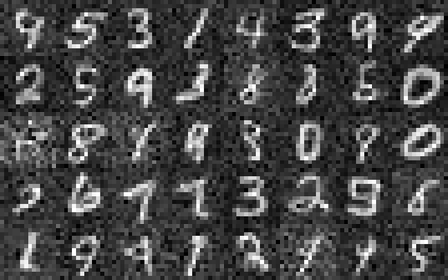}&
        \includegraphics[width=0.125\linewidth, trim=168 0 112 112, clip]{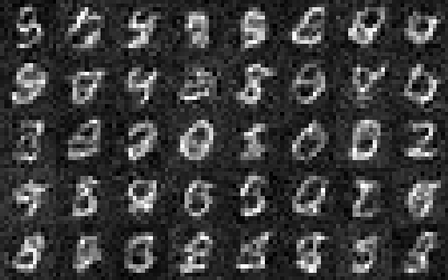} &
        \includegraphics[width=0.125\linewidth, trim=168 0 112 112, clip]{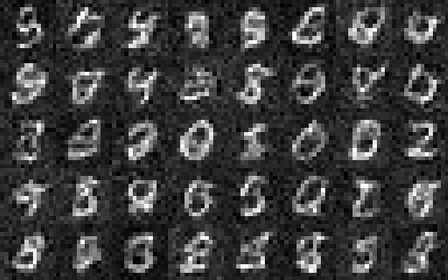} &
        \includegraphics[width=0.125\linewidth, trim=0 112 280 0, clip]{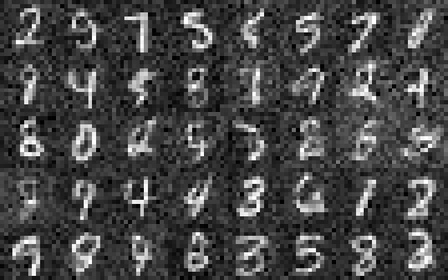} &
        \includegraphics[width=0.125\linewidth, trim=0 112 280 0, clip]{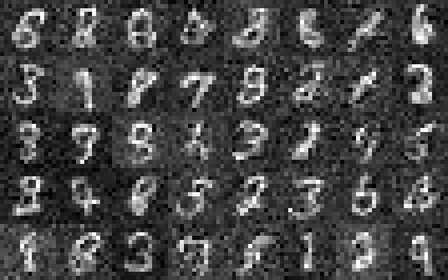} &
        \includegraphics[width=0.125\linewidth, trim=0 112 280 0, clip]{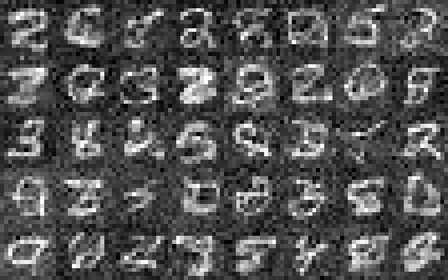} &
        \includegraphics[width=0.125\linewidth, trim=0 112 280 0, clip]{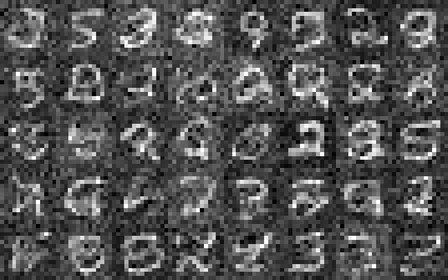} &
        \includegraphics[width=0.125\linewidth, trim=0 112 280 0, clip]{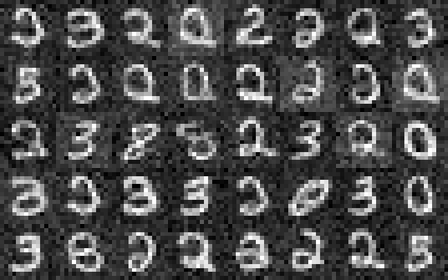} \\
    \end{tabular}
    }
    \vspace*{-1em}
    \caption{
        \looseness=-1
        Visualisation of samples generated by \osis and the baselines for 
        40GMM ($d=2$), Funnel ($d=10$), ManyWell ($d=32$), MNIST ($d=196$). Funnel and ManyWell are projected onto $\R^2$.
    }\vspace*{-1em}
    \label{fig:sampler-visualizations}
\end{figure}

\paragraph{Results.} 
\cref{tab:main-results} shows sample-based metrics on synthetic targets, with 2-Wasserstein and MMD evaluated between 10000 draws from a trained \osis model and an equal number of oracle samples from the target. \osis is not consistently outperformed by any of the MCMC, normalising flow, or diffusion-based baselines despite requiring only one neural function evaluation and no energy evaluations to draw a sample. While the diffusion-based samplers trained by off-policy methods \citep{sendera2024improved,choi2025reinforced} are the strongest overall, they require many sampling steps to perform well, \emph{cf.}\ the analysis in \citet{berner2026from}. On the higher-dimensional targets -- given by normalising flow densities trained on MNIST and FMNIST treated as energy-based models -- \osis again performs on par with FAB and stronger than diffusion-based samplers (\cref{tab:main-results-nice}).

\Cref{fig:sampler-visualizations} shows samples from various models on 40GMM ($d=2$), 
Funnel ($d=10$), ManyWell ($d=32$), and MNIST ($d=2$). It is clear from the figure that \osis performs comparably in terms of sampling quality while being significantly faster at inference.

\paragraph{Sampling efficiency.}\looseness=-1 The main advantage of one-step generative models is low sampling cost. \osis, a one-step sampler, is much more efficient at sampling time than most baselines, whose sampling procedures require hundreds of function evaluations. To quantify this exactly, in \cref{fig:efficiency} we show sampling time per batch of $2000$ samples on ManyWell ($d=32$), measured on an NVIDIA GH200 GPU, excluding compilation and warm-up. \osis yields a significant cost reduction with comparable or improved metrics (\cref{tab:main-results}). To assess the trade-off between number of sampling steps and sampling quality, we also train and evaluate the diffusion-based sampler DIS with $4$, $16$, $64$, and $128$ denoising steps. Although DIS performance improves with the number of denoising steps, \osis consistently outperforms at strictly lower sampling time.

\begin{figure}
    \centering
    \begin{tabular}{@{} *{3}{c@{\hspace{1mm}}} c @{}}
         \textit{smiling toddler} & 
         \textit{a young blond girl} & 
         \textit{a man with a beard} & 
         \textit{\!\!\!\!person wearing glasses\!\!\!\!} \\
         \cmidrule(r){1-1}\cmidrule(r){2-2}\cmidrule(r){3-3}\cmidrule{4-4}

         \includegraphics[width=0.24\linewidth, trim = 192 192 64 64, clip]{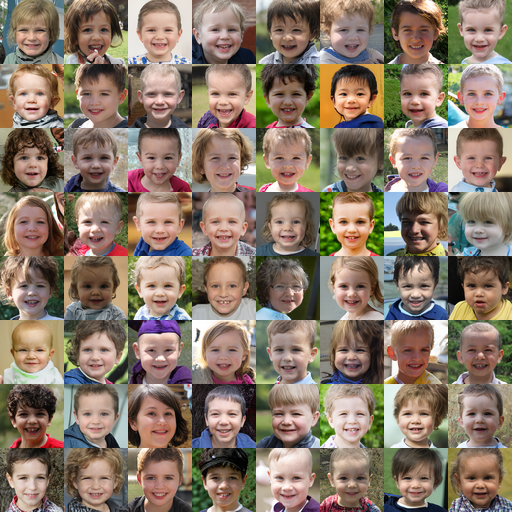} &
         \includegraphics[width=0.24\linewidth, trim = 256 256 0 0, clip]{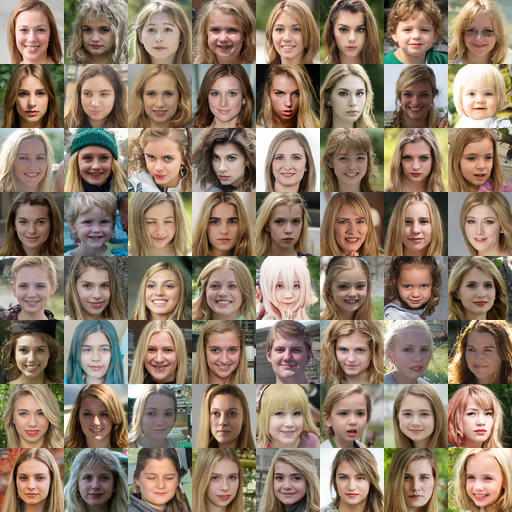} &
         \includegraphics[width=0.24\linewidth, trim = 256 192 0 64, clip]{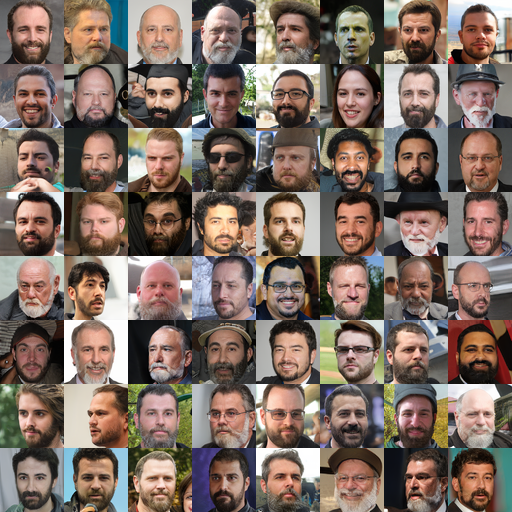} &
         \includegraphics[width=0.24\linewidth, trim = 0 0 256 256, clip]{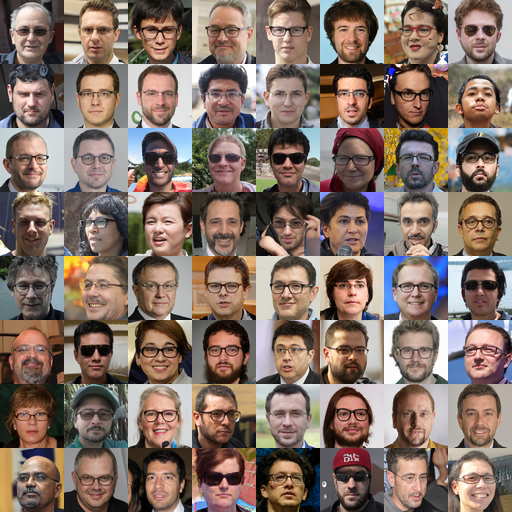} \\[-3pt]
         \bottomrule
    \end{tabular}\\[-0.5em]
    \caption{
        Uncurated samples obtained using outsourced
        sampling (\Cref{subsec:outsourced}). The samplers correspond to the ImageReward-based target conditioned on the given prompts. More examples in \Cref{sub:outsourced-additional-ffhq}.\label{fig:osis-outsourced-ir}
    }
    \vspace*{-2em}
\end{figure}

\begin{wraptable}[8]{r}{0.45\textwidth}
\vspace*{-1.4em}
\centering
\caption{
OSIS kernel ablation with $M=1$ inner iterations (\cref{alg:osis-div-ful}); 
details in \cref{apdx:experimental_setup}.
}\vspace*{-1em}
\label{tab:kernel-comparison}
\resizebox{\linewidth}{!}{%
\begin{tabular}{@{}lrrrr@{}}
\toprule
 Target density $\rightarrow$ & \multicolumn{2}{c}{40GMM ($d=2$)} & \multicolumn{2}{c}{ManyWell ($d=32$)} \\
\cmidrule(lr){2-3} \cmidrule(lr){4-5}
Kernel $\downarrow$ Metric $\rightarrow$ & \multicolumn{1}{c}{$\mathcal{W}_2^2$} & \multicolumn{1}{c}{MMD ($\times10^{-3}$)} & \multicolumn{1}{c}{$\mathcal{W}_2^2$} & \multicolumn{1}{c}{MMD ($\times10^{-3}$)} \\
\midrule
Reference & $7.68\std{3.22}$ & $7.17\std{1.51}$ & $23.61\std{0.07}$ & $8.75\std{0.72}$ \\
\midrule
RWMH & $90.29\std{1.08}$ & $49.88\std{0.56}$ & $30.08\std{0.62}$ & $112.31\std{6.76}$ \\
ULA & $85.42\std{25.65}$ & $60.98\std{13.75}$  & $25.07\std{0.29}$ & $39.54\std{6.08}$ \\
MALA & $103.26\std{41.41}$ & $64.37\std{13.38}$ & $26.79\std{0.41}$ & $55.92\std{6.00}$ \\
BAOAB & $40.24\std{16.81}$ & $41.66\std{15.91}$ & $43.88\std{3.70}$ & $227.46\std{52.12}$ \\
OBABO & $81.85\std{31.72}$ & $58.06\std{15.95}$ & $28.77\std{0.53}$ & $85.55\std{8.86}$ \\
HMC & $117.90\std{26.08}$ & $96.11\std{18.25}$ & $30.08\std{2.55}$ & $106.84\std{61.35}$ \\
SVGD & $463.12\std{263.92}$ & $260.72\std{125.03}$ & $38.57\std{3.62}$ & $269.97\std{50.16}$ \\
\bottomrule
\end{tabular}
}
\end{wraptable}

\paragraph{Ablation of kernel choice.} \cref{tab:kernel-comparison} compares different choices of the kernel in \osis (see \cref{subsec:kernel-options} for a description of the kernels). MALA, ULA, and OBABO tend to preform the strongest, and the higher-order splitting schemes for underdamped Langevin (OBABO and BAOAB) often improve over MALA, approaching the diffusion sampler baselines. However, based on the experiments, \osis does not consistently benefit from sophisticated MCMC kernels. 

Additionally, we compare with the RWMH kernel, which does not use energy gradients. Although it does not provide the best performance, it remains competitive and can serve as an alternative when target-energy gradients are unavailable. Note that these methods are compared in terms of \osis training steps, not at matched computational cost, and different kernels require different numbers of energy and gradient evaluations per step. Further ablations are provided in \cref{apx:osis-ablation}.

\subsection{Bayesian posterior sampling under deep generative model priors}\label{subsec:outsourced}

We show the scalability of \osis by applying it to the problem of sampling from a Bayesian posterior under a pretrained generative model prior in latent space, known otherwise as \emph{outsourced sampling} \citep{venkatraman2025outsourced}. Suppose that a pretrained generative model is expressed as a deterministic transformation of a latent variable: $x=g(z)$, where $z\sim q(z)$ follows a known base distribution. The model's distribution over $x$, $p=g_\#q$, is treated as a Bayesian prior. If $r$ is a likelihood or reward function, we wish to sample from the posterior, or reward-tilted distribution (which has density proportional to $p(x)r(x)$ if $p$ has a density).

It can be shown that this posterior is the pushforward by $g$ of the latent-space distribution with density proportional to $q(z)r(g(z))$, reducing sampling from the posterior in $x$-space to sampling from this distribution over the latent $z$.  Past work adopting such a formulation \citep{venkatraman2025outsourced,tang2025inferencetime, kalaivanan2025ess, om2025posterior,wang2026source,tamogashev2026data} used various amortised and Monte Carlo methods to sample from the latent-space posterior; here, we train \osis samplers.

We conduct two sets of experiments: 
\begin{enumerate}[left=0pt,nosep,label=(\arabic*)]
\item with $g$ the decoder of a VAE \citep{kingma2013auto} trained on MNIST with 16-dimensional latent, with a pretrained classifier's likelihood $p_{\text{classifier}}(c\mid x)$ used to define the reward;
\item with $g$ the generator of a R3GAN \citep{huang2024gan} trained on FFHQ-64 with 64-dimensional latent, with exponentiated ImageReward \citep{xu2023imagereward} as a reward, $r(x)=\exp(\beta R_{\text{ImageReward}}(x,c))$ for a given caption $c$.
\end{enumerate}

Details are given in \Cref{app:algorithm-additional}.

\paragraph{MNIST with classifier reward.} For the experiments with a
VAE pretrained on MNIST, we use a mixture of different classes as the reward:  $r(x) = \sum_{c=0}^9 w_cp(c\mid x)$. Samples are shown in \Cref{sub:outsourced-additional-mnist}.

\begin{wraptable}[8]{r}{0.4\textwidth}
\vspace*{-1.3em}
\centering
\caption{
    Comparison on the outsourced sampling problem.
}\vspace*{-1em}
\label{tab:outsourced}
\resizebox{\linewidth}{!}{
\begin{tabular}{@{}lcc}
    \toprule
    Algorithm $\downarrow$ Metric $\rightarrow$ &
    $\mathbb{E}[\log r(x)]$ ($\uparrow$) &
    FID $(\downarrow)$ \\ 
    \midrule
    Rejection sampling &
    $2.164$ & $2.345$ \\
    MALA &
    $2.127$ & $3.070$ \\
    \midrule
    OSIS &
    $1.387$ & $8.965$ \\
    OSIS + refinement &
    $1.854$ & $7.510 $\\
    TB$+$MALA &
    $0.992$ & $4.267$ \\
    \bottomrule
\end{tabular}
}
\end{wraptable}
\paragraph{FFHQ-64 with ImageReward.} We sample from the reward-reweighted distribution, which is defined by ImageReward conditioned on a prompt. For these experiments, we consider the four prompts shown in \Cref{fig:osis-outsourced-ir}, which shows the decoded samples from a trained \osis model. For the prompt \textit{a young blond girl}, we compare \osis to a diffusion sampler (\citet{sendera2024improved}, as used for this problem in \citet{venkatraman2025outsourced}), to an MCMC algorithm (MALA with adaptive step size), and to unbiased samples obtained by an expensive rejection sampling; see \cref{tab:outsourced}. \osis performs similarly -- higher mean reward, lower FID -- to the diffusion-based amortised baseline (TB$+$MALA) while requiring only one function evaluation per sample. \osis further improves with MCMC refinement. Visualisations in \Cref{sub:outsourced-additional-ffhq} show comparable quality of samples.

\section{Conclusion}\label{sec:conclusion}

This paper has proposed an algorithm for one-step amortised sampling based on matching the distribution induced by a generator with its MCMC refinement. With appropriate techniques to make the refinement and divergence optimisation practical, \osis performs surprisingly well compared to strong samplers that require many neural net and energy evaluations to draw each sample.

The most immediate challenge is to improve the sampling quality on highly multimodal targets, where diffusion samplers have been successful due to the geometric mixing induced by the noise, which is absent with tempering alone. The theoretical analysis should also be extended to include temperature annealing and inexact integration of the dynamics within various refinement kernels.

\subsubsection*{Acknowledgements}
The work of AC is supported by the UKRI EPSRC through the CDT in Machine Learning Systems hosted in the School of Informatics, University of Edinburgh (EP/Y03516X/1), as well as joint sponsorship from Level E Research. The work of SC and ESW is supported by the Advanced Research and Invention Agency (ARIA).  SB and ESW acknowledge support from the CIFAR Learning in Machines and Brains programme. This work has been partially supported by the Munich Center for Machine Learning (MCML) and G-Research.

This work was enabled by the computational resources of the Edinburgh International Data Facility (EIDF), the Edinburgh Generative AI Laboratory (GAIL), and the John von Neumann Institute for Computing (NIC) on the GCS Supercomputer JUPITER---JUWELS at J\"ulich Supercomputing Centre (JSC).

\bibliography{references}

@article{welling2011bayesian,
	title	= {Bayesian learning via stochastic gradient {Langevin} dynamics},
	author	= {Welling, Max and Teh, Yee W},
	journal	= {International Conference on Machine Learning (ICML)},
	year	= {2011},
}

@article{sohl2015deep,
	title	= {Deep unsupervised learning using nonequilibrium thermodynamics},
	author	= {Sohl-Dickstein, Jascha and Weiss, Eric and Maheswaranathan, Niru and Ganguli, Surya},
	journal	= {International Conference on Machine Learning (ICML)},
	year	= {2015},
}

@article{ho2020denoising,
	title	= {Denoising diffusion probabilistic models},
	author	= {Ho, Jonathan and Jain, Ajay and Abbeel, Pieter},
	journal	= {Neural Information Processing Systems (NeurIPS)},
	year	= {2020},
}

@article{song2020score,
	title	= {Score-Based Generative Modeling through Stochastic Differential Equations},
	author	= {Song, Yang and Sohl-Dickstein, Jascha and Kingma, Diederik P and Kumar, Abhishek and Ermon, Stefano and Poole, Ben},
    year    = 2021,
	journal	= {International Conference on Learning Representations (ICLR)},
}

@article{blessing2024beyond,
	title	= {Beyond {ELBOs}: A Large-Scale Evaluation of Variational Methods for Sampling},
	author	= {Blessing, Denis and Jia, Xiaogang and Esslinger, Johannes and Vargas, Francisco and Neumann, Gerhard},
	journal	= {International Conference on Machine Learning (ICML)},
	year	= {2024},
}

@article{del2006sequential,
	title	= {Sequential {Monte Carlo} samplers},
	author	= {Del Moral, Pierre and Doucet, Arnaud and Jasra, Ajay},
	journal	= {Journal of the Royal Statistical Society: Series B (Statistical Methodology)},
	year	= {2006},
}

@article{midgley2022flow,
	title	= {Flow annealed importance sampling bootstrap},
	author	= {Midgley, Laurence Illing and Stimper, Vincent and Simm, Gregor NC and Sch{\"o}lkopf, Bernhard and Hern{\'a}ndez-Lobato, Jos{\'e} Miguel},
	journal	= {International Conference on Learning Representations (ICLR)},
	year	= {2023},
}

@article{vargas2023denoising,
	title	= {Denoising diffusion samplers},
	author	= {Vargas, Francisco and Grathwohl, Will and Doucet, Arnaud},
	journal	= {International Conference on Learning Representations (ICLR)},
	year	= {2023},
}

@article{berner2024optimal,
	title	= {An optimal control perspective on diffusion-based generative modeling},
	author	= {Julius Berner and Lorenz Richter and Karen Ullrich},
	journal	= {Transactions on Machine Learning Research (TMLR)},
	year	= {2024},
}

@article{sendera2024improved,
	title	= {Improved off-policy training of diffusion samplers},
	author	= {Sendera, Marcin and Kim, Minsu and Mittal, Sarthak and Lemos, Pablo and Scimeca, Luca and Rector-Brooks, Jarrid and Adam, Alexandre and Bengio, Yoshua and Whitammer, Esmeralda S.},
	journal	= {Neural Information Processing Systems (NeurIPS)},
	year	= {2024},
}

@article{berner2026from,
	title	= {From discrete-time policies to continuous-time diffusion samplers: Asymptotic equivalences and faster training},
	author	= {Julius Berner and Lorenz Richter and Marcin Sendera and Jarrid Rector-Brooks and Nikolay Malkin},
	journal	= {Transactions on Machine Learning Research (TMLR)},
	year	= {2026},
}

@article{noe2019boltzmann,
	title	= {Boltzmann generators: Sampling equilibrium states of many-body systems with deep learning},
	author	= {No{\'e}, Frank and Olsson, Simon and K{\"o}hler, Jonas and Wu, Hao},
	journal	= {Science},
	year	= {2019},
}

@article{duane1987hybrid,
	title	= {Hybrid {Monte Carlo}},
	author	= {Duane, Simon and Kennedy, Anthony D and Pendleton, Brian J and Roweth, Duncan},
	journal	= {Physics Letters B},
	year	= {1987},
}

@article{roberts1996exponential,
	title	= {Exponential convergence of {Langevin} distributions and their discrete approximations},
	author	= {Roberts, Gareth O and Tweedie, Richard L},
    journal = {},
	year	= {1996},
}

@article{andrieu2008tutorial,
	title	= {A tutorial on adaptive {MCMC}},
	author	= {Andrieu, Christophe and Thoms, Johannes},
	journal	= {Statistics and computing},
	year	= {2008},
}

@article{vempala2019rapid,
	title	= {Rapid convergence of the unadjusted {Langevin} algorithm: Isoperimetry suffices},
	author	= {Vempala, Santosh and Wibisono, Andre},
	journal	= {Neural Information Processing Systems (NeurIPS)},
	year	= {2019},
}

@article{chewi2025analysis,
	title	= {Analysis of {Langevin Monte Carlo} from {Poincar\'e} to log-{Sobolev}},
	author	= {Chewi, Sinho and Erdogdu, Murat A and Li, Mufan and Shen, Ruoqi and Zhang, Matthew S},
	journal	= {Foundations of Computational Mathematics},
	year	= {2025},
}

@article{bonneel2015sliced,
	title	= {Sliced and {Radon} {Wasserstein} barycenters of measures},
	author	= {Bonneel, Nicolas and Rabin, Julien and Peyr{\'e}, Gabriel and Pfister, Hanspeter},
	journal	= {Journal of Mathematical Imaging and Vision},
	year	= {2015},
}

@article{kolouri2019generalized,
	title	= {Generalized sliced {Wasserstein} distances},
	author	= {Kolouri, Soheil and Nadjahi, Kimia and Simsekli, Umut and Badeau, Roland and Rohde, Gustavo},
	journal	= {Neural Information Processing Systems (NeurIPS)},
	year	= {2019},
}

@article{nadjahi2020statistical,
	title	= {Statistical and topological properties of sliced probability divergences},
	author	= {Nadjahi, Kimia and Durmus, Alain and Chizat, L{\'e}na{\"\i}c and Kolouri, Soheil and Shahrampour, Shahin and Simsekli, Umut},
	journal	= {Neural Information Processing Systems (NeurIPS)},
	year	= {2020},
}

@article{santambrogio2015optimal,
	title	= {Optimal Transport for Applied Mathematicians: Calculus of Variations, {PDEs}, and Modeling, volume 87 of Progress in Nonlinear Differential Equations and Their Applications},
	author	= {Santambrogio, Filippo},
    journal = {},
	year	= {2015},
}

@book{ambrosio2005gradient,
	title	= {Gradient flows: in metric spaces and in the space of probability measures},
	author	= {Ambrosio, Luigi and Gigli, Nicola and Savar{\'e}, Giuseppe},
	year	= {2005},
    publisher={Springer}
}

@article{jordan1998variational,
	title	= {The variational formulation of the {Fokker--Planck} equation},
	author	= {Jordan, Richard and Kinderlehrer, David and Otto, Felix},
	journal	= {SIAM journal on mathematical analysis},
	year	= {1998},
}

@article{flamary2021pot,
	title	= {Pot: Python optimal transport},
	author	= {Flamary, R{\'e}mi and Courty, Nicolas and Gramfort, Alexandre and Alaya, Mokhtar Z and Boisbunon, Aur{\'e}lie and Chambon, Stanislas and Chapel, Laetitia and Corenflos, Adrien and Fatras, Kilian and Fournier, Nemo and others},
	journal	= {Journal of Machine Learning Research},
	year	= {2021},
}

@article{cuturi2013sinkhorn,
	title	= {Sinkhorn distances: Lightspeed computation of optimal transport},
	author	= {Cuturi, Marco},
	journal	= {Neural Information Processing Systems (NeurIPS)},
	year	= {2013},
}

@article{feydy2019interpolating,
	title	= {Interpolating between optimal transport and {MMD} using {Sinkhorn} divergences},
	author	= {Feydy, Jean and S{\'e}journ{\'e}, Thibault and Vialard, Fran{\c{c}}ois-Xavier and Amari, Shun-ichi and Trouv{\'e}, Alain and Peyr{\'e}, Gabriel},
	journal	= {International Conference on Artificial Intelligence and Statistics (AISTATS)},
	year	= {2019},
}

@article{liu2016stein,
	title	= {Stein variational gradient descent: A general purpose {Bayesian} inference algorithm},
	author	= {Liu, Qiang and Wang, Dilin},
	journal	= {Neural Information Processing Systems (NeurIPS)},
	year	= {2016},
}

@article{loshchilov2017decoupled,
	title	= {Decoupled weight decay regularization},
	author	= {Loshchilov, Ilya and Hutter, Frank},
	journal	= {International Conference on Learning Representations (ICLR)},
	year	= {2019},
}

@article{he2016deep,
	title	= {Deep residual learning for image recognition},
	author	= {He, Kaiming and Zhang, Xiangyu and Ren, Shaoqing and Sun, Jian},
	journal	= {Computer Vision and Pattern Recognition (CVPR)},
	year	= {2016},
}

@article{ronneberger2015u,
	title	= {U-net: Convolutional networks for biomedical image segmentation},
	author	= {Ronneberger, Olaf and Fischer, Philipp and Brox, Thomas},
	journal	= {Medical Image Computing and Computer-Assisted Intervention (MICCAI)},
	year	= {2015},
}

@article{wirnsberger2020targeted,
	title	= {Targeted free energy estimation via learned mappings},
	author	= {Wirnsberger, Peter and Ballard, Andrew J and Papamakarios, George and Abercrombie, Stuart and Racani{\`e}re, S{\'e}bastien and Pritzel, Alexander and Jimenez Rezende, Danilo and Blundell, Charles},
	journal	= {The Journal of Chemical Physics},
	year	= {2020},
}

@article{holdijk2023stochastic,
	title	= {Stochastic optimal control for collective variable free sampling of molecular transition paths},
	author	= {Holdijk, Lars and Du, Yuanqi and Hooft, Ferry and Jaini, Priyank and Ensing, Berend and Welling, Max},
	journal	= {Neural Information Processing Systems (NeurIPS)},
	year	= {2023},
}

@article{bakry1985diffusions,
	title	= {Diffusions hypercontractives},
	author	= {Bakry, Dominique and {\'E}mery, Michel},
    journal = {Séminaire de probabilités},
	year	= {1985},
}

@article{otto2000generalization,
	title	= {Generalization of an inequality by {Talagrand} and links with the logarithmic {Sobolev} inequality},
	author	= {Otto, Felix and Villani, C{\'e}dric},
	journal	= {Journal of Functional Analysis},
	year	= {2000},
}

@article{cao2026one,
	title	= {One-Step Sampler for {Boltzmann} Distributions via Drifting},
	author	= {Wenhan Cao and Keyu Yan and Lin Zhao},
	year	= {2026},
	journal	= {arXiv preprint arXiv:2603.17579},

}

@article{deng2026generative,
	title	= {Generative Modeling via Drifting},
	author	= {Deng, Mingyang and Li, He and Li, Tianhong and Du, Yilun and He, Kaiming},
	journal	= {arXiv preprint arXiv:2602.04770},
	year	= {2026},
}

@book{douc2018markov,
	author	= {Douc, Randal and Moulines, Eric and Priouret, Pierre and Soulier, Philippe},
	title	= {Markov Chains},
	year	= {2018},
    publisher = {Springer},
}

@article{gretton2013two,
	author	= {Arthur Gretton and Karsten M. Borgwardt and Malte J. Rasch and Bernhard Sch{{\"o}}lkopf and Alexander Smola},
	title	= {A Kernel Two-Sample Test},
	journal	= {Journal of Machine Learning Research},
	year	= {2012},
}

@article{metropolis1953equation,
	title	= {Equation of state calculations by fast computing machines},
	author	= {Metropolis, Nicholas and Rosenbluth, Arianna W and Rosenbluth, Marshall N and Teller, Augusta H and Teller, Edward},
	journal	= {The Journal of Chemical Physics},
	year	= {1953},
}

@article{hastings1970monte,
	title	= {Monte {Carlo} sampling methods using {Markov} chains and their applications},
	author	= {Hastings, W Keith},
    journal = {},
	year	= {1970},
}

@article{neal2011mcmc,
	title	= {{MCMC} using {Hamiltonian} dynamics},
	author	= {Neal, Radford M},
	journal	= {Handbook of {Markov} chain {Monte Carlo}},
	year	= {2011},
}

@article{zhang2021path,
	title	= {Path integral sampler: a stochastic control approach for sampling},
	author	= {Zhang, Qinsheng and Chen, Yongxin},
	journal	= {International Conference on Learning Representations (ICLR)},
	year	= {2022},
}

@article{richter2020vargrad,
	title	= {{VarGrad}: A Low-Variance Gradient Estimator for Variational Inference},
	author	= {Lorenz Richter and Ayman Boustati and Nikolas Nüsken and Francisco J. R. Ruiz and Ömer Deniz Akyildiz},
	year	= {2020},
	journal	= {Neural Information Processing Systems (NeurIPS)},
}

@article{lahlou2023theory,
	title	= {A theory of continuous generative flow networks},
	author	= {Lahlou, Salem and Deleu, Tristan and Lemos, Pablo and Zhang, Dinghuai and Volokhova, Alexandra and Hern{\'a}ndez-Garc{\i}a, Alex and Ezzine, L{\'e}na N{\'e}hale and Bengio, Yoshua and Malkin, Nikolay},
	journal	= {International Conference on Machine Learning (ICML)},
	year	= {2023},
}

@article{richter2024improved,
	title	= {Improved sampling via learned diffusions},
	author	= {Richter, Lorenz and Berner, Julius and Liu, Guan-Horng},
	journal	= {International Conference on Learning Representations (ICLR)},
	year	= {2024},
}

@article{chen2025sequential,
	title	= {Sequential Controlled {Langevin} Diffusions},
	author	= {Junhua Chen and Lorenz Richter and Julius Berner and Denis Blessing and Gerhard Neumann and Anima Anandkumar},
	journal	= {International Conference on Learning Representations (ICLR)},
	year	= {2025},
}

@article{albergo2025nets,
	title	= {{NETS}: A Non-Equilibrium Transport Sampler},
	author	= {Michael S. Albergo and Eric Vanden-Eijnden},
	year	= {2025},
	journal	= {International Conference on Machine Learning (ICML)},
}

@article{rezende2015variational,
	title	= {Variational inference with normalizing flows},
	author	= {Rezende, Danilo and Mohamed, Shakir},
	year	= {2015},
	journal	= {International Conference on Machine Learning (ICML)},
}

@article{neal2000slice,
	title	= {Slice sampling},
	author	= {Neal, Radford M},
	journal	= {The Annals of Statistics},
	year	= {2003},
}

@article{fatras2021minibatch,
	title	= {Minibatch optimal transport distances; analysis and applications},
	author	= {Kilian Fatras and Younes Zine and Szymon Majewski and Rémi Flamary and Rémi Gribonval and Nicolas Courty},
	year	= {2021},
	journal	= {arXiv preprint arXiv:2101.01792},
}

@article{huang2024gan,
	title	= {The {GAN} is dead; long live the {GAN}! {A} Modern {GAN} Baseline},
	author	= {Yiwen Huang and Aaron Gokaslan and Volodymyr Kuleshov and James Tompkin},
	year	= {2024},
	journal	= {Neural Information Processing Systems (NeurIPS)},
}

@article{alfonsi2020squared,
	title	= {Squared quadratic {Wasserstein} distance: optimal couplings and {Lions} differentiability},
	author	= {Alfonsi, Aur{\'e}lien and Jourdain, Benjamin},
	journal	= {ESAIM: Probability and Statistics},
	year	= {2020},
}

@article{li2017approximate,
	title	= {Approximate inference with amortised {MCMC}},
	author	= {Li, Yingzhen and Turner, Richard E and Liu, Qiang},
	journal	= {arXiv preprint arXiv:1702.08343},
	year	= {2017},
}

@article{han2026one,
	title	= {One-Step Generative Modeling via {Wasserstein} Gradient Flows},
    author  = {Han, Jiaqi and Li, Puheng and Guo, Qiushan
               and Xu, Renyuan and Ermon, Stefano and Candès, Emmanuel J.},
	year	= {2026},
	journal	= {arXiv preprint arXiv:2605.11755},
}

@article{choi2025reinforced,
	title	= {Reinforced sequential {Monte Carlo} for amortised sampling},
    author  = {Choi, Sanghyeok and Mittal, Sarthak and Elvira, V{\'\i}ctor 
               and Park, Jinkyoo and Whitammer, Esmeralda S.},
	journal	= {International Conference on Machine Learning (ICML)},
	year	= {2026},
}

@article{song2017nice,
	title	= {{A}-{NICE}-{MC}: Adversarial training for {MCMC}},
	author	= {Song, Jiaming and Zhao, Shengjia and Ermon, Stefano},
	journal	= {Neural Information Processing Systems (NeurIPS)},
	year	= {2017},
}

@article{he2025no,
    title   = {No trick, no treat: Pursuits and challenges towards 
               simulation-free training of neural samplers},
    author  = {He, Jiajun and Du, Yuanqi and Vargas, Francisco and Zhang, Dinghuai 
               and Padhy, Shreyas and OuYang, RuiKang and Gomes, Carla 
               and Hern{\'a}ndez-Lobato, Jos{\'e} Miguel},
    journal = {Frontiers in Probabilistic Inference workshop 
               at International Conference on Machine Learining (ICML)},
	year	= {2025},
}

@article{gretton2026wasserstein,
	title	= {On the {Wasserstein} gradient flow interpretation of drifting models},
    author  = {Gretton, Arthur and Wenliang, Li Kevin and Galashov, Alexandre 
               and Thornton, James and De Bortoli, Valentin and Doucet, Arnaud},
	journal	= {arXiv preprint arXiv:2605.05118},
	year	= {2026},
}

@article{he2026sinkhorn,
	title	= {Sinkhorn-drifting generative models},
	author	= {He, Ping and Khangaonkar, Om and Pirsiavash, Hamed and Bai, Yikun and Kolouri, Soheil},
	journal	= {arXiv preprint arXiv:2603.12366},
	year	= {2026},
}

@article{mokrov2021large,
	title	= {Large-scale {Wasserstein} gradient flows},
    author  = {Mokrov, Petr and Korotin, Alexander and Li, Lingxiao 
               and Genevay, Aude and Solomon, Justin M and Burnaev, Evgeny},
	journal	= {Neural Information Processing Systems (NeurIPS)},
	year	= {2021},
}

@article{tian2026sobolev,
	title	= {Sobolev regularized {MMD} gradient flow},
	author	= {Tian, Chenyang and Sriperumbudur, Bharath K and Gretton, Arthur and Chen, Zonghao},
	journal	= {arXiv preprint arXiv:2605.11884},
	year	= {2026},
}

@article{galashov2025deep,
	title	= {Deep {MMD} gradient flow without adversarial training},
	author	= {Galashov, Alexandre and De Bortoli, Valentin and Gretton, Arthur},
	journal	= {International Conference on Learning Representations (ICLR)},
	year	= {2025},
}

@article{chen2025regularized,
	title	= {{(De)}-regularized maximum mean discrepancy gradient flow},
    author  = {Chen, Zonghao and Mustafi, Aratrika and Glaser, Pierre and Korba, Anna 
               and Gretton, Arthur and Sriperumbudur, Bharath K},
	journal	= {Journal of Machine Learning Research},
	year	= {2025},
}

@article{santambrogio2016euclidea,
  title={$\{$Euclidean, metric, and Wasserstein$\}$ gradient flows: an overview},
  author={Santambrogio, Filippo},
  journal={Bulletin of Mathematical Sciences},
  volume={7},
  number={1},
  pages={87--154},
  year={2017},
  publisher={Springer}
}

@article{leimkuhler2013rational,
    title={Rational construction of stochastic numerical methods for molecular sampling},
    author={Leimkuhler, Benedict and Matthews, Charles},
    journal={Applied Mathematics Research eXpress},
    year={2013},
}

@article{leimkuhler2013robust,
  title={Robust and efficient configurational molecular sampling via {Langevin} dynamics},
  author={Leimkuhler, Benedict and Matthews, Charles},
  journal={The Journal of Chemical Physics},
  volume={138},
  number={17},
  year={2013},
  publisher={AIP Publishing}
}

@article{havens2025adjoint,
    title   = {Adjoint sampling: Highly scalable diffusion samplers via adjoint matching},
    author  = {Havens, Aaron and Miller, Benjamin Kurt and Yan, Bing and Domingo-Enrich, Carles and 
               Sriram, Anuroop and Wood, Brandon and Levine, Daniel 
               and Hu, Bin and Amos, Brandon and Karrer, Brian and Fu, Xiang and Liu, Guan-Horng and Chen, Ricky T. Q.},
    journal = {International Conference on Machine Learining (ICML)},
    year    = {2025}
}

@article{strang1968construction,
  title={On the construction and comparison of difference schemes},
  author={Strang, Gilbert},
  journal={SIAM journal on numerical analysis},
  volume={5},
  number={3},
  pages={506--517},
  year={1968},
  publisher={SIAM}
}

@article{sahoo2024simple,
    title   = {Simple and effective masked diffusion language models},
    author  = {Sahoo, Subham S and Arriola, Marianne and Schiff, Yair and 
               Gokaslan, Aaron and Marroquin, Edgar and Chiu, Justin T and Rush, Alexander 
               and Kuleshov, Volodymyr},
    journal = {Neural Information Processing Systems (NeurIPS)},
    year    = {2024}
}

@article{hoogeboom2022equivariant,
    title   = {Equivariant diffusion for molecule generation in 3d},
    author  = {Hoogeboom, Emiel and Satorras, V{\i}ctor Garcia and Vignac, Cl{\'e}ment and Welling, Max},
    journal = {International Conference on Machine Learning (ICML)},
    year={2022},
}

@article{kong2020diffwave,
    title   = {Diffwave: A versatile diffusion model for audio synthesis},
    author  = {Kong, Zhifeng and Ping, Wei and Huang, Jiaji and Zhao, Kexin and Catanzaro, Bryan},
    journal = {International Conference on Learning Representations (ICLR)},
    year    = {2021}
}

@article{samsonov2022local,
    title   = {Local-global {MCMC} kernels: the best of both worlds},
    author  = {Samsonov, Sergey and Lagutin, Evgeny and Gabrié, Marylou 
               and Durmus, Alain and Naumov, Alexey and Moulines, Eric},
    journal = {Neural Information Processing Systems (NeurIPS)},
    year    = {2022}
}

@article{wang2024reinforcement,
    title   = {Reinforcement learning for adaptive {MCMC}},
    author  = {Wang, Congye and Chen, Wilson and Kanagawa, Heishiro and Oates, Chris},
    journal = {International Conference on Artificial Intelligence and Statistics (AISTATS)},
    year    = {2025}
}

@article{durmus2017nonasymptotic,
    title   = {Nonasymptotic convergence analysis for the unadjusted {Langevin} algorithm},
    author  = {Durmus, Alain and Moulines, Eric},
    journal = {The Annals of Applied Probability},
    year    = {2017}
}

@book{leimkuhler2015molecular,
    author  = {Leimkuhler, Benedict and Matthews, Charles},
    title   = {Molecular Dynamics: With Deterministic and Stochastic Numerical Methods},
    publisher = {Springer} ,
    year    = {2015} 
}

@article{mohamed2016learning,
    title={Learning in implicit generative models},
    author={Mohamed, Shakir and Lakshminarayanan, Balaji},
    journal={arXiv preprint arXiv:1610.03483},
    year={2016}
}

@article{feng2017learning,
    title={Learning to draw samples with amortized {Stein} variational gradient descent},
    author={Feng, Yihao and Wang, Dilin and Liu, Qiang},
    journal={Conference on Uncertainty in Artificial Intelligence (UAI)},
    year={2017}
}

@article{titsias2017learning,
    title={Learning model reparametrizations: Implicit variational inference by fitting {MCMC} distributions},
    author={Titsias, Michalis K},
    journal={arXiv preprint arXiv:1708.01529},
    year={2017}
}

@article{hoffman2017learning,
    title={Learning deep latent {Gaussian} models with {Markov} chain {Monte Carlo}},
    author={Hoffman, Matthew D},
    journal={International Conference on Machine Learning (ICML)},
    year={2017},
}

@article{ruiz2019contrastive,
    title={A contrastive divergence for combining variational inference and {MCMC}},
    author={Ruiz, Francisco and Titsias, Michalis},
    journal={International Conference on Machine Learning (ICML)},
    year={2019},
}

@article{gabrie2022adaptive,
    title={Adaptive {Monte Carlo} augmented with normalizing flows},
    author={Gabri{\'e}, Marylou and Rotskoff, Grant M and Vanden-Eijnden, Eric},
    journal={Proceedings of the National Academy of Sciences},
    volume={119},
    number={10},
    pages={e2109420119},
    year={2022},
    publisher={National Academy of Sciences}
}

@article{vonrenesse2005transport,
    author  = {von Renesse, Max-K. and Sturm, Karl-Theodor},
    title   = {Transport inequalities, gradient estimates, entropy and {Ricci} curvature},
    journal = {Communications on Pure and Applied Mathematics},
    year    = {2005}
}

@article{venkatraman2025outsourced,
    title   = {Outsourced diffusion sampling: Efficient posterior inference 
               in latent spaces of generative models},
    author  = {Venkatraman, Siddarth and Hasan, Mohsin and Kim, Minsu 
               and Scimeca, Luca and Sendera, Marcin and Bengio, Yoshua 
               and Berseth, Glen and Malkin, Nikolay},
    journal = {International Conference on Machine Learning (ICML)},
    year    = {2025}
}

@article{kingma2013auto,
    title   = {Auto-encoding variational {Bayes}},
    author  = {Kingma, Diederik P and Welling, Max},
    journal = {International Conference on Learning Representations (ICLR)},
    year    = {2014}
}

@article{xu2023imagereward,
    title   = {{ImageReward}: Learning and evaluating human 
               preferences for text-to-image generation},
    author  = {Xu, Jiazheng and Liu, Xiao and Wu, Yuchen 
               and Tong, Yuxuan and Li, Qinkai and Ding, Ming and Tang, Jie and Dong, Yuxiao},
    journal = {Neural Information Processing Systems (NeurIPS)},
    year    = {2023}
}

@article{tamogashev2026data,
    title   = {Data-to-energy stochastic dynamics},
    author  = {Tamogashev, Kirill and Whitammer, Esmeralda S.},
    journal = {International Conference on Learning Representations (ICLR)},
    year    = {2026}
}

@article{tang2025inferencetime,
    title   = {Inference-Time Alignment of Diffusion Models with Direct Noise Optimization},
    author  = {Zhiwei Tang and Jiangweizhi Peng and Jiasheng Tang and Mingyi Hong and Fan Wang and Tsung-Hui Chang},
    journal = {International Conference on Machine Learning (ICML)},
    year    = {2025}
}

@article{wang2026source,
    title   = {Source-guided flow matching},
    author  ={Wang, Zifan and Harting, Alice and Barreau, Matthieu and Zavlanos, Michael and Johansson, Karl H},
    journal = {International Conference on Learning Representations (ICLR)},
    year    = {2026}
}

@article{kalaivanan2025ess,
    title   = {{ESS-Flow}: Training-free guidance of flow-based models as inference in source space},
    author  = {Kalaivanan, Adhithyan and Zhao, Zheng and Sj{\"o}lund, Jens and Lindsten, Fredrik},
    journal = {arXiv preprint arXiv:2510.05849},
    year    = {2025}
}

@article{om2025posterior,
    title   = {Posterior Inference in Latent Space for Scalable Constrained Black-box Optimization},
    author  = {Om, Kiyoung and Sim, Kyuil and Yun, Taeyoung and Kang, Hyeongyu and Park, Jinkyoo},
    journal = {arXiv preprint arXiv:2507.00480},
    year    = {2025}
}

@article{besag1994mala,
    title = {Comments on “{Representations of knowledge in complex systems}” by {U. Grenander and M.I. Miller}},
    author = {Besag, Julian},
    journal = {Journal of the Royal Statistical Society: Series B (Statistical Methodology)},
    volume = {56},
    pages = {591--592},
    year = {1994}
}

@article{rossky1978brownian,
  title={Brownian dynamics as smart {Monte Carlo} simulation},
  author={Rossky, Peter J and Doll, Jimmie D. and Friedman, Harold L.},
  journal={The Journal of Chemical Physics},
  volume={69},
  pages={4628--4633},
  year={1978}
}

@article{fogliani2026annealing,
  title={Annealing in variational inference mitigates mode collapse: {A} theoretical study on {Gaussian} mixtures},
  author={Fogliani, Luigi and Loureiro, Bruno and Gabri{\'e}, Marylou},
  journal={arXiv preprint arXiv:2602.12923},
  year={2026}
}

@article{yi2022sliced,
    title   = {Sliced {Wasserstein} variational inference},
    author  = {Yi, Mingxuan and Liu, Song},
    year    = {2023},
    journal = {Asian Conference on Machine Learning (ACML)}
}
\bibliographystyle{iclr2027_conference}

\newpage
\appendix

\section{Additional discussion of the algorithm}\label{app:algorithm-additional}
\subsection{Vector-field-based loss}
It is possible to an consider alternative training procedure for \osis, which is similar to the one used for drifting models \citep{deng2026generative, gretton2026wasserstein}. Gradient descent on the parameters $\theta$ along the objective in \eqref{eq:div-objective} approximates the Wasserstein gradient descent of the distribution $\rho$ along the \emph{Wasserstein gradients of the divergence $\mathcal{D}$} (see \Cref{sec:connections}). 
The theory of Wasserstein geometry \citep{ambrosio2005gradient} tells us that the evolution of a distribution can be identified with the evolution of particles that follow this distribution along a gradient field. 
In this case, the evolution of particles follows a vector field $V$ such that $V = - \nabla\frac{\delta \mathcal{D}}{\delta \rho_\theta}$, with the term on the right being the gradient of the first variation of $\mathcal{D}$. 
This observation allows us to use an alternative training objective:
\begin{align}\label{eq:vf-objective}
    \mathcal{L}(\theta) = \mathbb{E}_{z\sim q}\left[\left||
    G_\theta(z) - \sg{G_\theta(z) + \eta V(G_\theta(z))}
    \right\|^2\right],
\end{align}
where now $V$, rather than the divergence $\mathcal{D}$, is estimated empirically using samples from $\rho_\theta$.
The algorithm corresponding to this objective is presented in \Cref{alg:osis-vf} and experiments are in \Cref{sec:drift-experiments}.

\subsection{Full algorithm}
\Cref{alg:osis-div,alg:osis-vf} describe the idealised version of \osis. However, as discussed in \Cref{sec:practical-considerations}, there are many design choices that can help improve performance of the algorithm in practice. 
\Cref{alg:osis-div-ful,alg:osis-vf-ful} present the complete version of \osis and include tempering, approximation of a Markov chain and the divergence with their empirical analogues, as well as the use of a replay buffer and inner iterations. We note that this work focuses on the variant of \osis described in \Cref{alg:osis-div,alg:osis-div-ful}. Pseudocode in \Cref{alg:osis-vf,alg:osis-vf-ful} is provided for completeness. 

\begin{algorithm}[p]
    \caption{OSIS using a vector field $V$}\label{alg:osis-vf}
    \begin{algorithmic}[1]
    \Require Prior distributon $q$,
        $\pi$-invariant kernel $K_\pi$,
        number of chain steps $M$,
        generator $G_\theta$,
        vector field $V_{(\cdot, \cdot)}$,
        step size $\eta$.
    \While{not converged}
        \State $\rho_\theta \leftarrow G_{\theta\,\#}q$
        \LineComment{Run a chain for time $h$}
        \State $\rho_h \leftarrow K^h\rho_{\sg{\theta}}$
        \LineComment{Compute loss using $V$}
        \State $\mathcal{L}(\theta) \leftarrow \mathbb{E}_{z\sim q(z),x=G_\theta(z)}
            \| G_\theta(z) - \sg{x + V_{\rho_{\sg{\theta}},\,\rho_h} (x)} \|^2$
        \State Gradient update on $\theta$ with $\nabla_\theta\mathcal{L}$
    \EndWhile
    \State \Return $G_\theta$
\end{algorithmic}\end{algorithm}

\begin{figure}[p]
\begin{minipage}[t]{0.48\linewidth}
\begin{algorithm}[H]
    \caption{Full \osis algorithm with $\hat{\mathcal{D}}$}
    \label{alg:osis-div-ful}
    \begin{algorithmic}[1]
    \Require Prior distribution $q$,
        generator $G_\theta$,
        $\pi$-invariant kernel $K$,
        empirical divergence $\hat{\mathcal{D}}$,
        number of outer steps $N$, number of inner steps $M$,
        number of chain steps $N_{\rm MCMC}$, 
        buffer size $B$, batch size $\tilde{B}$,
    \For{$\texttt{outer} = 1, \dots, N$}
        \State $\{z_i\}_{i=1}^{B} \sim \mathcal{N}(0, I),\;x_i\leftarrow G_{\sg{\theta}}(z_i)$
        \LineComment{Run a chain for $N_{\rm  MCMC}$ steps} 
        \LineComment{with inverse temperature $\beta$}
        \State $\beta \leftarrow \operatorname{get\_inverse\_temp}(\texttt{outer})$
        \State $y_i \leftarrow \operatorname{run\_mcmc}(x_i,\, K,\, N_{\rm MCMC}, \beta)$
        \LineComment{Instantiate buffer and save samples}
        \State Instantiate buffer $\mathcal{B}$ 
        \State with $\left(\{z_i\}_{i=1}^B,\, \{y_i\}_{i=1}^B\right)$
        \LineComment{Run several inner iterations} 
        \LineComment{for computational stability}
        \For{$\texttt{inner} = 1, \dots, M$}
            \If{\texttt{resample} is \textbf{True}}
                \State $\left(\{z_i\}_{i=1}^{\tilde{B}} ,\, \{y_i\}_{i=1}^{\tilde{B}} \right)\sim \mathcal{B}$ 
            \Else
                \State Get  $\left(\{z_i\}_{i=1}^{\tilde{B}} ,\, \{y_i\}_{i=1}^{\tilde{B}} \right)$ from $\mathcal{B}$
            \EndIf
            \State $x^\theta_i \leftarrow G_\theta(z_i)$
            \State $\hat{p}^{\theta}_x\leftarrow\sum_{j=1}^{\tilde{B}} \delta_{x^\theta_j}$, 
            $\hat{p}_y\leftarrow\sum_{j=1}^{\tilde{B}} \delta_{y_j}$  
            \State $\mathcal{L}(\theta) \leftarrow \hat{\mathcal{D}}(\hat{p}^{\theta}_x, \hat{p}_y)$
            \State Gradient update on $\theta$ with $\nabla_\theta\mathcal{L}$
            \EndFor
    \EndFor
    \State \Return $G_\theta$
\end{algorithmic}\end{algorithm}\end{minipage}
\hfill
\begin{minipage}[t]{0.51\linewidth}
\begin{algorithm}[H]
    \caption{Full \osis algorithm with $V$}
    \label{alg:osis-vf-ful}
    \begin{algorithmic}[1]
    \Require Prior distribution $q$,
        generator $G_\theta$,
        $\pi$-invariant kernel $K$,
        vector field $V_{(\cdot, \cdot)}$,
        number of outer steps $N$, number of inner steps $M$,
        number of chain steps $N_{\rm MCMC}$,
        buffer size $B$, batch size $\tilde{B}$, step size $\eta$.
    \For{$\texttt{outer} = 1, \dots, N$}
        \State $\{z_i\}_{i=1}^{B} \sim \mathcal{N}(0, I),\;x_i\leftarrow G_\theta(z_i)$
        \LineComment{Run a chain for $N_{\rm  MCMC}$ steps} 
        \LineComment{with inverse temperature $\beta$}
        \State $\beta \leftarrow \operatorname{get\_inverse\_temp}(\texttt{outer})$
        \State $y_i \leftarrow \operatorname{run\_mcmc}(\sg{x_i},\, K,\, N_{\rm MCMC}, \beta)$
        \LineComment{Instantiate buffer and save samples}
        \State Instantiate buffer $\mathcal{B}$ with 
               $\left(\{z_i\}_{i=1}^B,\, \{y_i\}_{i=1}^B\right)$

        \LineComment{Run several inner iterations} 
        \LineComment{for computational stability}
        \For{$\texttt{inner} = 1, \dots, M$}
            \If{\texttt{resample} is \textbf{True}}
                \State $\left(\{z_i\}_{i=1}^{\tilde{B}} ,\, \{y_i\}_{i=1}^{\tilde{B}} \right)\sim \mathcal{B}$ 
            \Else
                \State Get  $\left(\{z_i\}_{i=1}^{\tilde{B}} ,\, \{y_i\}_{i=1}^{\tilde{B}} \right)$ from $\mathcal{B}$
            \EndIf
            \State $x^\theta_i \leftarrow G_\theta(z_i)$
            \LineComment{$z\sim q$, $x=G_\theta(z)$}
            \State$\begin{aligned}%
                \mathcal{L}(\theta)&\leftarrow\\
                &\mathbb{E}_{z, x}\| G_\theta(z) - \sg{x + \eta V_{\rho_{\sg{\theta}},\,\rho_h} (x)} \|^2
            \end{aligned}$
            \State Gradient update on $\theta$ with $\nabla_\theta\mathcal{L}$
            \EndFor
    \EndFor
    \State \Return $G_\theta$
\end{algorithmic}\end{algorithm}\end{minipage}
\end{figure}

\clearpage\newpage

\section{Extended theoretical analysis}
\label{app:extended-population}

\subsection{Additional notation and temperature scaling
}\label{app:additional-notation}

\paragraph{KL divergence and Fisher information.}
For distributionsw $\rho,\mu\in\mathcal P_2(\R^d)$, we define KL divergence and Fisher information as follows:
\par\noindent
\begin{minipage}{0.48\textwidth}
\begin{equation}\label{eq:kl}
  \KL{\rho}{\mu}\coloneqq
  \int\rho(x)\log\frac{\rho(x)}{\mu(x)}\,\dd x,
\end{equation}
\end{minipage}%
\hfill
\begin{minipage}{0.48\textwidth}
\begin{equation}\label{eq:fi}
    \FI{\rho}{\mu} \coloneqq \int
    \left|\nabla\log\frac{\rho(x)}{\mu(x)}\right|^2 \rho(x)\,\dd x.
\end{equation}
\end{minipage}
\par\noindent
We assume that for all cases we consider both KL divergence and Fisher information are finite.

\paragraph{2-Wasserstein distance.}
For distributions $\rho,\mu\in\mathcal P_2(\R^d)$, we define 2-Wasserstein distance as
\begin{equation}\label{appxeq:kantorovich-w2}
    \Wp{2}^2(\rho,\mu) = \inf_{\gamma\in\Gamma(\rho,\mu)}
    \int_{\R^d\times\R^d}
    |x - y|^2\,\dd\gamma(x,y),
\end{equation}
where $\Gamma(\rho,\mu)$ is the set of probability measures on $\R^d\times\R^d$ whose marginals are $\rho$ and $\mu$. 

\eqref{appxeq:kantorovich-w2} defines the Kantorovich version of OT. For measures in $\mathcal{P}_2(\R^d)$, the infimum is achieved (\ie, the minimal coupling exists) and the minimiser is unique. Moreover, for $\mathcal{P}_2(\R^d)$ the optimal coupling has the form $\gamma^* = (\Id, T)_{\#}\rho$ for some $T:\R^d\to\R^d$. This $T$ is the solution to the Monge OT problem, which appears in another (in this case, equivalent) definition of 2-Wasserstein distance:
 \begin{equation}\label{eq:wasserstein}
    \sqW{\rho}{\mu} = \inf_{T : T_{\#}\rho = \mu}
    \int
    |T(x) - x|^2\,\rho(x)\,\dd x
    .  
 \end{equation}
Yet a third useful definition of 2-Wasserstein distance, which we use to prove \Cref{appxlemma:sqw-kl-bound}, is due to Benamou-Brenier \citep[Equation~8.0.3]{ambrosio2005gradient} and employs the dynamic formulation of optimal transport:
\begin{equation}\label{eq:benamou-brenier}
    \sqW{\rho}{\mu} = \inf 
        \int_0^1\int |v_t|^2\rho_t(x)\,\dd x\,\dd t,
\end{equation}
where the infimum is taken over time-dependent vector fields $v_t$ and densities $\rho_t$ satisfying the continuity equation $\pd{t}{\rho_t} + \dv(\rho_t\,v_t)=0$ and boundary conditions $\rho_0=\rho,\rho_1=\mu$.

In any of these definitions, $\mathcal{W}_2$ satisfies the triangle inequality, meaning that 2-Wasserstein is a distance metric in $\mathcal{P}_2(\R^d)$. 

\subsection{Proofs}\label{subsec:proofs}

We first recall some standard definitions and results in Wasserstein geometry (see, \eg, \citet{santambrogio2015optimal}), then prove the results from the main text.

\begin{definition}[Geodesic convexity]\label{def:geodesic-convexity}
A functional $\mathcal{J}(\rho):\mathcal{P}_2(\R^d)\to\R$ is said to be 
\begin{itemize}[left=0pt]
    \item \textbf{geodesically convex} if, for every $\rho_0,\rho_1\in\mathcal{P}_2(\R^d)$, when $\rho_t$ is the geodesic between $\rho_0$ and $\rho_1$,
    \begin{equation}
        \forall t\in[0, 1]\quad
        \mathcal{J}(\rho_t) \leq t\mathcal{J}(\rho_1) + (1 - t)\mathcal{J}(\rho_0); 
    \end{equation}
    \item \textbf{$\lambda$-strongly geodesically convex} (for some $\lambda>0$) if, for every $\rho_0,\rho_1\in\mathcal{P}_2(\R^d)$, when $\rho_t$ is the geodesic between $\rho_0$ and $\rho_1$,
    \begin{equation}
        \forall t\in[0, 1]
        \quad\mathcal{J}(\rho_t) \leq t\mathcal{J}(\rho_1) + (1 - t)\mathcal{J}(\rho_0) 
        - \frac{\lambda}{2} t(1-t)\sqW{\rho_0}{\rho_1}.
    \end{equation}
\end{itemize}
\end{definition}

\printProofs

\section{Extended practical considerations}
\label{apx:ext-practical-considerations}
This appendix details the main design choices of \osis. It
expands on the kernel, loss, tempering, and inference
choices in \cref{sec:practical-considerations}. This section aims
to provide a general overview. 
Specific details about the experiments are provided in
\cref{apdx:experimental_setup}.

\subsection{Choice of MCMC kernel}\label{subsec:kernel-options}
We present here various methods to update the generated samples towards the target $\pi$. They often consist of a $\pi$-invariant MCMC kernel, but other methods, such as Stein variational gradient descent \citep{liu2016stein}, can also be investigated. In this subsection, $q(y\mid x)$ denotes a proposal density. We describe kernels for a fixed target $\pi$. Under an annealing schedule, the same constructions use $\pi_{\beta_t}$ instead.

\paragraph{Metropolis-Hastings.} The Metropolis-Hastings \citep[MH][]{metropolis1953equation, hastings1970monte} kernel modifies a continuous full-support proposal to be $\pi$-invariant by mixing it with a rejection kernel. To be precise, given a proposal kernel $q$, define
\begin{equation}
    \label{eq:mh_kernel}
    K(y\mid x) = \alpha(x,y)q(y\mid x) + (1-A(x))\delta_{x}(y),
\end{equation}
where $\delta_x$ is a Dirac delta at $x$ and
\begin{equation}\label{eq:MH_ratio}
    \alpha(x,y) = \min\left(1,\frac{\pi(y)q(x\mid y)}{\pi(x)q(y\mid x)}\right),\quad
    A(x)=\int\alpha(x,y)q(y\mid x)\,\dd y.
\end{equation}
By construction, $K$ is $\pi$-invariant (in particular, if $q$ is reversible with respect to $\pi$, then $A(x)=1$ for $\pi$-almost every $x$ and $K=q$). However, the choice of $q$ greatly affects the rate of convergence to $\pi$ of a Markov chain with transition operator $K$.

\paragraph{Random-walk proposals.} Random-walk Metropolis (RWM) is a commonly used form of MH. A standard variant takes $q$ to be the kenrel of a Gaussian randmo walk: $q(y\mid x) = \mathcal{N}(y; x,\sigma^2 I)$ for some chosen variance $\sigma^2$.
The proposal is symmetric, so its acceptance probability simplifies to $\min(1,\pi(y)/\pi(x))$. RWM requires target energy evaluations but no target energy gradients.

\paragraph{Langevin-based proposals.} One can make use of target gradient information when defining a proposal distribution. Metropolis-Adjusted Langevin \citep[MALA;][]{rossky1978brownian, besag1994mala, roberts1996exponential} is MH based on a proposal $q$ that is an Euler-Maruyama integration step of a Langevin SDE:
\begin{equation*}
    q(y\mid x)=\mathcal{N}\left(y;x-\delta\nabla\mathcal{E}(x),2\delta I_d\right)
\end{equation*}
for some chosen step size $\delta>0$.
At inverse temperature $\beta_t$, the drift becomes $-\delta\beta_t\nabla\mathcal E(x)$. Unlike RWM, this proposal is generally asymmetric, so both proposal-density terms in \eqref{eq:MH_ratio} must be retained. Omitting the accept/reject step gives the unadjusted Langevin algorithm (ULA), which does not in general preserve the target at a finite step size \citep{roberts1996exponential}. Neither a MALA step nor a ULA step is the exact Langevin semigroup transition $K^h$ used in \cref{sec:theory}.

\paragraph{Higher-order splitting schemes.}
Splitting integrators provide further discretisations of Langevin dynamics and can reduce finite-step sampling bias \citep{leimkuhler2013rational}.
The idea is to divide the dynamics into simpler parts that can each be integrated exactly, then compose their updates. We consider the matrix formulation of the underdamped Langevin targeting $\pi(x)\mathcal N(v;0,I_d)$, where $\pi\propto e^{-\mathcal E}$:

\begin{align}
\begin{bmatrix}
\dd X_s\\
\dd V_s
\end{bmatrix}
=
\underbrace{
\begin{bmatrix}
V_s\\
0
\end{bmatrix}\dd s
}_{\mathrm A}
+
\underbrace{
\begin{bmatrix}
0\\
-\nabla\mathcal E(X_s)
\end{bmatrix}\dd s
}_{\mathrm B}
+
\underbrace{
\begin{bmatrix}
0\\
-\zeta V_s\,\dd s+\sqrt{2\zeta}\,\dd B_s
\end{bmatrix}
}_{\mathrm O},
\label{eq:obabo-sde}
\end{align}
where $\zeta>0$ is the friction coefficient and $v\in\R^d$ is the auxiliary velocity variable. The underdamped Langevin dynamics is split into three parts:
free transport of the position at fixed velocity (A),
force-driven evolution of the velocity at fixed position (B),
and Ornstein-Uhlenbeck evolution of the velocity,
combining friction and Gaussian noise at fixed position (O). Their exact updates over a duration $t\ge0$ are
\begin{align}
\mathrm A_t:\quad &(x,v)\longmapsto(x+tv,v),\\
\mathrm B_t:\quad &(x,v)\longmapsto(x,v-t\nabla\mathcal E(x)),\\
\mathrm O_t:\quad &(x,v)\longmapsto
\left(x,e^{-\zeta t}v+\sqrt{1-e^{-2\zeta t}}\,\xi\right),
\qquad \xi\sim\mathcal N(0,I_d).
\label{eq:langevin-split-updates}
\end{align}
These exact substeps generally do not commute, so their composition provides a specific approximation of the full dynamics \citep[Section~IV]{leimkuhler2013robust}. For example, the chronological order of a BAOAB step of size $\delta$ is
\begin{equation}
\mathrm B_{\delta/2}\ \longrightarrow\
\mathrm A_{\delta/2}\ \longrightarrow\
\mathrm O_\delta\ \longrightarrow\
\mathrm A_{\delta/2}\ \longrightarrow\
\mathrm B_{\delta/2}.
\label{eq:baoab-composition}
\end{equation}
The letters specify the order of the updates; each repeated letter here uses a half step. Each update acts on the current state, with independent noise for each O substep. 
Symmetric Strang composition (\ie, letters forming a palindrome) give second-order weak accuracy under appropriate conditions \citep{strang1968construction}.

\paragraph{HMC.}
Hamiltonian Monte Carlo introduces auxiliary momentum variables and uses Hamiltonian trajectories, with a correction for numerical integration error, to construct a target-invariant transition \citep{neal2011mcmc}. 

\paragraph{Stein variational gradient descent.}
Stein variational gradient descent (SVGD) transports a set
of interacting particles using a deterministic, kernel-based
field \citep{liu2016stein}. For a differentiable
positive-definite kernel $k$ and step size $\delta>0$,
its update for particles $\{x_i\}_{i=1}^{B}$ is
\begin{equation}
y_i
=
x_i+\frac{\delta}{B}\sum_{j=1}^{B}
\left[
-k(x_j,x_i)\nabla\mathcal E(x_j)
+\nabla_{x_j} k(x_j,x_i)
\right],
\qquad i=1,\ldots,B.
\label{eq:svgd-particle-update}
\end{equation}

\paragraph{Multiple-step kernels.} If $K$ is a $\pi$-invariant kernel, so is $K^n$ for $n\geq1$. Thus one can construct several-step kernels by iterating the primitive ones defined above, as we do in the experiments.

\paragraph{Acceptance-targeting step adaptation.} 
We adapt the step size $\delta$ of random-walk or Langevin MCMC during training to maintain a target acceptance rate $\alpha^{\star}$ \citep{andrieu2008tutorial}. The update is log-space stochastic approximation,
$$
  \log h_t \leftarrow \log h_t + \eta_{\mathrm{adapt}}(\hat\alpha-\alpha^\star),
$$
where $\hat\alpha$ is the batch mean acceptance and $h_t\coloneqq \frac{\delta}{\beta_t}$ is the temperature-rescaled step size. We then set the sampler’s step size to $\delta_t=h_t\beta_t$ and clip it to $[\delta_{\min},\delta_{\max}]$.

\subsection{Choice of divergence}\label{subsec:div-options}

To compute the distance between two distributions, we rely on sample-based distribution metrics. In practice, these distances are computed using samples from both distributions.

\paragraph{Maximum mean discrepancy.}

One possible way to compute the distance between two distributions is use maximum mean discrepancy \citep[MMD;][]{gretton2013two}. Given a positive definite kernel function, such as 
$k(x, y) = \exp\left( -\frac{\| x - y \|}{h} \right)$, where $h$ is a positive bandwidth and $\|\cdot\|$ is $L_2$ distance, the MMD between two distributions $\mu_1$ and $\mu_2$ is defined as:
\begin{align}
    \text{MMD}^2(\mu_1,\mu_2) &\coloneqq 
    \mathbb{E}_{x,x'\sim\mu_1} [k(x, x')]
    + \mathbb{E}_{y,y'\sim\mu_2} [k(y, y')]\nonumber\\
    &\phantom{=\;}- 2\mathbb{E}_{x\sim\mu_1,y\sim\mu_2} [k(x, y)] ,\label{eq:mmd}
\end{align}
which can be estimated by Monte Carlo sampling.

\paragraph{Squared 2-Wasserstein distance.}
For probability distributions $\mu_1$ and $\mu_2$ on $\R^d$ with finite second moments, recall from \eqref{eq:wasserstein} that the squared 2-Wasserstein distance is an optimal transport cost
\begin{align}
  \sqW{\mu_1}{\mu_2} \coloneqq  
  \inf_{\gamma\in\Gamma(\mu_1,\mu_2)}\int_{\R^d\times\R^d}|x-y|^2\,\dd\gamma(x,y),
  \label{eq:w2-def}
\end{align}
where $\Gamma(\mu_1,\mu_2)$ contains joint distributions over  $\R^d\times\R^d$ whose marginals are $\mu_1$ and $\mu_2$.The  $\mathcal{W}_2$ distance between empirical approximations to $\mu_1$ and $\mu_2$ is known as the minibatch 2-Wasserstein, and it converges to \eqref{eq:w2-def} as the number of samples increases \citep{fatras2021minibatch}.

\paragraph{Sliced Wasserstein distance.}
The sliced Wasserstein distance
\citep[SWD;][]{bonneel2015sliced, kolouri2019generalized,
nadjahi2020statistical} is a tractable surrogate that computes
OT distances using $1$-dimensional projections.
It is defined by
\begin{equation}\label{eq:swd-def}
    \mathbb{S}_2^2(\mu_1,\mu_2) \coloneqq
    \mathbb{E}_\omega\left[
    \sqW{P_{\omega\,\#}\mu_1}{P_{\omega\,\#}\mu_2}\right],
\end{equation}
where the expectation is taken over $\omega$ uniform on the unit sphere in $\R^d$.
In practice, the expectation over the projection direction $\omega$ is estimated with
random projections. The quantity inside the expectation is a 1D OT distance between empirical distributions and is thus easily computed.

\paragraph{Sinkhorn divergence.}
Entropic regularisation is introduced to the transport problem \eqref{eq:w2-def} and provides a smoother optimisation problem that can be solved using Sinkhorn iterations. For regularisation strength $\varepsilon>0$, define \citep{cuturi2013sinkhorn}
\begin{equation}
\operatorname{OT}_\varepsilon(\mu_1,\mu_2)
=\inf_{\gamma\in\Gamma(\mu_1,\mu_2)}
\left[\int|x-y|^2\,\dd\gamma(x,y)
+\varepsilon\KL{\gamma}{\mu_1\otimes\mu_2}\right].
\label{eq:practical-entropic-ot}
\end{equation}
The corresponding debiased Sinkhorn divergence subtracts the two self-interaction terms,
\begin{equation}
\operatorname{S}_\varepsilon(\mu_1,\mu_2)
=\operatorname{OT}_\varepsilon(\mu_1,\mu_2)
-\tfrac12\operatorname{OT}_\varepsilon(\mu_1,\mu_1)-\tfrac12\operatorname{OT}_\varepsilon(\mu_2,\mu_2),
\label{eq:practical-sinkhorn}
\end{equation}
so that $\operatorname{S}_\varepsilon(\mu,\mu)=0$. 
For empirical measures and the squared-Euclidean cost used here,
$\operatorname{S}_\varepsilon(\mu_1,\mu_2)$ approaches
$\sqW{\mu_1}{\mu_2}$ as $\varepsilon\downarrow0$.
As $\varepsilon\to\infty$, it approaches the squared distance
between the means of $\mu_1$ and $\mu_2$, equivalently, the
squared MMD associated with the linear kernel
$k(x,y)=\langle x,y\rangle$
\citep[Section~1.2]{feydy2019interpolating}.

\section{Detailed experiment setup}
\label{apdx:experimental_setup}

\subsection{Synthetic targets}

\paragraph{Setup.}
We train \osis and the baselines for five random seeds. After training, we draw five independent batches of $10{,}000$ samples with different seeds at the selected checkpoint: the final training iteration for \osis{} and the checkpoint with
the highest ELBO for baselines, following
\citet{blessing2024beyond}.

Each sampler's evaluation batch is compared with the same batch of $10{,}000$ reference
samples for all samplers. We average the results within each training seed, then report the mean and sample standard deviation across the five training-seed means. SMC uses five repetition groups instead of training
seeds. The reference rows provide a floor on each metric by comparing two independent target batches of samples.
We report empirical $\mathcal W_2^2$ with unnormalised squared Euclidean cost, computed using POT \citep{flamary2021pot} with convergence checks. MMD corresponds to the square root of the biased Gaussian kernel MMD-squared estimate, with bandwidth fixed per target to the median pairwise distance in an independent $10{,}000$-sample batch from the target distribution.

\paragraph{Baselines.}
We retrain FAB~\citep{midgley2022flow}, PIS~\citep{zhang2021path},
and DIS~\citep{berner2024optimal}, and run SMC~\citep{del2006sequential}
using the reference implementation of \citet{blessing2024beyond}.
FAB uses an 8-layer Real-NVP flow with $[64,64]$ conditioner MLPs,
12 intermediate distributions, 10 HMC transitions per
intermediate distribution, a replay buffer, and $\alpha$-divergence loss
with $\alpha=2$. It trains for 3,000 iterations with four replay
updates per iteration, at batch size 2,000 (1,000 for both NICE targets).

PIS and DIS use the PISGRADNet with width-64 MLPs and Fourier time features, trained for 40,000 updates with a batch size of 2,000 and 128 integration steps. For other hyperparameters, we use those reported in \citet{blessing2024beyond}. For ManyWell $d=32,64$ and 40GMM $d=10$, since the paper does not provide default hyperparameters, we performed a grid search over the key hyperparameters, as provided in \cref{tab:baseline-hparam-search}, and selected the one with the best ELBO. SMC uses 2,000 particles, 10 HMC transitions per
annealing interval, and resampling threshold of $0.3$. It uses 256 temperatures for GMM-10D and ManyWell (dimension $32$ and $64$) targets, and 128 otherwise. 

For TB$+$MALA \citep{sendera2024improved} and SubTB$+$SMC \citep{choi2025reinforced}, we use the original implementation of \citet{choi2025reinforced} and reimplement TB$+$MALA within the same codebase to simplify benchmarking. While the reimplemented TB$+$MALA is not identical to the original from \citet{sendera2024improved}, we ensure that the algorithm performs at least as well as, or better than, the original on synthetic targets. For these two baselines, we set the network architectures and common hyperparameters (learning rate and initial std.) to match those of DIS, while fixing algorithm-specific hyperparameters as reported in the original papers.

\begin{table}[t]
\centering
\caption{Baseline hyperparameter search grids.}\vspace*{-1em}
\label{tab:baseline-hparam-search}
\begin{tabular}{@{}llcc}
\toprule
Method & Parameter & GMM ($d=10$) & ManyWell ($d=32,64$) \\
\midrule
FAB / PIS / DIS & Learning rate
& \multicolumn{2}{c}{$\{10^{-5},10^{-4},5\cdot10^{-4},10^{-3}\}$} \\
\midrule
FAB & Base scale
& $\{20,40,80\}$ & $\{1,2,4\}$ \\
DIS & Initial std.
& $\{20,40,80\}$ & $\{1,2,4\}$ \\
PIS & Max.\ diffusion amplitude
& $\{80,160,320\}$ & $\{2,4,8\}$ \\
\midrule
SMC & Initial std.
& $\{20,40,80\}$ & $\{1,2,4\}$ \\
& HMC step size
& $\{0.1,0.2\}$ & $\{0.05,0.1\}$ \\
& Number of temperatures
& $\{128,256\}$ & $\{128,256\}$ \\
\bottomrule
\end{tabular}
\end{table}

\paragraph{\osis{}.}
For the main evaluation study \cref{tab:main-results,tab:main-results-nice}, \osis was trained using a squared SWD~\citep{bonneel2015sliced} with $2048$ projections, AdamW with weight decay $10^{-2}$, batch size of $4096$,
$T=10{,}000$ training updates, $20$ steps of MCMC refinement and $M=50$ inner gradient iterations \cref{alg:osis-div-ful}. All the other experiments were run for $M=1$.

We parameterise the function $G_\theta$ with a ResNet~\citep{he2016deep} architecture for
synthetic targets and a small UNet~\citep{ronneberger2015u} for the NICE targets.
The temperature follows the schedule of \citet{fogliani2026annealing}: at time $t$, the inverse temperature is
\begin{align*}
\beta(t)=\beta_0^{\max(0,\,1-t/(0.8T))},
\end{align*}
so the last $20\%$ of training uses $\beta=1$. Funnel keeps $\beta=1$ during its entire training. Additional hyperparameters are reported in \cref{tab:osis-hparams}.

\begin{table}[t]
\centering
\caption{\osis hyperparameters. \textsc{ResNet}(hidden, blocks) and
\textsc{UNet}(base channels, attention heads).
The temperature schedule is as defined in \cref{apdx:experimental_setup}.}\vspace*{-1em}
\label{tab:osis-hparams}
\begin{tabular}{@{}lllccc}
\toprule
Target & Loss & Model & lr & $\tau_{\mathrm{start}}$ & target accept. \\
\midrule
40GMM ($d=2$) & SWD & \textsc{ResNet}(256, 3) & $3\cdot10^{-3}$ & 128 & 0.3 \\
40GMM ($d=10$) & SWD & \textsc{ResNet}(256, 3) & $3\cdot10^{-3}$ & 256 & 0.3 \\
ManyWell ($d=32$) & SWD & \textsc{ResNet}(256, 3) & $3\cdot10^{-3}$ & 256 & 0.90 \\
ManyWell ($d=64$) & SWD & \textsc{ResNet}(256, 3) & $3\cdot10^{-3}$ & 16 & 0.574 \\
Funnel ($d=10$) & SWD & \textsc{ResNet}(128, 2) & $3\cdot10^{-5}$ & 1 & 0.90 \\
NICE-MNIST ($d=196$) & SWD & \textsc{UNet}(16, 4) & $1\cdot10^{-3}$ & 32 & 0.574 \\
NICE-FMNIST ($d=784$) & SWD & \textsc{UNet}(16, 4) & $1\cdot10^{-4}$ & 32 & 0.574 \\
\bottomrule
\end{tabular}
\end{table}

\paragraph{Kernel comparison.}
For this study the model is kept fixed across kernels as described above. Each kernel uses $20$ refinement steps. Other parameters are tuned per kernel and presented in Table~\ref{tab:kernel-comparison-detail}. MALA in this study uses $M=1$ inner iterations (see \cref{alg:osis-div-ful}), so its parameters differ from \cref{tab:osis-hparams}.

\begin{table*}[t]
\centering
\caption{Selected hyperparameters for ManyWell ($d=32$).}\vspace*{-1em}
\label{tab:kernel-comparison-detail}
\resizebox{\linewidth}{!}{%
\begin{tabular}{@{}lrrrrrrrrrr}
\toprule
Kernel & \shortstack{Learning\\rate\vphantom{d}}
& \shortstack{Initial\vphantom{g}\\temperature\vphantom{d}}
& \shortstack{Initial proposal\\standard\vphantom{g}\\deviation}
& \shortstack{Target\vphantom{p}\\acceptance\vphantom{d}\vphantom{g}\\rate\vphantom{d}}
& \shortstack{ULA\vphantom{g}\\drift step}
& \shortstack{Integration\\step size}
& Friction & \shortstack{Leapfrog\\steps\vphantom{d}}
& \shortstack{SVGD\vphantom{g}\\update step}
& \shortstack{RBF kernel\vphantom{g}\\variance} \\
\midrule
RWMH & $10^{-3}$ & 4 & 1 & 0.5 & --- & --- & --- & --- & --- & --- \\
MALA & $10^{-3}$ & 16 & --- & 0.90 & --- & --- & --- & --- & --- & --- \\
ULA & $10^{-3}$ & 16 & --- & --- & 0.008 & --- & --- & --- & --- & --- \\
BAOAB & $1.9\cdot10^{-5}$ & 2 & --- & --- & --- & 0.0025 & 2 & --- & --- & --- \\
OBABO & $10^{-3}$ & 16 & --- & --- & --- & 0.006 & 2 & --- & --- & --- \\
HMC & $1.7\cdot10^{-3}$ & 16 & --- & --- & --- & 0.019 & --- & 10 & --- & --- \\
SVGD & $10^{-3}$ & 16 & --- & --- & --- & --- & --- & --- & $1.6\cdot10^{-4}$ & 2 \\
\bottomrule
\end{tabular}%
}
\end{table*}

\paragraph{Parameter counts.}
Trainable parameter counts are reported in Table~\ref{tab:param-counts}. SMC is non-amortised
and has no learned parameters. PIS, DIS, and TB$+$MALA share the same backbone and
parameter count. \osis uses a \textsc{ResNet} on the synthetic
targets and a \textsc{UNet} on the NICE targets.

\begin{table}[t]
\centering
\caption{Number of trainable parameters in the benchmark study ($\times10^3$).}\vspace*{-1em}
\label{tab:param-counts}
\begin{tabular}{@{}lrrrrrrr}
\toprule
Target $\rightarrow$ & \multicolumn{2}{c}{40GMM} & \multicolumn{2}{c}{ManyWell} & Funnel & \multicolumn{2}{c}{NICE} \\
\cmidrule(lr){2-3}\cmidrule(lr){4-5}\cmidrule(lr){6-6}\cmidrule(lr){7-8}
Method $\downarrow$ & $d=2$ & $d=10$ & $d=32$ & $d=64$ & $d=10$ & MNIST & FMNIST \\
\midrule
SMC & --- & --- & --- & --- & --- & --- & --- \\
FAB & 70.7 & 83.1 & 117.2 & 166.9 & 83.1 & 371.8 & 1284.4 \\
PIS / DIS & 37.8 & 39.3 & 43.6 & 49.8 & 39.3 & 75.4 & 189.5 \\
\osis{} & 198.7 & 202.8 & 214.0 & 230.5 & 35.7 & 227.9 & 689.9 \\
\bottomrule
\end{tabular}
\end{table}

\paragraph{Sampling time.}
Figure~\ref{fig:efficiency} provides the median time over 250 runs for a batch of 2,000 samples on one NVIDIA GH200 GPU, excluding compilation and warm-up. The mean $\mathcal W^2_2$ metric is computed independently following the previously described 10,000-sample protocol. \osis, TB$+$MALA, and SubTB$+$SMC use compiled PyTorch, while FAB, PIS, DIS and SMC use native JAX JIT.

\subsection{Outsourced sampling}

\paragraph{Target densities.}
For MNIST we use a VAE \citep{kingma2013auto} which we train ourselves. Then, we use VAE decoder with a $16$-dimensional latent,  and a Bernoulli likelihood, trained for $100$ epochs. For the reward function, we train a CNN classifier $p(c\mid x)$. The reward is defined as  $\log r(z)=\log\sum_c \tilde w_c, p(c\mid x(z))$, and the target log-density is $\log\pi(z)=\log r(z)-\tfrac12| z|^2$. 
Both networks remain frozen during the training of \osis. 

For the experiments on FFHQ-64, we use a pretrained frozen R3GAN \citep{huang2024gan} generator. The reward is ImageReward \citep{xu2023imagereward}. The target log-density is defined as $\log\pi(z)=3.0\cdot\mathrm{IR}(G(z), \text{caption}) - \tfrac12|z|^2$. Reward and gradient passes run \texttt{bfloat16}.

\paragraph{Model architecture for $G_\theta$.} In all our experiments we use a ResNet architecture with MLP layers. The inputs are first projected to a $256$-dimensional latent, then follow 3 residual  blocks of the form $x \mapsto x + \mathrm{SiLU}(Wx+b),\,W \in \mathbb{R}^{256\times256}$, then the outputs are projected to the output dimension.

\paragraph{Training.} We train \osis with $1200$ outer steps and $50$ inner steps, following the procedure presented in \Cref{alg:osis-div-ful}. We use an empirical approximation of 
OT loss (\Cref{subsec:div-options}) and use  MALA with adaptive step size (\Cref{subsec:kernel-options}) as the MCMC algorithm, which we run for $20$ steps. For the initial $1000$ outer iterations we train the algorithm on a tempered target, the temperature follows a cosine schedule with initial temperature $2.0$ and final temperature $1.0$. Next, we train \osis for $200$ more iterations keeping the temperature fixed at $1.0$. We use the AdamW \citep{loshchilov2017decoupled} optimiser 
with learning rate $0.001$ and no weight decay. 

\paragraph{Evaluations.} Each method in \Cref{tab:outsourced} is evaluated using  $N=10{,}000$. The reference is rejection sampling from the same tilted density. Since ImagerReward is unbounded by default, we bound it with a constant to perform rejection sampling. $\mathbb{E}[\log r(x)]$ corresponds to the averaged ImageReward score computed with $10,000$ samples. FID is computed against the rejection-sampling reference. We compute FID score using \texttt{torchmetrics}.

\section{Additional experiment results}

\subsection{Extended evaluation of \osis}
\label{apdx:extended_eval_osis}

In this section, we report results for \osis with OBABO kernels on the main synthetic targets. This provides further evidence on whether splitting schemes can improve \osis performance \cref{tab:obabo-osis,tab:obabo-mnist-fmnist}. Overall, the benefits depend on the training objective: with OT, OBABO achieves lower mean squared 2-Wasserstein distance and MMD than MALA on all evaluated targets, whereas with SWD, the results are mixed. These experiments compare equal numbers of \osis training steps, which does not necessarily imply equal computational cost. Each OBABO step requires two energy-gradient evaluations, whereas each MALA step requires two energy-gradient evaluations and two additional energy-value evaluations for the Metropolis-Hastings correction.

\begin{table*}[t]
\centering
\caption{Sampler comparison on GMM, Funnel, and ManyWell targets. OSIS rows specify the training kernel and loss. This table was computed using $M=1$ inner iteration (see \cref{alg:osis-div-ful}).}\vspace*{-1em}
\label{tab:obabo-osis}
\resizebox{\linewidth}{!}{%
\begin{tabular}{@{}lrrrrrrrrrr}
\toprule
Target $\rightarrow$
 & \multicolumn{2}{c}{40GMM ($d=2$)}
 & \multicolumn{2}{c}{40GMM ($d=10$)}
 & \multicolumn{2}{c}{Funnel ($d=10$)}
 & \multicolumn{2}{c}{ManyWell ($d=32$)}
 & \multicolumn{2}{c}{ManyWell ($d=64$)} \\
\cmidrule(lr){2-3} \cmidrule(lr){4-5} \cmidrule(lr){6-7}
\cmidrule(lr){8-9} \cmidrule(lr){10-11}
Method $\downarrow$ Metric $\rightarrow$
 & \multicolumn{1}{c}{$\mathcal{W}_2^2$} & \multicolumn{1}{c}{MMD ($\times10^{-3}$)}
 & \multicolumn{1}{c}{$\mathcal{W}_2^2$} & \multicolumn{1}{c}{MMD ($\times10^{-3}$)}
 & \multicolumn{1}{c}{$\mathcal{W}_2^2$} & \multicolumn{1}{c}{MMD ($\times10^{-3}$)}
 & \multicolumn{1}{c}{$\mathcal{W}_2^2$} & \multicolumn{1}{c}{MMD ($\times10^{-3}$)}
 & \multicolumn{1}{c}{$\mathcal{W}_2^2$} & \multicolumn{1}{c}{MMD ($\times10^{-3}$)} \\
\midrule
Reference
 & $7.68\std{3.22}$ & $7.17\std{1.51}$
 & $202.81\std{16.77}$ & $7.98\std{0.78}$
 & $87.60\std{1.20}$ & $9.22\std{1.54}$
 & $23.61\std{0.07}$ & $8.75\std{0.72}$
 & $78.36\std{0.11}$ & $9.58\std{0.65}$ \\
\midrule
\rowcolor{osisgreen}
\osis{} (OBABO, OT)
 & $81.85\std{31.72}$ & $58.06\std{15.95}$
 & $2359.76\std{535.07}$ & $123.10\std{39.78}$
 & $189.86\std{0.33}$ & $246.80\std{0.72}$
 & $35.54\std{0.01}$ & $255.33\std{0.34}$
 & $78.70\std{0.04}$ & $274.45\std{0.43}$ \\
\rowcolor{osisgreen}
\osis{} (OBABO, SWD)
 & $848.79\std{388.69}$ & $378.69\std{132.12}$
 & $7465.66\std{7.91}$ & $477.83\std{0.51}$
 & $99.37\std{2.09}$ & $57.69\std{8.07}$
 & $28.77\std{0.53}$ & $85.55\std{8.86}$
 & $91.24\std{2.01}$ & $170.13\std{9.52}$ \\
\rowcolor{osisgreen}
\osis{} (MALA, OT)
 & $111.24\std{94.65}$ & $83.77\std{39.73}$
 & $2913.22\std{489.84}$ & $140.34\std{31.18}$
 & $243.15\std{3.27}$ & $474.04\std{16.40}$
 & $36.60\std{0.02}$ & $276.63\std{0.29}$
 & $81.37\std{0.10}$ & $293.66\std{0.61}$ \\
\rowcolor{osisgreen}
\osis{} (MALA, SWD)
 & $553.89\std{305.14}$ & $277.34\std{116.75}$
 & $9854.75\std{1328.11}$ & $482.27\std{132.16}$
 & $89.41\std{0.53}$ & $45.32\std{3.21}$
 & $25.51\std{0.43}$ & $48.57\std{7.55}$
 & $89.93\std{14.74}$ & $159.27\std{20.76}$ \\
\midrule
FAB \citep{midgley2022flow}
 & $72.03\std{19.29}$ & $46.42\std{6.32}$
 & $6257.04\std{455.10}$ & $380.98\std{45.19}$
 & $582.97\std{233.17}$ & $11.87\std{1.74}$
 & $25.52\std{0.30}$ & $36.63\std{5.56}$
 & $132.55\std{0.48}$ & $299.11\std{1.51}$ \\
FAB (flow only) \citep{midgley2022flow}
 & $296.01\std{333.93}$ & $81.05\std{38.73}$
 & $6217.94\std{472.86}$ & $382.64\std{44.65}$
 & $583.48\std{233.71}$ & $12.94\std{2.34}$
 & $27.10\std{0.46}$ & $47.95\std{4.56}$
 & $125.58\std{0.04}$ & $401.28\std{0.64}$ \\
SMC \citep{del2006sequential}
 & $14.67\std{1.79}$ & $16.85\std{2.11}$
 & $4362.81\std{139.06}$ & $208.62\std{11.73}$
 & $149.56\std{1.37}$ & $125.72\std{8.99}$
 & $26.04\std{0.16}$ & $22.48\std{1.10}$
 & $85.76\std{0.29}$ & $42.63\std{1.36}$ \\
DIS \citep{berner2024optimal}
 & $29.26\std{7.00}$ & $34.71\std{10.99}$
 & $1650.04\std{60.09}$ & $85.37\std{5.26}$
 & $122.58\std{21.01}$ & $214.06\std{103.49}$
 & $36.44\std{2.44}$ & $162.78\std{20.94}$
 & $100.34\std{3.52}$ & $162.04\std{18.66}$ \\
PIS \citep{zhang2021path}
 & $40.35\std{25.29}$ & $34.40\std{12.85}$
 & $2702.24\std{66.69}$ & $142.54\std{5.24}$
 & $165.77\std{1.72}$ & $194.21\std{3.72}$
 & $35.47\std{0.10}$ & $227.64\std{1.74}$
 & $83.39\std{0.13}$ & $230.37\std{1.05}$ \\
\bottomrule
\end{tabular}%
}
\end{table*}

\begin{table*}[t]
\centering
\caption{Sampler comparison on high-dimensional targets. OSIS rows specify the training kernel and loss. This table was computed using $M=1$ inner iteration (see \cref{alg:osis-div-ful}).}\vspace*{-1em}
\label{tab:obabo-mnist-fmnist}
\resizebox{1\linewidth}{!}{
\begin{tabular}{@{}lrrrr}
\toprule
 Target $\rightarrow$ & \multicolumn{2}{c}{NICE-MNIST ($d=196$)}
 & \multicolumn{2}{c}{NICE-FMNIST ($d=784$)} \\
\cmidrule(lr){2-3} \cmidrule(lr){4-5}
Method $\downarrow$ Metric $\rightarrow$
 & \multicolumn{1}{c}{$\mathcal{W}_2^2$} & \multicolumn{1}{c}{MMD ($\times10^{-3}$)}
 & \multicolumn{1}{c}{$\mathcal{W}_2^2$} & \multicolumn{1}{c}{MMD ($\times10^{-3}$)} \\
\midrule
Reference
 & $129.56\std{0.22}$ & $8.75\std{0.08}$
 & $549.42\std{1.35}$ & $9.24\std{1.09}$ \\
\midrule
\rowcolor{osisgreen}
\osis{} (OBABO, OT)
 & $270.02\std{4.60}$ & $428.07\std{10.29}$
 & $1631.93\std{63.53}$ & $412.35\std{14.51}$ \\
\rowcolor{osisgreen}
\osis{} (OBABO, SWD)
 & $183.18\std{5.45}$ & $122.34\std{27.80}$
 & $2269.09\std{475.05}$ & $410.57\std{89.89}$ \\
\rowcolor{osisgreen}
\osis{} (MALA, OT)
 & $277.40\std{8.45}$ & $462.69\std{18.73}$
 & $1776.79\std{82.07}$ & $450.47\std{21.16}$ \\
\rowcolor{osisgreen}
\osis{} (MALA, SWD)
 & $209.74\std{39.21}$ & $248.96\std{125.42}$
 & $2506.97\std{29.95}$ & $507.93\std{31.28}$ \\
\midrule
FAB \citep{midgley2022flow}
 & $143.83\std{3.10}$ & $80.83\std{14.39}$
 & $1746.47\std{335.35}$ & $438.79\std{71.94}$ \\
FAB (flow only) \citep{midgley2022flow}
 & $162.20\std{1.91}$ & $57.94\std{11.09}$
 & $1768.75\std{335.78}$ & $430.25\std{72.05}$ \\
SMC \citep{del2006sequential}
 & $268.32\std{4.16}$ & $246.17\std{13.54}$
 & $2856.36\std{15.26}$ & $629.18\std{3.24}$ \\
DIS \citep{berner2024optimal}
 & $307.73\std{55.89}$ & $208.21\std{22.97}$
 & $5321.68\std{2656.55}$ & $421.53\std{124.32}$ \\
PIS \citep{zhang2021path}
 & $237.63\std{3.08}$ & $225.89\std{16.69}$
 & $2793.73\std{19.11}$ & $290.08\std{4.09}$ \\
\bottomrule
\end{tabular}
}
\end{table*}

\subsection{\osis against its long run refinement}
In this section we compare \osis to its long run MCMC kernel. Both methods are run with the same MALA settings, number of MCMC steps and temperature annealing schedule. This table assesses how much of the refinement signal \osis captures relative to running the equally long rollout of its refinement kernel. We note that the MALA parameters used here come from \osis runs meaning, that they were  not tuned for \osis training and may not be optimal for direct MCMC sampling.

We observe that \osis has higher reported discrepancies than MALA on 40GMM, motivating further investigation of whether additional inner optimisation steps
help \osis track its refinement targets \cref{tab:main-results}.
On ManyWell, however, SWD-trained \osis outperforms its refinement kernel run for the same number of steps.
These results show effective amortisation of the refinement signal in our sampler on specific targets. 

\begin{table*}[h!]
\centering
\caption{Comparison of \osis+MALA with long-run MALA only.
\osis rows specify the training kernel and loss. MALA-only runs start from Gaussian noise. We compare one set of MALA samples against five reference sets and report the average and standard deviation. This table was computed using $M=1$ inner iteration (see \cref{alg:osis-div-ful}).}\vspace*{-1em}
\label{tab:mala-comparison-low-dimensional}
\resizebox{\linewidth}{!}{%
\begin{tabular}{lrrrrrrrrrr}
\toprule
 Target $\rightarrow$ & \multicolumn{2}{c}{40GMM ($d=2$)}
 & \multicolumn{2}{c}{40GMM ($d=10$)}
 & \multicolumn{2}{c}{Funnel ($d=10$)}
 & \multicolumn{2}{c}{ManyWell ($d=32$)}
 & \multicolumn{2}{c}{ManyWell ($d=64$)} \\
\cmidrule(lr){2-3} \cmidrule(lr){4-5} \cmidrule(lr){6-7}
\cmidrule(lr){8-9} \cmidrule(lr){10-11}
Method $\downarrow$ Metric $\rightarrow$
 & \multicolumn{1}{c}{$\mathcal{W}_2^2$} & \multicolumn{1}{c}{MMD ($\times10^{-3}$)}
 & \multicolumn{1}{c}{$\mathcal{W}_2^2$} & \multicolumn{1}{c}{MMD ($\times10^{-3}$)}
 & \multicolumn{1}{c}{$\mathcal{W}_2^2$} & \multicolumn{1}{c}{MMD ($\times10^{-3}$)}
 & \multicolumn{1}{c}{$\mathcal{W}_2^2$} & \multicolumn{1}{c}{MMD ($\times10^{-3}$)}
 & \multicolumn{1}{c}{$\mathcal{W}_2^2$} & \multicolumn{1}{c}{MMD ($\times10^{-3}$)} \\
\midrule
Reference
 & $7.68\std{3.22}$ & $7.17\std{1.51}$
 & $202.81\std{16.77}$ & $7.98\std{0.78}$
 & $87.60\std{1.20}$ & $9.22\std{1.54}$
 & $23.61\std{0.07}$ & $8.75\std{0.72}$
 & $78.36\std{0.11}$ & $9.58\std{0.65}$ \\
\midrule
\rowcolor{osisgreen}
\osis{} (MALA, OT)
 & $111.24\std{94.65}$ & $83.77\std{39.73}$
 & $2913.22\std{489.84}$ & $140.34\std{31.18}$
 & $243.15\std{3.27}$ & $474.04\std{16.40}$
 & $36.60\std{0.02}$ & $276.63\std{0.29}$
 & $81.37\std{0.10}$ & $293.66\std{0.61}$ \\
\rowcolor{osisgreen}
\osis{} (MALA, SWD)
 & $553.89\std{305.14}$ & $277.34\std{116.75}$
 & $9854.75\std{1328.11}$ & $482.27\std{132.16}$
 & $89.41\std{0.53}$ & $45.32\std{3.21}$
 & $25.51\std{0.43}$ & $48.57\std{7.55}$
 & $89.93\std{14.74}$ & $159.27\std{20.76}$ \\
\midrule
MALA only
 & $82.82\std{2.28}$ & $52.04\std{1.67}$
 & $200.14\std{12.26}$ & $8.58\std{0.62}$
 &  $90.71\std{3.41}$ & $22.47\std{3.52}$
 & $36.58\std{0.05}$ & $149.82\std{0.74}$
 & $104.71\std{0.11}$ & $166.18\std{0.86}$\\
\bottomrule
\end{tabular}%
}
\end{table*}

\clearpage\newpage

\subsection{Drift parameterisation of \osis}\label{sec:drift-experiments}

We evaluate whether \osis can benefit from a drift-based regression formulation motivated by Wasserstein gradient flows. This formulation expresses the generator update as a regression task towards samples displaced by an estimated drift field, with the regression targets held fixed during differentiation. For OT-based objectives, computing these targets still requires the corresponding transport information. The drift parameterisation yields lower mean values for both metrics on ManyWell ($d=32$). On 40GMM ($d=2$), both mean Sinkhorn divergence and mean MMD are higher, although the differences are within the scale of the reported variability.

\begin{table}[h!]
\centering
\caption{Comparison of discrepancy and drift objectives for \osis trained with the OBABO refinement kernel. Results are mean $\pm$ standard deviation over five training seeds, after averaging five sampling-seed evaluations per checkpoint. This table was computed using $M=1$ inner iteration (see \cref{alg:osis-div-ful}).}\vspace*{-1em}
\label{tab:drift-discrepancy-student-w2}
\providecommand{\std}[1]{\ensuremath{\,{\scriptstyle\pm #1}}}
\begin{tabular}{@{}lrrrr}
\toprule
Target $\rightarrow$
 & \multicolumn{2}{c}{40GMM ($d=2$) + OT}
 & \multicolumn{2}{c}{ManyWell ($d=32$) + SWD} \\
\cmidrule(lr){2-3} \cmidrule(lr){4-5}
Method $\downarrow$ Metric $\rightarrow$
 & $\mathcal{W}_2^2$ & MMD ($\times10^{-3}$)
 & $\mathcal{W}_2^2$ & MMD ($\times10^{-3}$) \\
\midrule
Reference
 & $7.68\std{3.22}$ & $7.17\std{1.51}$
 & $23.61\std{0.07}$ & $8.75\std{0.72}$ \\
\midrule
Discrepancy
 & $81.85\std{31.72}$ & $58.06\std{15.95}$
 & $28.77\std{0.53}$ & $85.55\std{8.86}$ \\
Drift
 & $95.93\std{46.34}$ & $77.97\std{35.87}$
 & $27.11\std{1.29}$ & $69.34\std{13.05}$ \\
\bottomrule
\end{tabular}
\end{table}

\clearpage\newpage
\subsection{Ablations}
\label{apx:osis-ablation}

We provide in this section detailed ablations of the different design choices of \osis.

\begin{table*}[h!]
\centering
\caption{
Ablation on ManyWell ($d=32$).
The default \osis{} training configuration uses 20-step MALA,
batch size $4096$, SWD-$2048$,
and $50,000$ training updates.
Results are mean $\pm$ standard deviation over five training seeds,
after averaging five sampling-seed evaluations per checkpoint. This table was computed using $M=1$ inner iteration (see \cref{alg:osis-div-ful}).
}\vspace*{-1em}
\label{tab:ablation_manywell32_student_reference_w2}
\begingroup
\begin{tabularx}{\textwidth}{@{}Yrr@{}}
\toprule
Configuration
& \multicolumn{1}{c}{$\mathcal{W}_2^2$ ($\downarrow$)}
& \multicolumn{1}{c}{MMD ($\times10^{-3}$)} \\
\midrule
\rowcolor{osisgreen}
\emph{Default} & $25.51\std{0.43}$ & $48.57\std{7.55}$ \\
\midrule
\multicolumn{3}{@{}l}{\emph{MCMC steps} (default: $20$)} \\
$1$ & $28.40\std{1.19}$ & $77.78\std{20.34}$ \\
$5$ & $30.17\std{1.32}$ & $109.10\std{29.69}$ \\
$10$ & $27.98\std{1.87}$ & $85.43\std{41.68}$ \\
$40$ & $25.30\std{0.55}$ & $51.75\std{12.59}$ \\
$80$ & $25.91\std{0.33}$ & $43.79\std{9.80}$ \\
\midrule
\multicolumn{3}{@{}l}{\emph{Batch size} (default: $4096$)} \\
$1024$ & $26.37\std{1.19}$ & $60.70\std{16.79}$ \\
$2048$ & $25.58\std{0.32}$ & $47.69\std{5.16}$ \\
$8192$ & $24.96\std{0.59}$ & $48.07\std{16.14}$ \\
\midrule
\multicolumn{3}{@{}l}{\emph{SWD projections} (default: $2048$)} \\
$16$ & $29.24\std{0.34}$ & $150.38\std{6.05}$ \\
$64$ & $25.63\std{0.20}$ & $82.13\std{3.71}$ \\
$256$ & $24.88\std{0.32}$ & $52.94\std{12.83}$ \\
$1024$ & $25.01\std{0.32}$ & $48.03\std{4.93}$ \\
$4096$ & $25.36\std{0.43}$ & $45.81\std{8.22}$ \\
\midrule
\multicolumn{3}{@{}l}{\emph{Divergence} (default: SWD-$2048$)} \\
Exact OT & $36.60\std{0.02}$ & $276.63\std{0.29}$ \\
MMD & $36.70\std{0.05}$ & $230.28\std{1.27}$ \\
\bottomrule
\end{tabularx}
\endgroup
\end{table*}

\clearpage
\newpage

\subsection{
    Additional samples from Bayesian posteriors on MNIST
}\label{sub:outsourced-additional-mnist}
\begin{figure}[!h]
    \centering
    \begin{tabular}{@{} *{3}{c@{}}}
         \osis & \osis + MALA & Rejection sampling  \\
         \includegraphics[width=0.31\linewidth, trim=0 0 0 50, clip]{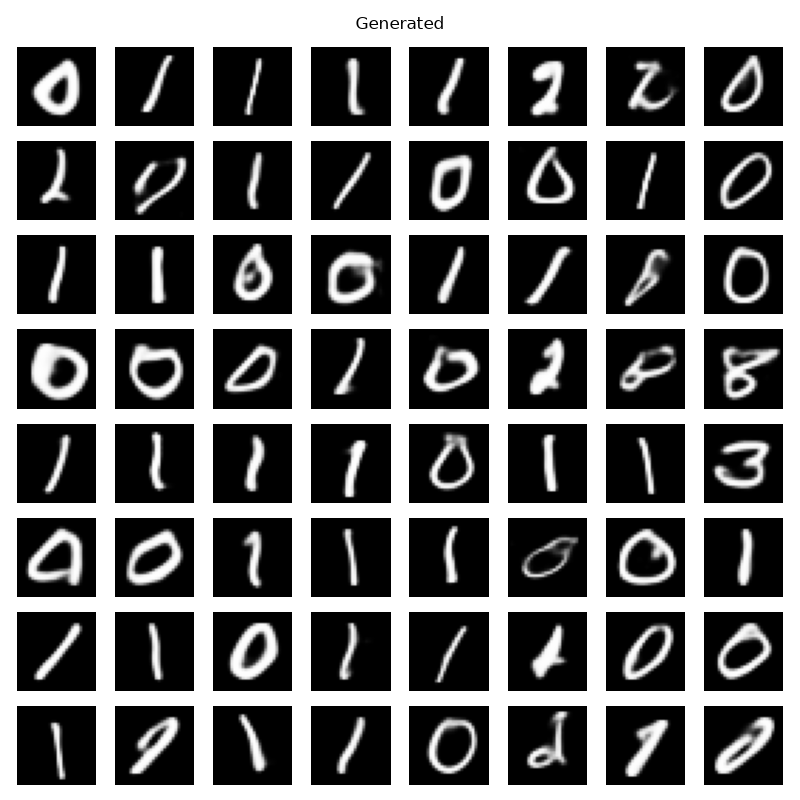} & 
         \includegraphics[width=0.31\linewidth, trim=0 0 0 50, clip]{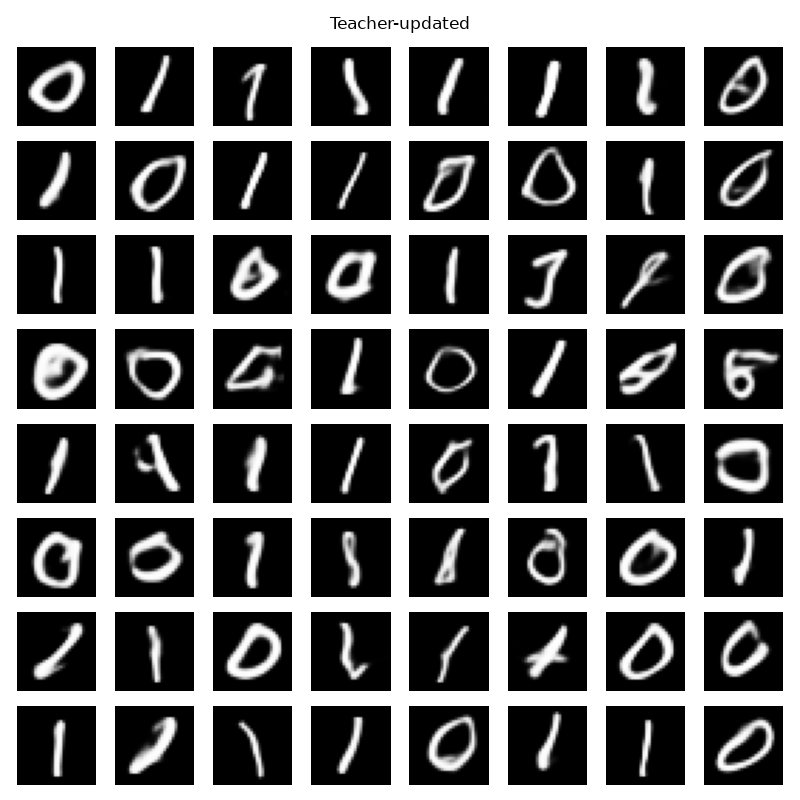} & 
         \includegraphics[width=0.31\linewidth, trim=0 0 0 50, clip]{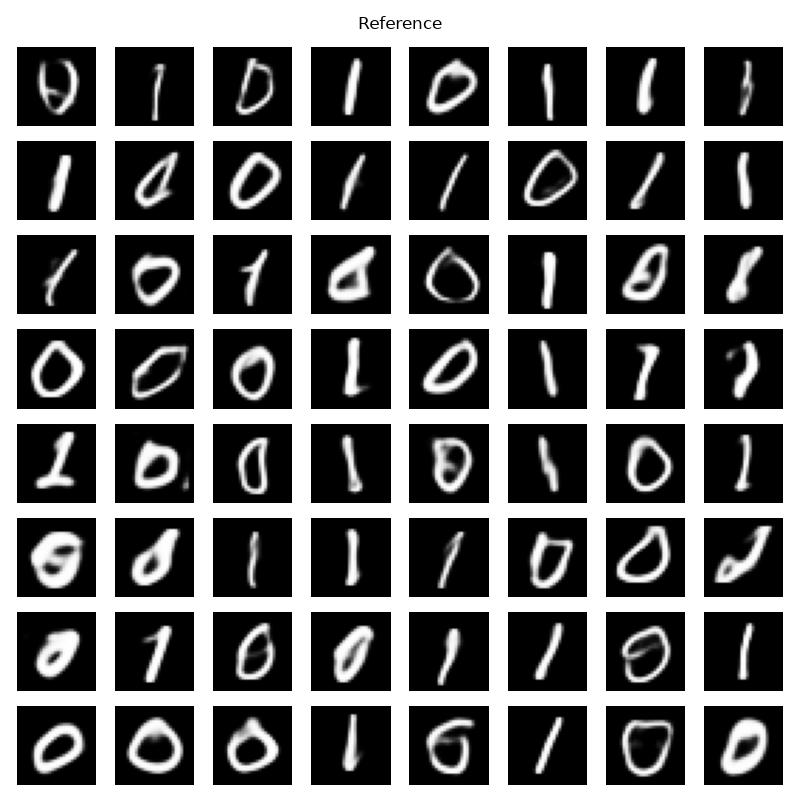} \\
    \end{tabular}
    \caption{
        Samples obtained using outsourced sampling. Samples correspond to the 
        classifier-based target with the reward $r(x)=0.5p(0|x) + 0.5p(1|x)$. 
        From left to right, samples are obtained with \osis, \osis with 20-step MALA refinement, 
        and rejection sampling.
        }
\end{figure}
\begin{figure}
    \centering
    \begin{tabular}{@{} *{3}{c@{}}}
         \osis & \osis + MALA & Rejection sampling  \\
         \includegraphics[width=0.31\linewidth, trim=0 0 0 50, clip]{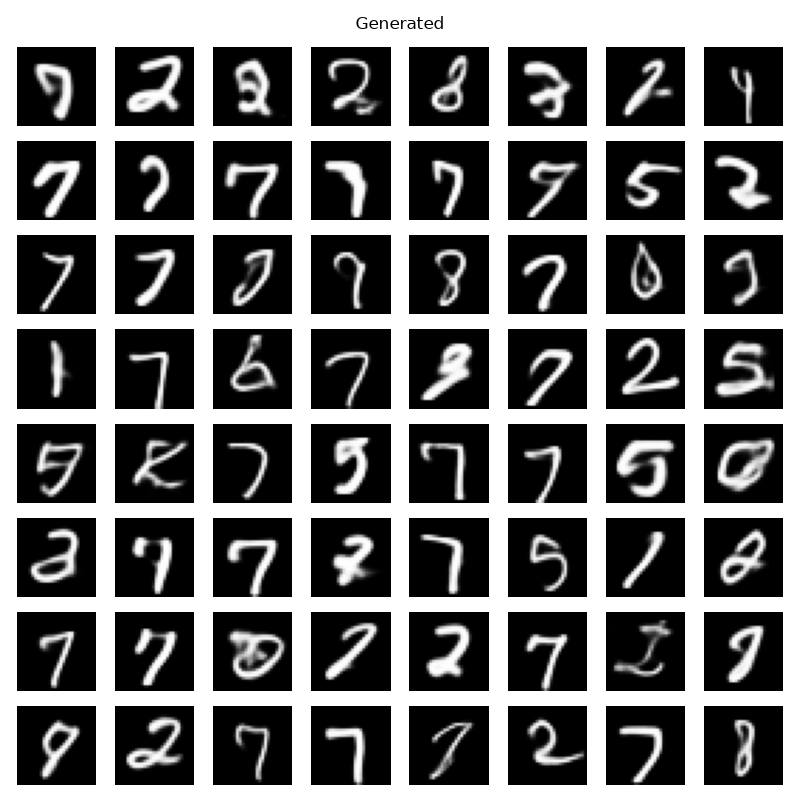} & 
         \includegraphics[width=0.31\linewidth, trim=0 0 0 50, clip]{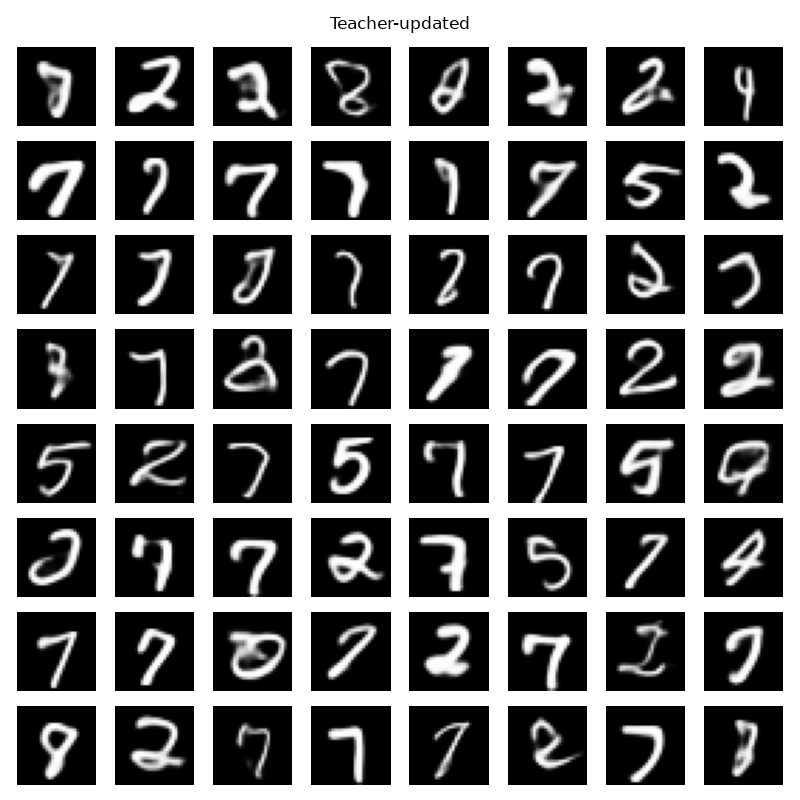} & 
         \includegraphics[width=0.31\linewidth, trim=0 0 0 50, clip]{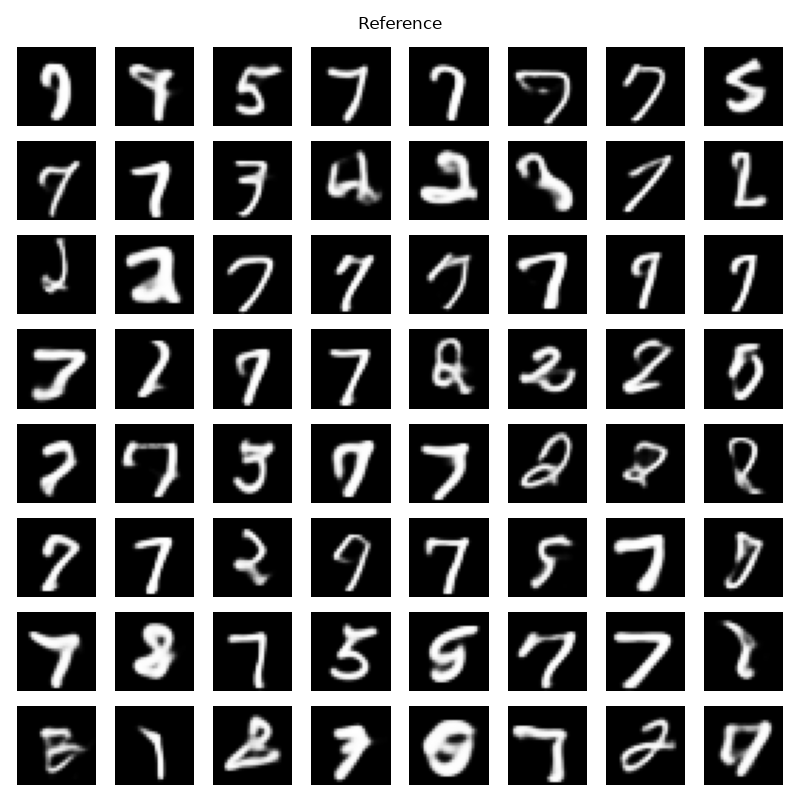} \\
    \end{tabular}
    \caption{
        Samples obtained using outsourced sampling. Samples correspond to the 
        classifier-based target with the reward $r(x)=1p(2\mid x) + 0.5p(5\mid x) + 2p(7\mid x)$.
        From left to right, samples are obtained with \osis, \osis with 20-step MALA refinement, 
        and rejection sampling.
        }
\end{figure}

\clearpage
\newpage
\subsection{
    Additional samples from Bayesian posteriors on FFHQ-64
}\label{sub:outsourced-additional-ffhq}

\begin{figure}[h!]
    \centering
    \begin{tabular}{@{} *{4}{c@{\hspace{3pt}}}}
         & \osis & \osis + MALA & Rejection sampling  \\
         \raisebox{7mm}{\small \raisebox{0.5cm}{\rotatebox[origin=c]{90}{\textit{a young blond girl}}}} &
         \includegraphics[width=0.31\linewidth, trim=0 0 0 0, clip]{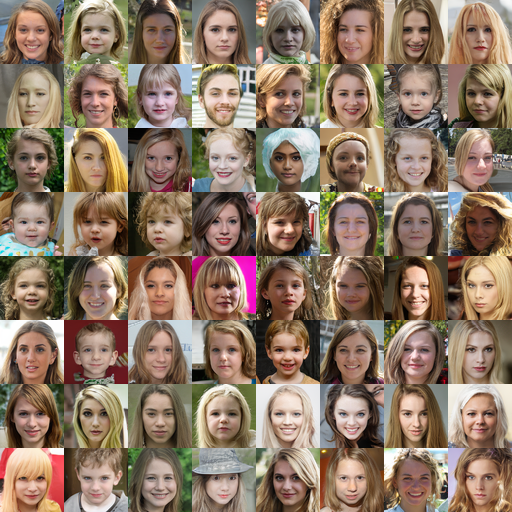} &
         \includegraphics[width=0.31\linewidth, trim=0 0 0 0, clip]{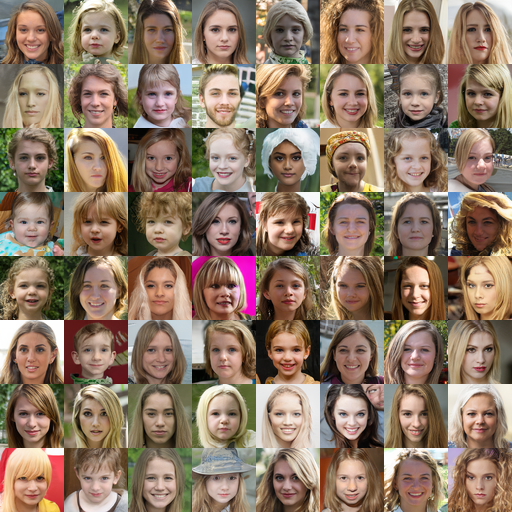} &
         \includegraphics[width=0.31\linewidth, trim=0 0 0 0, clip]{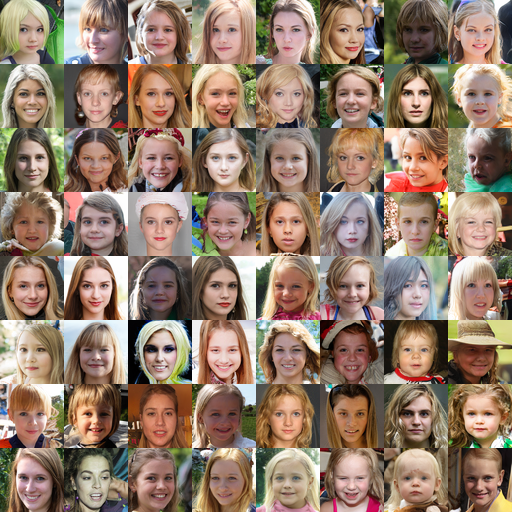} \\

         \raisebox{5mm}{\small \raisebox{0.5cm}{\rotatebox[origin=c]{90}{\textit{smiling toddler}}}} &
         \includegraphics[width=0.31\linewidth, trim=0 0 0 0, clip]{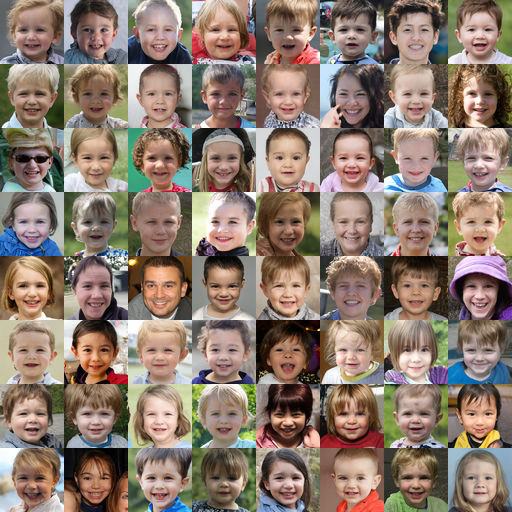} &
         \includegraphics[width=0.31\linewidth, trim=0 0 0 0, clip]{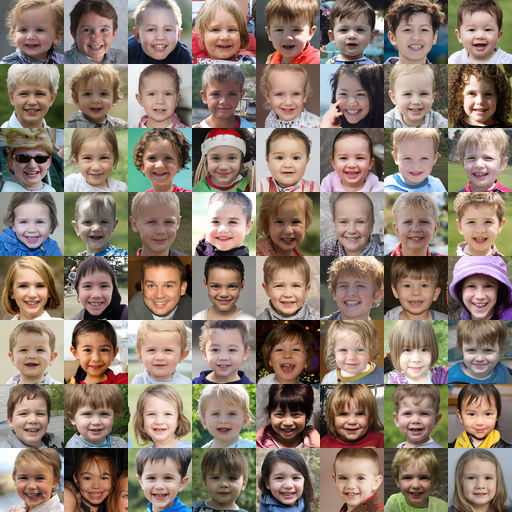} &
         \includegraphics[width=0.31\linewidth, trim=0 0 0 0, clip]{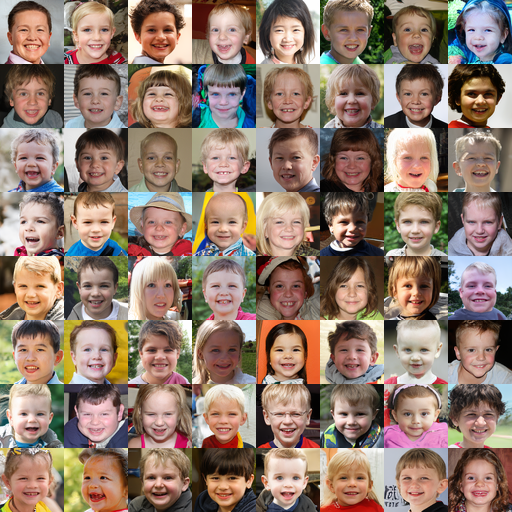} \\

         \raisebox{7mm}{\small \raisebox{0.5cm}{\rotatebox[origin=c]{90}{\textit{a man with a beard}}}} &
         \includegraphics[width=0.31\linewidth, trim=0 0 0 0, clip]{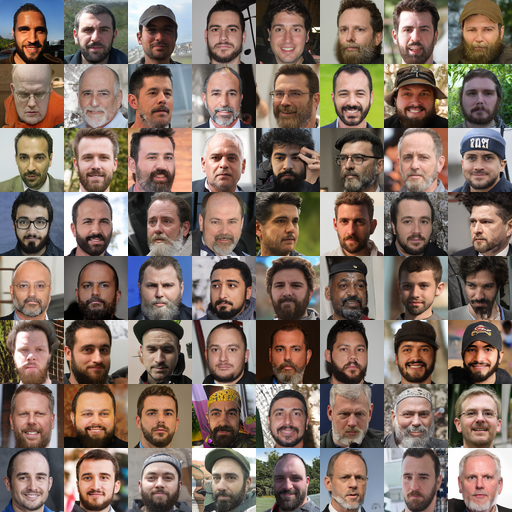} &
         \includegraphics[width=0.31\linewidth, trim=0 0 0 0, clip]{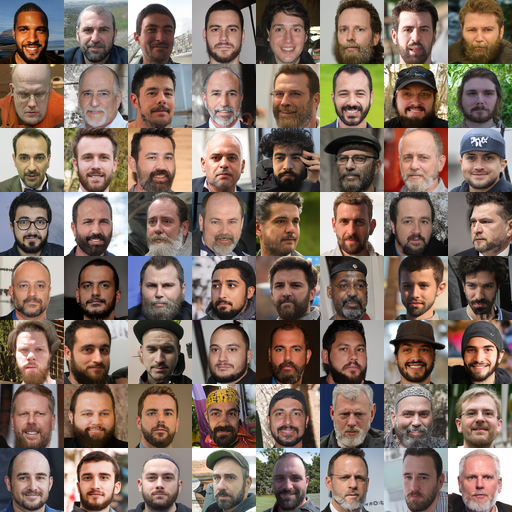} &
         \includegraphics[width=0.31\linewidth, trim=0 0 0 0, clip]{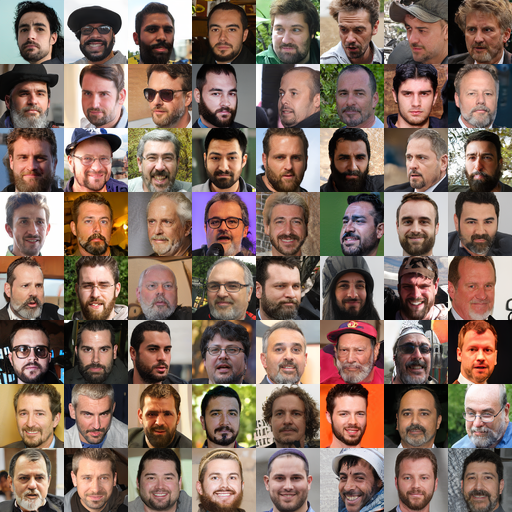} \\

         \raisebox{9mm}{\small \raisebox{0.5cm}{\rotatebox[origin=c]{90}{\textit{person wearing glasses}}}} &
         \includegraphics[width=0.31\linewidth, trim=0 0 0 0, clip]{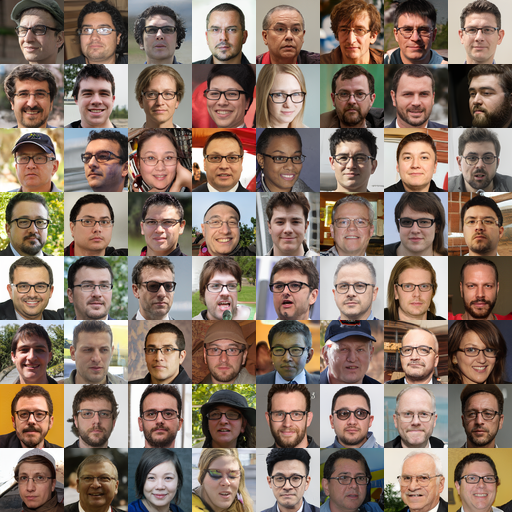} &
         \includegraphics[width=0.31\linewidth, trim=0 0 0 0, clip]{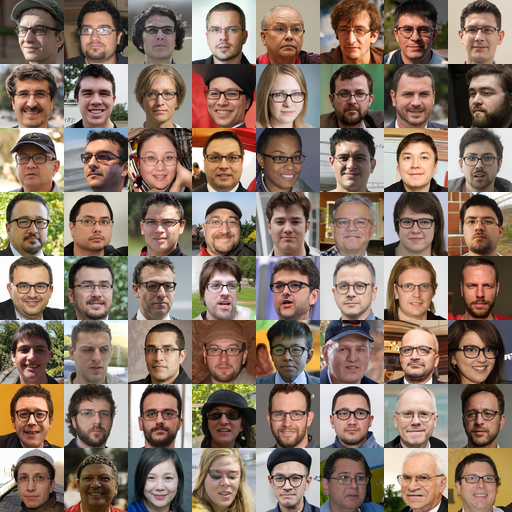} &
         \includegraphics[width=0.31\linewidth, trim=0 0 0 0, clip]{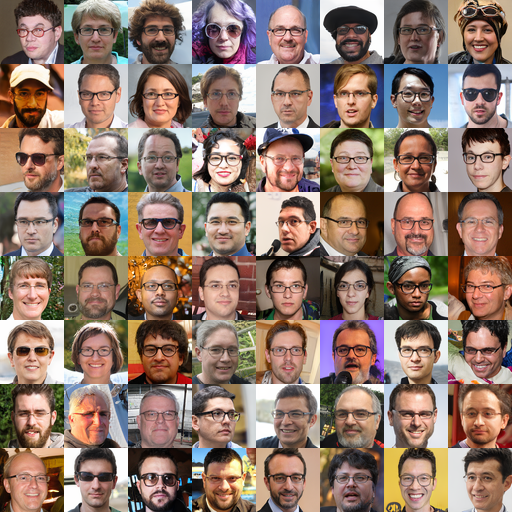} \\
    \end{tabular}
    \caption{
        Comparison of samples obtained with \osis, \osis with 20-step MALA refinement, 
        and rejection sampling on outsourced sampling task with 
        ImageReward-based target. The following 4 prompts are considered (from to to bottom): 
        \textit{a young blond girl}, \textit{smiling toddler}, \textit{a man with a beard}, \textit{person wearing glasses}.
        }
\end{figure}
\begin{figure}[h!]
    \centering
    \begin{tabular}{@{} *{2}{c@{\hspace{3pt}}}}
         \osis &  Rejection sampling  \\
         \includegraphics[width=0.49\linewidth, trim=0 0 0 0, clip]{images/outsourced/appx-ffhq64/blond_osis.png} &
         \includegraphics[width=0.49\linewidth, trim=0 0 0 0, clip]{images/outsourced/appx-ffhq64/blond_rs.png} \\
         TB$+$MALA \citep{sendera2024improved} &  MALA  \\
         \includegraphics[width=0.49\linewidth, trim=0 0 0 0, clip]{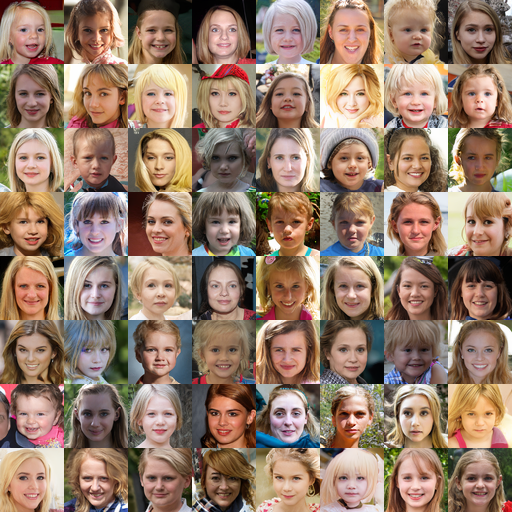} &
         \includegraphics[width=0.49\linewidth, trim=0 0 0 0, clip]{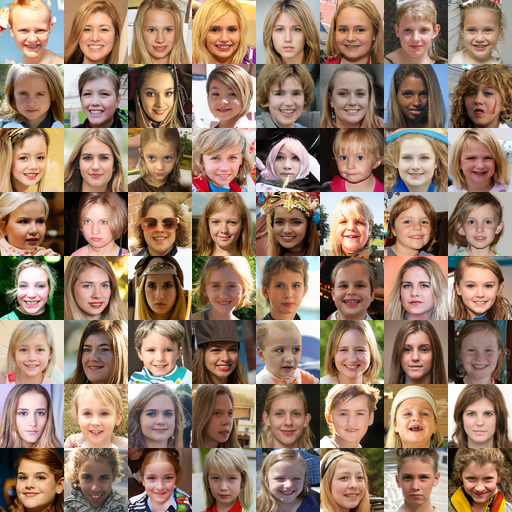} \\
    \end{tabular}
    \caption{
        Visualisation of samples from \osis, rejection sampling, TB$+$MALA \citep{sendera2024improved} 
        and MALA. All algorithms, except rejection sampling, are matched in terms of the number of log-density evaluations. 
        }
\end{figure}
\end{document}